%% file: main_arxiv.tex
\documentclass[journal=jprobs,manuscript=article]{achemso}

\usepackage{chemformula}
\usepackage[T1]{fontenc}
\usepackage{tabularx}
\usepackage{graphicx}
\usepackage{subcaption}
\usepackage[symbol]{footmisc}
\usepackage{amsfonts}
\usepackage{bbm}
\usepackage{color}
\usepackage{wrapfig}
\usepackage[normalem]{ulem}

\usepackage{makecell}
\usepackage{titlesec}
\usepackage{mciteplus}
\mciteErrorOnUnknownfalse % Add this line to suppress the error

\newcommand{\G}{{\mathcal{G}}}
\newcommand{\V}{{\mathcal{V}}}
\newcommand{\E}{{\mathcal{E}}}
\newcommand{\N}{{\mathcal{N}}}

\renewcommand{\sout}[1]{\skip}
\definecolor{red}{rgb}{0,0,0}

\author{Zheng Ma$^{1, \dag}$, Jiazhen Chen$^{2, \dag}$, Lei Xin$^{3}$, Ali Ghodsi$^{2,\ddag,}$}

\affiliation{%
\textit{$^{1}$Cheriton School of Computer Science, University of Waterloo, Waterloo, ON, Canada, N2L 3G1} \\
\textit{$^{2}$Department of Statistical and Actuarial Science, University of Waterloo, Waterloo, ON, Canada, N2L 3G1} \\
\textit{$^{3}$Bioinformatics Solutions Inc., Waterloo, ON, Canada, N2L 3K8}\\
\thanks{$^{\dag}$Equal Contribution}\\
\thanks{$^{*}$Corresponding author} \\
\textit{$\ddag$Current address: 200 University Ave W, Waterloo, ON, Canada, N2L 3G1}
}

\email{ali.ghodsi@uwaterloo.ca}
\phone{+1 (519) 888-4567 x47316}

\title[GraphPI]
{GraphPI: Efficient Protein Inference with Graph Neural Networks}

\abbreviations{GNN,PI,ML}

\begin{document}

\footnote{This article is published in Journal of Proteome Research, 2024, American Chemical Society. DOI: \url{https://doi.org/10.1021/acs.jproteome.3c00845}.

The article can be accessed via ACS Article on Request, \url{https://pubs.acs.org/articlesonrequest/AOR-GESZT64SEDG6BQKSFSTX?_gl=1*j2intf*_ga*MTg3NjgwOTE1Ni4xNzUyMTAzMTM1*_ga_XP5JV6H8Q6*czE3NTIxMDMxMzQkbzEkZzAkdDE3NTIxMDMxMzQkajYwJGwwJGgw}.}

\begin{abstract}
The integration of deep learning approaches in biomedical research has been transformative, enabling breakthroughs in various applications. Despite these strides, its application in protein inference is impeded by the scarcity of extensively labeled datasets, a challenge compounded by the high costs and complexities of accurate protein annotation. In this study, we introduce GraphPI, a novel framework that treats protein inference as a node classification problem. We treat proteins as interconnected nodes within a protein-peptide-PSM graph, utilizing a Graph Neural Network-based architecture to elucidate their interrelations. To address label scarcity, we train the model on a set of unlabeled public protein datasets with pseudo-labels derived from an existing protein inference algorithm, enhanced by self-training to iteratively refine labels based on confidence scores. Contrary to prevalent methodologies necessitating dataset-specific training, our research illustrates that GraphPI, due to the well normalized nature of Percolator features, exhibits universal applicability without dataset-specific fine-tuning, a feature that not only mitigates the risk of overfitting but also enhances computational efficiency. Our empirical experiments reveal notable performance on various test datasets and deliver significantly reduced computation times compared to common protein inference algorithms.

\end{abstract}

%%%%%%%%%%%%%%%%%%%%%%%%%%%%%%%%%%%%%%%%%%%%%%%%%%%%%%%%%%%%%%%%%%%%%
%% Main paper
%%%%%%%%%%%%%%%%%%%%%%%%%%%%%%%%%%%%%%%%%%%%%%%%%%%%%%%%%%%%%%%%%%%%%
\input{chapters/introduction_revision}

\input{chapters/Methodology_revision}

\input{chapters/Experiments_revision}

\input{chapters/Conclusion}

%%%%%%%%%%%%%%%%%%%%%%%%%%%%%%%%%%%%%%%%%%%%%%%%%%%%%%%%%%%%%%%%%%%%%
%% Acknowledgements
%%%%%%%%%%%%%%%%%%%%%%%%%%%%%%%%%%%%%%%%%%%%%%%%%%%%%%%%%%%%%%%%%%%%%
\section*{Acknowledgements}
The authors received support from Natural Sciences and Engineering Research Council of Canada (NSERC) Discovery Grant (no. 50503-10904-500).

%%%%%%%%%%%%%%%%%%%%%%%%%%%%%%%%%%%%%%%%%%%%%%%%%%%%%%%%%%%%%%%%%%%%%
%% Supporting information summary
%%%%%%%%%%%%%%%%%%%%%%%%%%%%%%%%%%%%%%%%%%%%%%%%%%%%%%%%%%%%%%%%%%%%%
\begin{suppinfo}

The codes and test data analysis are available via Github: \\ 
\url{https://github.com/hearthewind/graphpi_protein_inference.git}

The supporting information is attached to the end of this manuscript.

\begin{itemize}
\item{Table S1}: PSM features utilized by our model, along with their descriptions.
\item{Table S2}: Information about training datasets, including search parameters, identifier, and other metainfo.
\item{Material S1}: Information for testing datasets, including search parameters, identifier/link, data size, and number of raw files.
\item{Figure S1}: Pie chart for each testing dataset, the portion of peptides that are shared by multiple proteins.
\item{Figure S2}: Pie chart for each testing dataset, the portion of groundtruth proteins sharing peptides with contaminate proteins.
\item{Table S3}: Software versions and configurations in our experiments.
\item{Figure S3}: The relationship between decoy FDR and entrapment FDR for various methods under various test datasets.
\item{Figure S4}: The number of identified proteins under different decoy FDR values, for GraphPI and Epifany.
\item{Material S2}: Alternative settings for pretraining, showing the model performance when GraphPI is pretrained using non-human training datasets, or non-isoform database.
\item{Figure S5}: pAUC score of GraphPI on test datasets trained on human isoform,
human non-isoform, or non-human isoform database.
\item{Material S3}: Evaluation of our model on non-human datasets with decoy FDR.
\item{Figure S6}: pAUC scores of GraphPI and Epifany on PXD datasets with decoy FDR.
\item{Table S4}: Information about additional non-human datasets.
\end{itemize}

\end{suppinfo}

\newpage

%%%%%%%%%%%%%%%%%%%%%%%%%%%%%%%%%%%%%%%%%%%%%%%%%%%%%%%%%%%%%%%%%%%%%
%% Bibliography
%%%%%%%%%%%%%%%%%%%%%%%%%%%%%%%%%%%%%%%%%%%%%%%%%%%%%%%%%%%%%%%%%%%%%
\bibliography{reference}

%%%%%%%%%%%%%%%%%%%%%%%%%%%%%%%%%%%%%%%%%%%%%%%%%%%%%%%%%%%%%%%%%%%%%
%% Supporting material / appendix
%%%%%%%%%%%%%%%%%%%%%%%%%%%%%%%%%%%%%%%%%%%%%%%%%%%%%%%%%%%%%%%%%%%%%
\clearpage
\appendix

% Reset counters for supporting material
\setcounter{table}{0}
\setcounter{figure}{0}
\setcounter{page}{1}

% Use S-numbering only for supporting material
\renewcommand{\thetable}{S\arabic{table}}
\renewcommand{\thefigure}{S\arabic{figure}}
\renewcommand{\thepage}{S\arabic{page}}

\input{chapters/Supplementary}

\input{chapters/Supplementary_nonhuman_comparison}

\end{document}

%% file: chapters/introduction_revision.tex
\section{Introduction}

%{Protein inference refers to the process of determining the proteins present in a biological sample by analyzing the peptides detected through tandem mass spectrometry (MS/MS) experiments. It is an important area of research in proteomics because accurate identification of proteins is critical for understanding their functions and roles in biological systems. Protein inference results serve as the foundation for various downstream applications, including the identification of biomarkers, drug targets, and the annotation of protein functions. }

Understanding the proteins in a biological sample is crucial for unraveling their functions and roles in biological systems. Protein inference, which involves identifying proteins through peptides detected in tandem mass spectrometry (MS/MS) experiments, is fundamental to proteomics. Accurate protein identification is essential for applications such as discovering biomarkers, identifying drug targets, and annotating protein functions, which are vital for advancing personalized medicine and therapeutic strategies~\citep{anderson2002humanplasmaproteome}~\citep{diamandis2004massspec}~\citep{hopkins2002druggable}.

% Despite technological advances, challenges such as peptide-to-protein ambiguity persist, making reliable protein inference critical. Addressing these challenges improves the interpretation of proteomics data, thereby enhancing our understanding of complex biological processes ~\citep{rask2011trends}~\citep{kanehisa2004kegg}~\citep{ashburner2000gene}.
% }

% \textcolor{red}{Accurate protein inference is essential in proteomics, as it serves as the foundation for understanding protein functions and their roles in biological systems. This accuracy is critical for numerous downstream applications, such as identifying biomarkers, drug targets, and annotating protein functions, which are vital for advancing personalized medicine and therapeutic strategies~\citep{anderson2002humanplasmaproteome}~\citep{diamandis2004massspec}~\citep{hopkins2002druggable}. However, achieving precise protein inference from tandem mass spectrometry (MS/MS) data remains challenging due to the inherent complexities and limitations that arise in the process~\citep{rask2011trends}~\citep{kanehisa2004kegg}~\citep{ashburner2000gene}.}

In an MS/MS experiment, proteins are initially digested with some proteolytic enzyme, like trypsin, into peptides. The peptide mixture is then passed through a mass spectrometer, generating MS1 spectra. One prominent peptide, or a group of peptides, at a time is selected from the MS1 spectra and further fragmented into fragment ions, and mass spectrometer will capture the $m/z$ values and intensities of these fragment ions, resulting in a specific mass spectrum signature for each peptide, named the MS2 spectrum. Next, the acquired MS2 spectra are matched to a peptide database to detect which peptides are present in the sample. Finally, the peptide profile is used to predict which proteins are more likely to produce the observed peptide set. 

However, high dynamic range of protein abundance, as well as limitations in digestion and mass spectrometry often lead to ambiguous peptide identification result~\citep{pfeuffer2020epifany}~\citep{serang2010efficient}. In addition, the existence of shared peptides and one-hit proteins further complicates the problem~\citep{pfeuffer2020epifany}. Shared peptides are those that can be derived from multiple (degenerate) proteins. Those peptides introduce ambiguity in associating detected peptides with their respective proteins. One-hit proteins are those that are backed by single-peptide evidence, making it difficult to confidently infer the presence of this protein. The issue is amplified as some peptides are more prone to detection, skewing protein identification toward those producing such peptides. These elements together complicate the protein inference process, creating an intricate, complex landscape for protein identification. 

% \sout {Various approaches have been developed to address the challenges associated with protein inference. However, protein inference remains a difficult problem due to challenges such as the limitations of digestion~\citep{pfeuffer2020epifany}, which can result in subtle patterns that are difficult for algorithms to decode. This necessitates the development of better algorithms capable of interpreting more subtle details revealed in the data, thereby advancing the performance of protein inference. } 
Various approaches have been developed to address the challenges associated with protein inference, with limited success due to the fundamental challenges mentioned above. ProteinProphet~\citep{nesvizhskii2003statistical}, an early innovator in this field, employs a heuristic-based probabilistic model to estimate protein probabilities, accounting for factors such as shared peptides and the quantity of peptides identified for each protein. PIA~\citep{uszkoreit2015pia}, a rule-based algorithm, conducts inference by identifing the minimal set of proteins that most accurately accounts for the observed Peptide-Spectrum Matches (PSMs).

In addition to these methods, Bayesian networks have emerged as a highly efficacious technique for protein inference. Fido~\citep{serang2010efficient} was the first to implement Bayesian networks in this domain, incorporating simple yet rational assumptions. Specifically, for proteins sharing peptides, the method gives preference to those with independent evidence. Concurrently, the ``explain away'' effect reduces the scores for proteins without distinct evidence. For instance, if two proteins share a common peptide with no other evidence, their scores should be equal due to symmetry. If one protein gains  unique peptide evidence, its score should be elevated due to this new evidence. Concurrently, the score of the protein without unique evidence should be lowered, as the unique peptide evidence for the first protein effectively ``explains away'' the shared peptide, reducing the likelihood of the second protein being present.
Epifany~\citep{pfeuffer2020epifany} utilizes a Bayesian network akin to Fido's, but incorporates additional priors to regularize the inference process and employs a rapid approximation inference algorithm to enhance computational efficiency. 
Nevertheless, these techniques are constrained by the inherent limitations of Bayesian networks,
which include high computational costs in terms of time and memory, and susceptibility to prior probabilities~\citep{Ahmed2008}~\citep{Neapolitan2009}.

With the advent of neural network advancements, deep learning techniques have delivered promising results in biomedical research, including, but not limited to, prediction of peptide properties from tandem mass spectra~\citep{guan2019prediction}, peak detection~\citep{zohora2019deepiso}, peptide database search~\citep{diann}, and de novo sequencing of peptides~\citep{tran2017novo}~\citep{qiao2021computationally}~\citep{tran2019deep}. However, given that deep learning methods usually require a massive amount of labeled data for training, it is challenging to apply them in the field of protein inference. The scarcity of labeled data in this field, possibly because of the prohibitive costs of accurately annotating the proteins in a biological sample, presents a considerable barrier.

In recent years, a limited number of deep learning-based methods have been proposed to address the protein inference problem. For instance, Barista~\citep{spivak2012direct} trains a binary classification network using decoy proteins (in silico-generated by shuffling or reversing real sequences) as negative labels and real proteins as positive labels.
Thereby the classification scores can be directly adopted as the protein scores. However, this methodology can inadvertently infer an inaccurate classification boundary, as not all assumed positive proteins are actually present in a given sample. Secondly, the strategy also runs the risk of overfitting decoy proteins, which are conventionally employed to discern truly present proteins based on a predefined FDR threshold~\citep{elias2007target}. Moreover, the neural network design of Barista \citep{spivak2012direct} does not allow the feature of one protein to affect another, making it unable to take advantage of the ``explain away'' effect of a Bayesian network. 

Contrastingly, DeepPep~\citep{kim2017deeppep} leverages self-supervised learning, using peptide scores to circumvent the problem of protein label scarcity. Specifically, it utilizes peptide identification scores as labels and trains a model to predict these scores based on all proteins. By sequentially removing each protein from the input, the model can infer the contribution of each protein to each peptide. The final protein score is an aggregation of each protein's contribution to all potential peptides based on their identification scores. Nonetheless, this approach suffers from computational inefficiency, as it necessitates iterating over all proteins for each peptide. 
% \sout{Furthermore, using an self-supervised objective is less than ideal, since training objective does not directly align with testing objective.} 
Moreover, the self-supervised objective introduces a misalignment between the training and testing objectives, further diminishing its effectiveness. Therefore, despite the potential of existing deep learning methods, they remain somewhat marginalized within the protein inference field due to their subpar performance relative to Bayesian networks. For instance, DeepPep~\citep{kim2017deeppep} underperforms by over 25\% compared to the best performing Bayesian method.

In this study, we present GraphPI, a novel deep learning-based framework to address the protein inference problem. Drawing inspiration from Bayesian network techniques, we design a protein-peptide-spectrum graph structure with uniquely crafted node and edge features. 
% \textcolor{red}{A graph is a collection of nodes (vertices) and edges (connections). In our context, the nodes represent proteins, peptides, and Peptide-Spectrum Matches (PSMs), forming a tripartite graph with edges forming between proteins and peptides, and between peptides and PSMs, representing the connections established through the protein digestion and mass spectrometry process.}

\begin{figure}[t!]
\centering
    \includegraphics[width=6.3in]{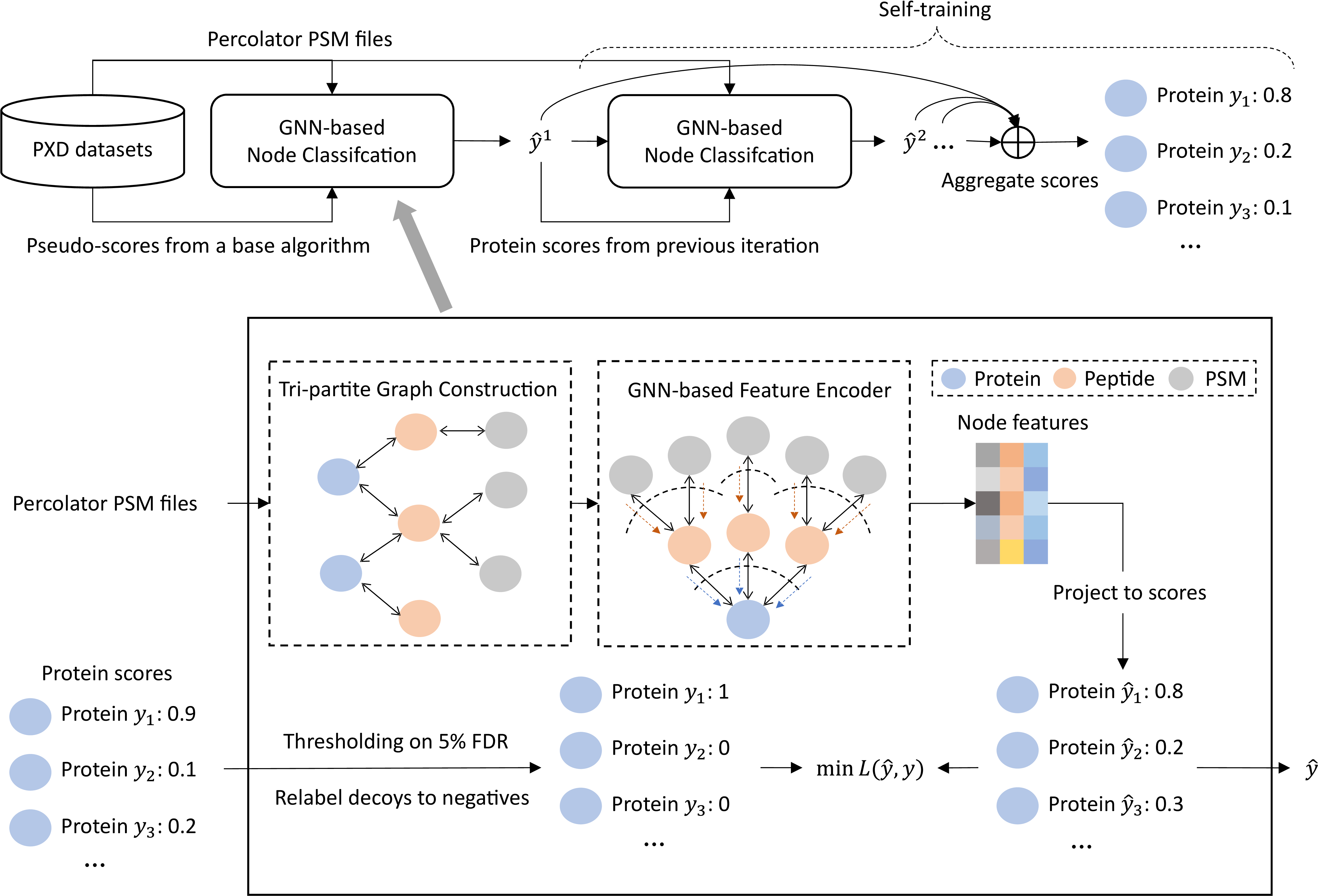}  
    %\vspace{-0.55cm}
    \caption{The architecture of our protein inference algorithm leveraging GNN for node classification. The process begins with the input of Percolator~\citep{kall2007semi} PSM files into the GNN-based Node Classification module. This module utilizes pseudo-scores from an established protein algorithm to guide the initial classification. Within the module, a tripartite graph is constructed, linking proteins, peptides, and PSMs, as indicated in the detailed view. The algorithm then employs a self-training strategy over multiple iterations to refine the protein scores, as illustrated in the top sequence. Finally, the iterative process yields aggregated confidence scores for each protein, denoted by $y_i$, which are presented on the right. These resulting scores reflect the cumulative learning and adjustment from the iterative self-training, yielding a robust set of protein identifications.}
\label{fig:overall_structure}
\end{figure}

A graph consists of nodes (vertices) and edges (connections). In our case, the nodes represent proteins, peptides, and Peptide-Spectrum Matches (PSMs), forming a tripartite structure. Edges connect proteins to peptides, and peptides to PSMs, reflecting the relationships formed during the protein digestion and mass spectrometry process.
This enables us to perceive candidate proteins as interconnected entities rather than isolated individuals. The protein inference problem can be then formulated as a node classification problem with protein scores generated directly from the node classification scores. 
To process the protein-peptide-spectrum tripartite graph, we employ Graph Neural Networks (GNNs)\citep{kipf2016semi}, which learn node representations by recursively aggregating information from neighbors. However, standard GNNs are not well-suited to handle the heterogeneous nature of our graph, which consists of different types of nodes and edges. To address this issue, we develop a tailored GNN architecture based on GraphSAGE\citep{hamilton2017inductive}, designed to effectively manage this heterogeneity.
To mitigate the label scarcity issue and enhance computation efficiency, our model is primarily trained on a set of large and unlabeled public protein datasets from ProteomeXchange (\url{https://proteomecentral.proteomexchange.org/}) in a semi-supervised learning setting, using pseudo-labels generated by an existing protein inference algorithm as the base model. We further refine these labels by introducing hard negative decoy protein information, allowing the model to surpass capabilities of the base model and produce improved test results. Finally, we perform self-training to further enhance our model's performance by iteratively refining labels based on their confidence scores. Figure~\ref{fig:overall_structure} presents the overall pipeline of GraphPI. 
Contrary to alternative approaches which necessitate the execution of training or fine-tuning processes for each individual dataset, our analysis indicates that the data pertaining to peptide identification exhibits considerable normalization across all test datasets. This standardization facilitates the application of a single model, which can be trained and subsequently assessed universally, thus circumventing the issue of overfitting. Additionally, this method enhances computational efficiency by mitigating the need for repetitive training processes for disparate datasets.

To the best of our knowledge, we are the first to apply GNNs and a semi-supervised training scheme to the protein inference problem, and the experiments demonstrate that our approach achieves competitive performance across diverse test datasets. Additionally, we leverage the inherently parallelizable structure of neural networks, leading to considerably faster computations in comparison to existing methods. Moreover, our model, pre-trained on a wealth of publicly available datasets, is adept at performing inference instantaneously during real-time applications. As a result, the improvements in efficiency facilitate our model to handle protein inference tasks, seamlessly scaling to accommodate large protein datasets.

%% file: chapters/Methodology_revision.tex
\section{Methods}

\subsection{Overview}
We present a deep-learning framework, GraphPI, for protein inference that avoids the need for labeled protein datasets, and offers improved computational efficiency compared to existing methods. Our approach leverages a semi-supervised binary classification model that is trained on a set of protein datasets, which are pseudo-labeled by an existing protein inference method. Each dataset is represented as a tripartite graph and encoded using a GNN network that accommodates the heterogeneity of node and edge types.
Following training on the pseudo-labels, we implement a self-training procedure that refines the labels based on protein probability scores followed by a fresh retraining, and we iterate this process for multiple rounds. The final protein scores are calculated as an ensemble of the models from all rounds. To ensure the generalizability of our model to real-world data, we utilize a variety of experimental datasets from published biological research. The following sections delineate our methodology in three parts: 1) Construction of Tri-partite Graph: the conversion of protein datasets into tripartite graphs; 2) GNN Model Architecture: the elaboration of our model's architecture for encoding these graphs into latent representations amenable to deep learning; 3) Training: a detailed exposition of our training paradigm. This paradigm leverages the power of semi-supervised learning with an iterative self-training mechanism that begins with pseudo-labels and evolves to deliver refined, reliable inferences.

\subsection{Construction of Tri-partite Graph}

\label{sec:tri-partite graph}
For the protein inference problem, understanding the relationship between proteins, peptides, and their associated PSMs is paramount since the identification of each protein is closely tied to its constituent peptides, which in turn is validated by the confidence of their PSMs.
Given the inherent interconnectedness of these components, a graphical formulation naturally emerges as an apt choice. 
% \sout{By representing the data as a graph, we can capture the complex dependencies among proteins, peptides, and PSMs through GNNs, and then, we could leverage the learned protein node representation, which consists of multi-hop information from its related peptides and PSMs to generate the protein score.}
By representing the data as a graph, we can capture the complex dependencies among proteins, peptides, and PSMs through GNNs, which learn robust node representations by recursively aggregating information from neighboring nodes. Then, we leverage the learned protein node representation, which includes information propagated through connected peptides and PSMs, to generate the protein score.
% {\color{blue} Here the aggregated information is multi-hop because by design of GNN, each layer will aggregate information from its immediate neighbors. Thus with $k$ layers of GNN, the information from distance $k$ is aggregated.}
In the following paragraphs, the details of the graph are presented, focusing primarily on the nodes and edges, and their associated features. 

\begin{figure*}[t!]
\centering
\begin{subfigure}[t]{1\linewidth}
\centering
        \includegraphics[height=0.5cm]{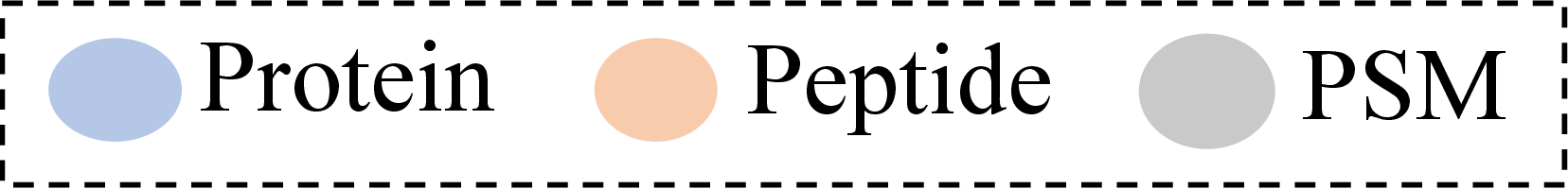} 
\end{subfigure}
\vspace{0.1cm}
\newline
\centering
\begin{subfigure}[t]{0.49\linewidth}
    \centering
    % \includesvg[height=2.9cm]{figures/swat_graph_deviation.svg}
    \includegraphics[width=6.5cm]{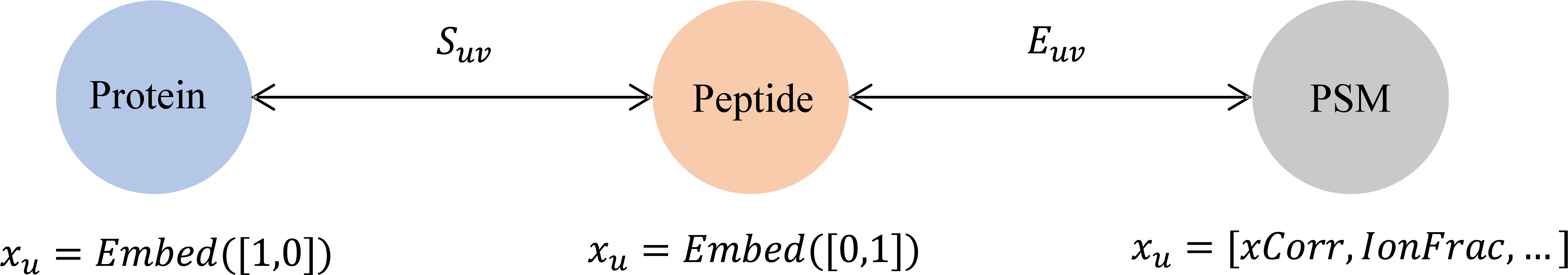}  
    %\vspace{-0.55cm}
    \caption{}
    \label{fig: schema}
\end{subfigure}
%\hspace{\fill} % maximize horizontal separation
\centering
\begin{subfigure}[t]{0.49\linewidth}
    \centering
\includegraphics[height=5.5cm]{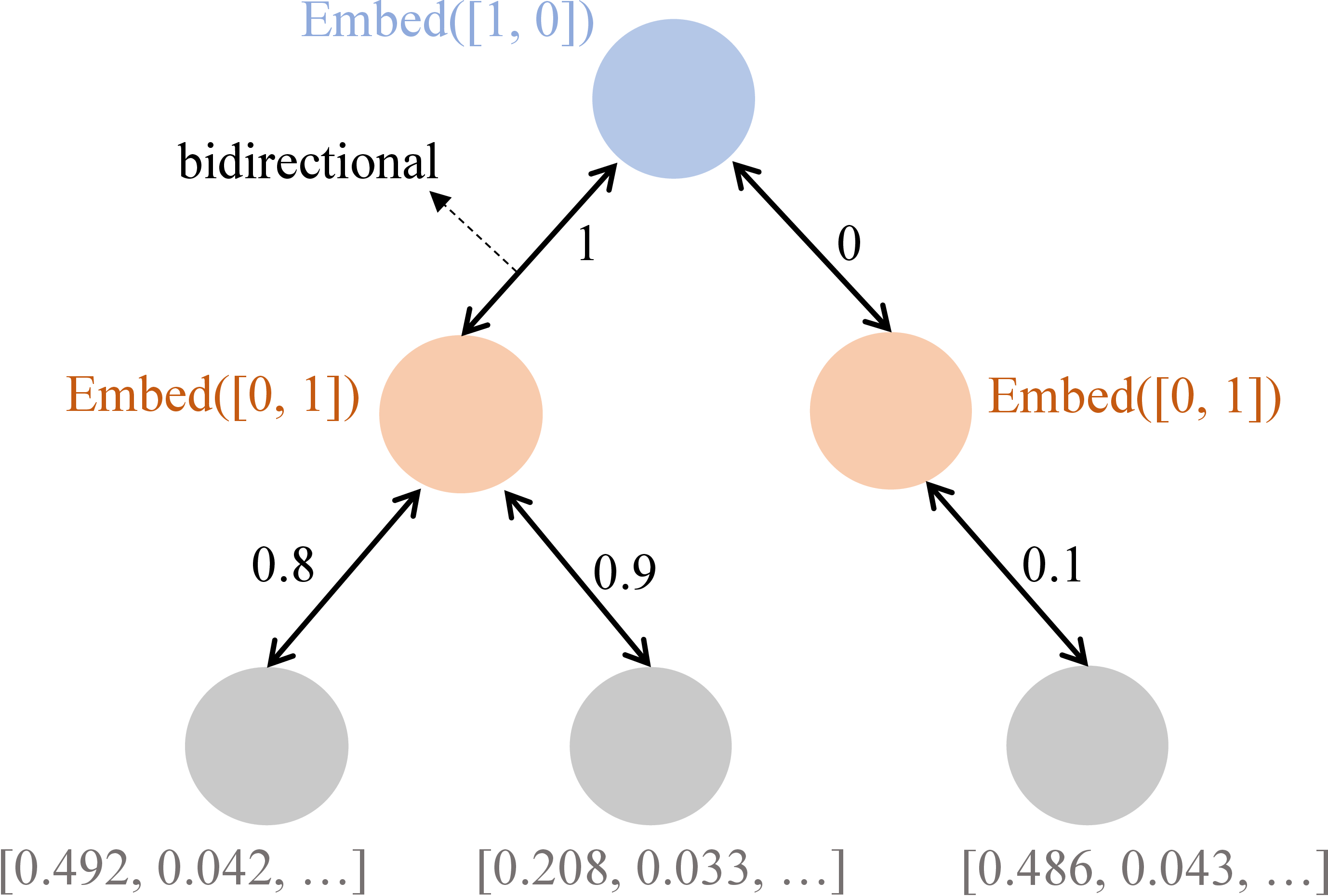}  
       % \vspace{-0.55cm}
    \caption{}
    \label{fig: graph example}
\end{subfigure}
\caption{(a): The schema of the bidirectional tri-partite graph, with $S_{uv}$ and $E_{uv}$ denote as the edge attribute for (protein, peptide) and (peptide, PSM) node pairs respectively, and $x_u$ is an example of the node feature vector for each type of the three nodes (one-hot embedding for protein and peptide nodes, and database search engine features for PSM nodes). (b): An illustrative example of the tripartite graph containing one protein surrounded by its peptide and PSM nodes within 2-hop.}
\end{figure*}

\subsubsection{Nodes in the Tri-partite Graph} 
The foundation of our graph lies in its nodes, which are categorized into three specific types: proteins, peptides, and PSMs. We prioritize PSMs over spectra as a node type due to their inherent adaptability in incorporating supplementary features. For instance, peptide search engines such as Percolator~\citep{kall2007semi} rely on a suite of PSM-centric features to compute peptide identification scores. In this study, we primarily utilize the PSM features from Percolator, which are presented as Table S1. In our model, these PSM features are harnessed not just for peptide identification but are further refined to aid in the prediction of protein scores. By doing so, we facilitate an end-to-end training process where peptide identification scores are optimized in alignment with the ultimate goal of predicting protein scores. Moreover, the volume of PSM nodes corresponding to a peptide is deemed valuable; a higher number of matches between peptides and spectra, especially with elevated identification scores, often indicates a heightened likelihood of peptide presence.

While PSM nodes come with pre-existing features from a database search engine, the nodes representing peptides and proteins lack such attributes. Nonetheless, GNNs necessitate node features. Possible strategies for feature assignment range from designating unique one-hot vectors for each node to allotting universal features (e.g., a vector with all ones) across all nodes. However, a unique one-hot vector for every node tends to restrict the graph learning to be transductive, inhibiting its generalization to datasets with varying node counts and structures. On the other hand, uniform features for peptide and protein nodes might
compromise the GNN's ability to differentiate between messages from diverse node types. To circumvent these challenges while maintaining inductive capabilities, we allocate two separate one-hot vectors for protein and peptide nodes. Additionally, learnable embedding layers are applied on top of the one-hot vectors to make them dimensionally equivalent to the PSM node features. This strategy enables our model to distinguish between different types of nodes while retaining the capacity to adapt to diverse datasets.

\subsubsection{Connecting the Dots: Edges in the Tri-partite Graph} 
While nodes represent distinct entities, the true essence of their relationships is captured through edges. Peptides and PSMs are interconnected in a one-to-many relationship, with each PSM associated with a unique peptide-spectra pair. To enable our model to selectively control the flow of information from a specific PSM node to a peptide node, we choose the peptide identification score as the edge weight for each (peptide, PSM) pair, which is defined as $E_{uv}$ for a given source node $u$ and a target node $v$ in the remaining sections. In doing so, a lower identification score signals a weaker presence of a peptide in the sample mixture, thereby indicating that the corresponding PSM node is less reliable for computing the score of its parent protein nodes.

An edge is formed between a protein and a peptide if the peptide appears in the protein. Instead of setting all edge weights to 1, which simply indicates connectivity between peptide and protein nodes, we could incorporate certain prior knowledge into the design. To this end, we integrate a peptide-sharing feature into our graph design by creating a specialized edge attribute $S_{uv}$ between a peptide $u$ and a protein $v$, which is defined as
\begin{align}
    S_{uv} = \left \{
        \begin{aligned}
             & \frac{1}{|C|}, \  \text{if} \ v={\arg max}_k {f(k)} \ \forall k \in C, \text{where} \ C=\{k| A_{uk}=1\} \\
             & 0, \  \text{if} \ v\neq{\arg max}_k {f(k)}
        \end{aligned}
        \right.  
            \label{eq: protein score 1}
\end{align}
Here, $A$ is the bipartite adjacency matrix for peptide and protein, and $A_{uv}=1$ indicates a connection between peptide $u$ and protein $v$. $C$ is the node index set of the protein nodes that connect to the peptide $u$, and $|C|$ represents the size of the set $C$. $f(\cdot)$, a score indicating if a protein is connected to many high-scoring peptide, is defined as
\begin{align}
    f(k) = \sum_{l \in D}(\mathbbm{1}_{d_l > \epsilon}), \text{where} \ D=\{l | A_{lk} = 1\}
    \label{eq: protein score 2}
\end{align}
where $d_l$ is the maximum peptide identification scores of peptide $l$ (i.e., each score associates with one PSM), and $\epsilon \in [0, 1]$ is a hyper-parameter. $\mathbbm{1}_{d_l > \epsilon}$ is an indicator function, which is 1 when $d_l > \epsilon$, and 0 otherwise. $D$ is the node index set of the peptide nodes that connect to the protein $k$.
The peptide-sharing feature discounts a peptide's relevance to its parental proteins if it is connected to multiple proteins. Elaborating, for a peptide that connects to multiple proteins, a surrogate score (i.e., $f(k)$ in Eq.\eqref{eq: protein score 2}) is ascertained for each protein, gauged by the count of top-scoring peptides tethered to it. Subsequent to this, solely the edge weight for the protein with the highest surrogate score is retained, with all others being nullified. The penultimate step involves diminishing the edge weight of this top-scoring protein by a factor of $\frac{1}{|C|}$ as laid out in Eq.\eqref{eq: protein score 1}. The intuition here is that we aim to further mitigate the contribution of a peptide to its parental protein if it is not unique evidence of the protein. In practice, the threshold $\epsilon$ is set at 0.9, a decision stemming from our intention to prioritize peptides with high confidence. Note that in this feature design, although the edge weight between a peptide and a weakly connected protein is set to 0, it is only used as a feature in the subsequent model, and the edge still exists. 
% An example of the tripartite graph is shown in Figure~\ref{fig: protein example}.
A schematic depiction of the tri-partite graph is available in Figure~\ref{fig: schema}, and an example of the tripartite graph containing the related information of one protein is presented in Figure~\ref{fig: graph example}.

\subsection{GNN Model Architecture} 
\label{sec:gnn_arch}

Using the tripartite graph as input, we can naturally formulate the protein score generation as a protein node prediction task, utilizing GNNs as the foundational model architecture. Our design draws inspiration from GraphSAGE, a network that elevates message-passing operations through a customized aggregation function. This offers greater flexibility compared to GCNs (Graph Convolutional Neural Networks~\citep{kipf2016semi}), which rely solely on a linear transformation of node features and a subsequent mean aggregation from neighboring nodes. Notably, our model augments GraphSAGE by addressing the inherent heterogeneity of the tripartite graph. This refined architecture is underpinned by two core principles: 1) distinct edge types propagate messages uniquely. 2) different node types update their hidden representations differently.

\begin{figure*}[t!]
\centering
    %\centering
    % \includesvg[height=2.9cm]{figures/wadi_graph_deviation.svg}
\includegraphics[width=5.5in]{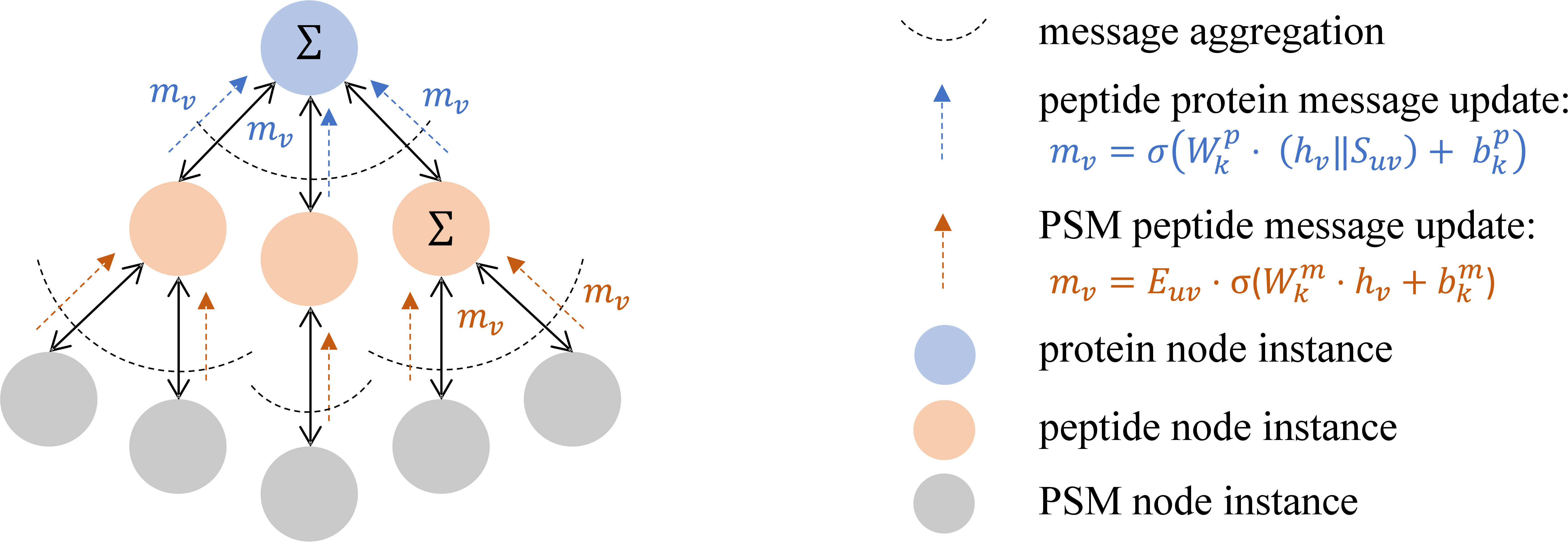}  
       % \vspace{-0.55cm}
    %\caption{}
\caption{An illustrative example of the message passing operation of GNN, where the message update function of each edge type is processed differently.}
    \label{fig: gnn message passing}
\end{figure*}

Formally, given the tri-partite graph $\G=(\E, \V)$, where the set of nodes $\V$ is partitioned into three disjoint subsets: $\V_{\text{pep}}$, $\V_{\text{pro}}$, and $\V_{\text{psm}}$, representing the nodes for peptides, proteins, and PSMs, respectively.
We define the message passing operation between a source node $v \in \V$ and a target node $u \in \V$ as
\begin{align}
a_u^{(k)} &= \sum_{v \in \N_u} m_v^{(k-1)}\\
h_u^{(k)} &= \sigma(W_k \cdot (h_u^{k-1} \mathbin\Vert a_u^{(k)}) + b_k)
\label{eq. message passing}
\end{align}
where $h_u^{k}$ denotes the output feature representation of node $u$ at the $k$-th GNN layer ($h_u^0 = x_u$, the feature of node $u$), $\N_u$ represents the set of neighboring nodes of $u$, $W_k$ and $b_k$ are the learnable weights and biases, $\sigma$ is a ReLU function, and $\mathbin\Vert$ denotes a vector concatenation. The message $m_v$ propagating from node $v$ to $u$ is uniquely defined for different edge pairs. For node pairs comprising peptides and proteins, i.e., $(u \in \V_{pro}, v \in \V_{pep})$ or vice versa, the message $m_v$ at the $k$-th layer is computed as
\begin{align}
m_v^{(k)} = \sigma(W^{p}_k \cdot (h_v^{(k-1)} \mathbin\Vert S_{uv})+ b^{p}_k)
\label{eq. message function for protein and peptide}
\end{align}
where $S_{uv}$ represents the edge attribute associated with the ($u$, $v$) pair, i.e., the feature that penalizes the peptide having non-unique evidence for the target protein. $W_k^p$ and $b_k^p$ are the weights and biases for the peptide-protein pairs. Conversely, the message between peptides and PSMs, i.e., $(u \in \V_{psm}, v \in \V_{pep})$ or vice versa, is defined as
\begin{align}
m_v^{(k)} = E_{uv} \cdot \sigma(W^{m}_k \cdot h_v^{(k-1)}+ b^{m}_k)
\label{eq. message function for psm and peptide}
\end{align}
where $E_{uv}$ corresponds to the identification score of the source PSM node, $W_k^m$ and $b_k^m$ are the weights and biases for the peptide-PSM pairs. See Figure~\ref{fig: gnn message passing} for a visual illustration of the proposed GNN message passing operation.

It's noteworthy that our model employs a unique set of learnable weights and biases for different edge pair types, capturing the varied distributions across node types. For example, the number of PSM nodes serves as important prior information, as a peptide connected to more PSMs suggests a higher likelihood of occurrence. Similarly, the count of peptides linked to a protein indicates the probability of its target protein's presence. By using specific transformations for these node distributions, our model captures information intrinsic to each node type.

Moreover, we use different message functions for pairs of nodes that have different edge attributes. For the edge attribute between peptides and PSMs, which reflects the probability that a PSM originates from a given peptide, we selectively filter out PSMs that are unlikely to have originated from the peptide directly based on the edge score (shown in Eq.~\eqref{eq. message function for psm and peptide}). On the other hand, for the edge attribute between proteins and peptides, we treat the edge attribute as a feature rather than a filter (shown in Eq.~\eqref{eq. message function for protein and peptide}), as it provides relatively weak prior information. In this way, we let the neural network learn the importance of peptides in relation to a given protein primarily in a data-driven way. 

% \subsection{Model Optimization}
% \subsubsection{Semi-supervised training based on Pseudo-labels}
% We optimize the GNN model architecture through a binary cross-entropy loss on protein nodes, which is defined as
% \begin{align}
%     L = \frac{1}{|\V_{pro}|}\sum_{i\in \V_{pro}} \left[ {\hat{y}_i \log{y_i} + (1-\hat{y}_i) \log {(1-y_i)}} \right]
% \end{align}
% where $\hat{y}_i$ is obtained by passing the GNN output representation $h_i$ of protein $i$ through a learnable one-layer feedforward neural network, i.e., $f_o(\cdot): R^d \rightarrow R$, and $y_i$ corresponds to the true label of protein $i$. That is, the node classification is primarily based on the protein nodes. Since we typically don't have access to the ground truth labels for proteins, we start by using an existing protein algorithm to generate the pseudo-scores for each protein, and then we convert the scores into 1 or 1 based on why it passes the $5\%$ decoy FDR threshold or not. In addition, we intentionally relabel the decoy proteins to negatives, i.e., 0, since this is the information that we already know before the training. 

\subsection{Training}
\label{sec:training}

We train our model under a self-training scheme, which is a classic semi-supervised learning technique that iteratively uses the label generated by the trained model as the training label for the next round of training procedure. This approach hinges on the premise that the efficacy of a new model iteration is closely tied to the quality of the pseudo protein labels from its predecessor.

In a typical self-training setting, a model is initially trained using a small labeled dataset and subsequently applied to a large unlabeled dataset to generate pseudo labels. The same model is then trained on a combination of labeled and pseudo-labeled data, with unlabeled data iteratively added to the training set. However, in our problem setting, we lack labeled data initially, while having access to other protein inference models to generate pseudo labels for the unlabeled data. As a result, we adopt a self-training variation similar to the work~\citep{pang2020self}, in which an existing benchmark model provides the initial labeled dataset.

In this study, our model undergoes training within a binary classification framework. The training loss, binary cross entropy, is mathematically defined as:
\begin{align}
L = \frac{1}{|\V_{pro}|}\sum_{i\in \V_{pro}} \left[ {\hat{y}_i \log{y_i} + (1-\hat{y}_i) \log {(1-y_i)}} \right]
\end{align}
In this formulation, $\hat{y}_i$ represents the predicted label, derived from the GNN output representation $h_i$ of protein $i$ through a trainable one-layer feedforward neural network followed by a sigmoid layer, denoted as $f_o(\cdot): R^d \rightarrow R$. The term $y_i$ corresponds to the actual label of protein $i$, and $|\V_{pro}|$ is the number of proteins used for training. The classification process predominantly focuses on protein nodes.

Given the typical unavailability of ground truth labels for proteins, our approach employs pseudo labels generated by thresholding the scores from a benchmark model. The threshold is determined based on the False Discovery Rate (FDR); specifically, proteins exceeding a defined FDR threshold (e.g., 0.05 in our experiments) are categorized as positive samples, with the rest deemed negative. Additionally, decoy proteins are explicitly labeled as negative, owing to their inherent absence in biological samples. After the initial training round based on the labels provided by the benchmark model, we replace the previous pseudo labels with the new ones generated by our latest model and retrain our deep learning model accordingly in subsequent self-training rounds.

In each round $i$ (including the initial round, where the training is based on labels provided by the selected benchmark model), the self-learning procedure outputs a learned model $\phi_i$ and returns a list of trained models. Analogous to ensemble learning, we perform an average aggregation of all learned models to obtain the optimal protein score. Specifically, the final protein score for a given protein x is computed as follows:
\begin{align}
score(x) = \frac{1}{t}\sum_{i=1}^{t}\phi_i(x)
\end{align}
where $t$ denotes the total number of self-training rounds. In our experiments, we set $t$ to 10 rounds.

We employ self-training on several public datasets. Once trained, the model is directly applied to each test dataset to produce the respective protein scores, eliminating the need for re-training. In contrast to methods demanding distinct training or fine-tuning for individual datasets, our findings underscore a pronounced consistency in peptide identification data across all examined datasets. Such uniformity allows us to deploy a single, universally adaptable model, mitigating the problem of overfitting. Moreover, this approach increases computational efficiency by reducing the need for redundant training procedures for different datasets.

\subsection{Implementation}

\subsubsection{Experimental Setting}
In our experiments, we select Epifany \cite{pfeuffer2020epifany} as the base model to generate the initial pseudo labels due to its relative computational efficiency compared to other models and competing performance, as demonstrated in their original paper.

We adopted the Adam optimizer with a learning rate of 0.001. 
The model is composed of six GNN layers, with node and edge hidden dimensions set to 100. In addition, our model is trained on a single Nvidia RTX4090 graphics card over 1000 epochs, and the parameters leading to the best validation loss are stored. 

The software versions and configurations in our experiments are all listed in Table S3.

\subsubsection{Trainging Datasets} 
\label{sec: training datasets}
The training datasets are public protein datasets downloaded from ProteomeXchange. To avoid over-fitting to a specific benchmark dataset used in our experiment, those datasets are selected randomly, and processed with Comet~\citep{eng2015deeper} search and Percolator~\citep{kall2007semi} for peptide database search, with respect to their experimental specifications. For convenience, we selected only human data, but this does not affect our generalizability, as shown in Figure S5.

The following are the datasets used for training: PXD004789~\citep{kreutz2017response}, PXD005388~\citep{meier2017organoruthenium}, PXD006640~\citep{venkatraman2017proteomic}, PXD010319~\citep{meier2017organoruthenium}, PXD022881~\citep{bhat2021combined}, 
PXD023034~\citep{Zawily2023}, PXD023593~\citep{lupton2021cryo}, PXD025701~\citep{mumtaz2022secreted}, PXD026991~\citep{fischer2022supt3h}, 
PXD030330~\citep{Jaffray2023}, PXD030448~\citep{hotta2022eml2}, PXD032035~\citep{liu2022sars}, PXD032284~\citep{lo2023development}, 
PXD034012~\citep{cela2022proteomic},  PXD035125~\citep{Herman2022}, PXD036171~\citep{su2023study}, and
PXD039272~\citep{hsu2020trop2}, the links and search parameters of which are listed in Table S2.

\subsubsection{Test Datasets}
We evaluated our algorithm on the following MS/MS datasets: iPRG2016~\citep{iprg2016}, UPS2~\citep{ups2}, 18Mix~\citep{18mix}, Yeast~\citep{ramakrishnan2009gold} and Hela-3T3~\citep{saltzman2018gpgrouper}. All datasets except Hela-3T3 provides us with ground truth labels for evaluating entrapment FDR, while for Hela-3T3 we evaluate based on two-species FDR. Among these datasets, iPRG2016 was specifically curated to test protein inference algorithms on proteins that share peptides. 18Mix and Yeast also contain a small portion of peptide-sharing proteins. Hela-3T3 provide us with human hela or mouse 3t3 cells, with large amount of shared peptides. Table~\ref{table:test_data_summary} offers the summary statistics of these datasets, while a detailed description of these datasets are provided in the Material S1.

\begin{table}[H]
    \centering
    \begin{tabular}{| c|c|c|c| }
    \hline
    \textbf{Dataset} & \textbf{\# True Proteins} & \textbf{\# Contaminate Proteins}  & \textbf{\# Identified}\\
    \hline
    iPRG2016 A & 191 & 1,191 & 179 \\
    \hline
    iPRG2016 B & 191 & 1,191 & 187 \\
    \hline
    iPRG2016 AB & 382 & 1000 & 368 \\
    \hline
        Yeast & 4,265 & 6,330 & 551 \\
    \hline
        UPS2 & 48 & 48 & 23 \\
    \hline
        18Mix & 18 & 1,802 & 13 \\
    \hline
        Hela & 20,419 & 17,202 & 1335 \\
    \hline 
        3T3 & 17,202 & 20,419 & 799 \\
    \hline
    \end{tabular}
    \caption{Number of true proteins and contaminate proteins of each test dataset, along with numbers of protein identified at 5\% FDR by GraphPI.}
    \label{table:test_data_summary}
\end{table}

To process the tandem MS data, we first converted and centroided the raw files with msConvert \cite{chambers2012msconvert}. For UPS2, we do not need to generate decoy proteins since they are already provided by the fasta file. For other datasets, the provided fasta file was used to generate a decoy database through shuffling of amino acid with the OpenMS \cite{rost2016openms} DecoyDatabase tool. Then, spectra were searched using Comet allowing 10 ppm precursor mass tolerance, 0.01 Da fragment mass tolerance, and one missed cleavage for fully tryptic peptides (for 18Mix, the precursor and fragment mass tolerance is set to 1.005 Da since it comes from a low resolution instrument). Carbamidomethylation(C) is selected as the fixed modification except Yeast (which does not undergo alkylation), and oxidation(M) the variable one. Then, we extracted additional features from the Comet search, and feed them into Percolator to obtain better peptide scores.

\subsubsection{Benchmark Methods}
We used the following four popular and representative methods to compare against our model: Epifany \citep{pfeuffer2020epifany}, Fido \citep{serang2010efficient}, PIA \citep{uszkoreit2015pia}, and DeepPep \citep{kim2017deeppep}. They are either based on parsimony, probability, Bayesian, or
deep learning approaches, covering the majority of the approaches to this problem. 

Leveraging PSM features from Percolator, each method derives a probability score for individual proteins. The probability output of each model is used to rank the proteins. Groups of identically connected proteins are treated as one single protein group during inference. When we are referring to the number of proteins, such groups contribute only one per group, instead of contributing once per protein.

%% file: chapters/Experiments_revision.tex
\section{Results and Discussion}

\subsection{Comparison Study}

\begin{figure}[t!]
%\captionsetup[subfigure]%{justification=Centering}
\centering
\begin{subfigure}[t]{0.49\textwidth}
\centering
    \includegraphics[width=3.3in]{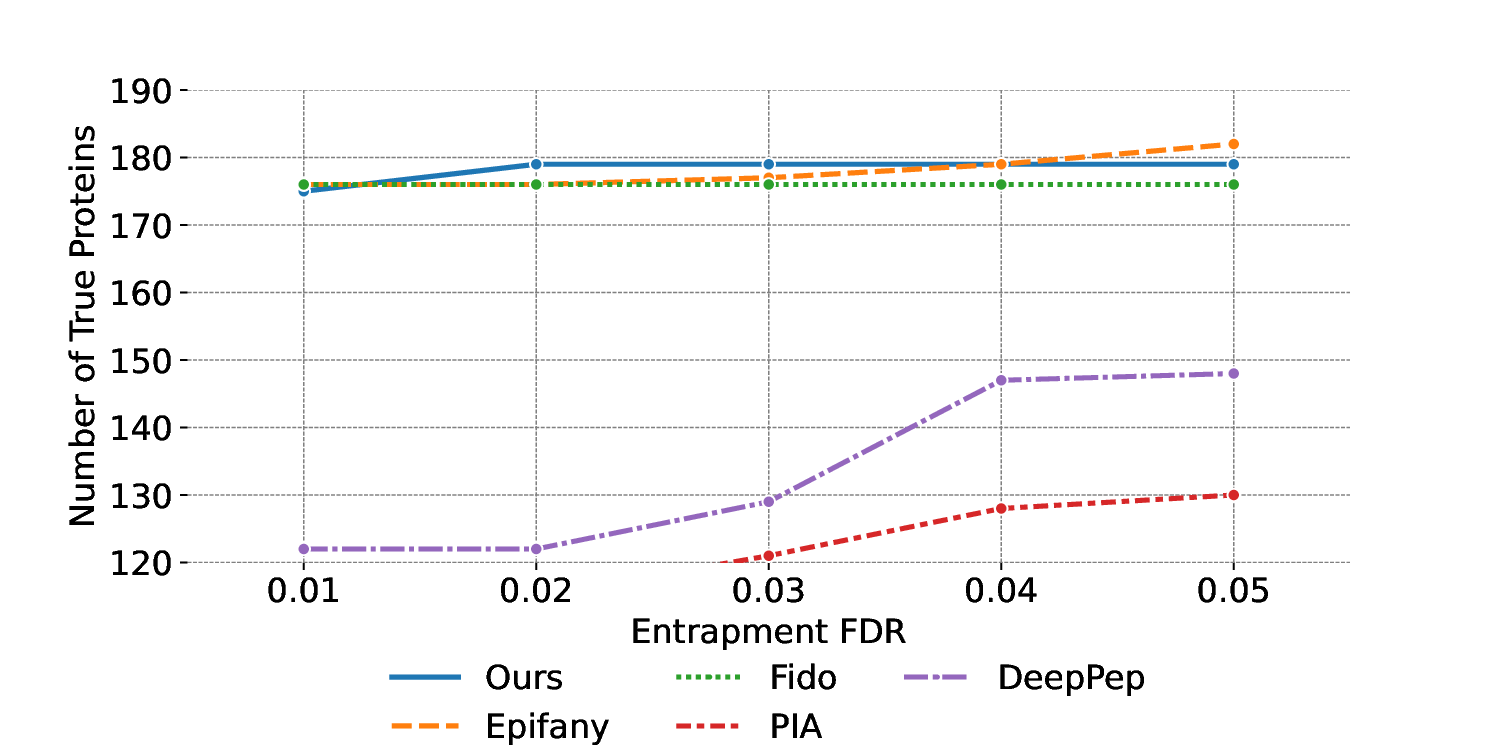}
    \caption{iPRG2016 A}
    \label{fig: iprg_a_roc}
\end{subfigure}\hspace{\fill} % maximize horizontal separation
\hfill
\begin{subfigure}[t]{0.49\textwidth}
\centering
    \includegraphics[width=3.3in]{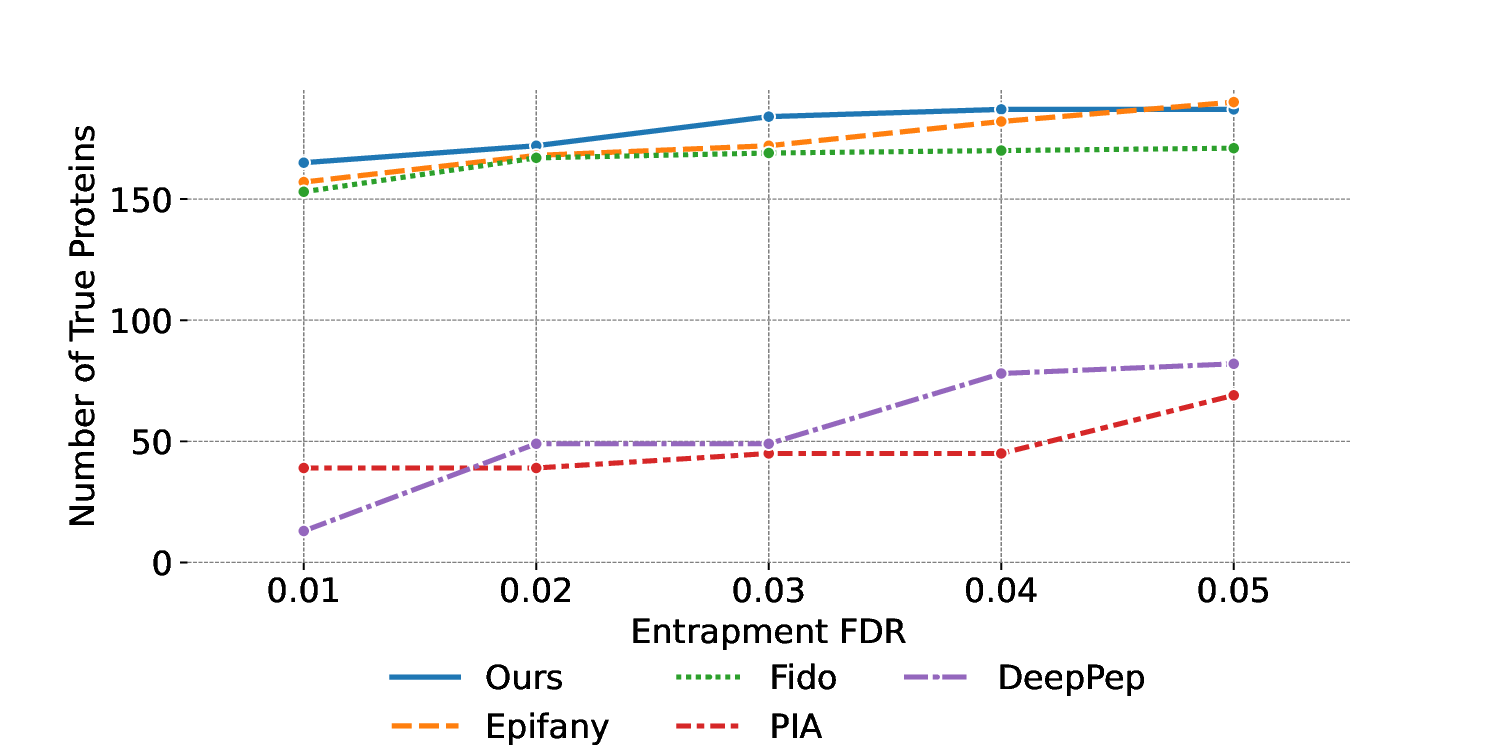}
    \caption{iPRG2016 B}
    \label{fig: iprg_b_roc}
\end{subfigure}

 % Add a line break after the second image

\begin{subfigure}[t]{0.49\textwidth}
\centering
    \includegraphics[width=3.3in]{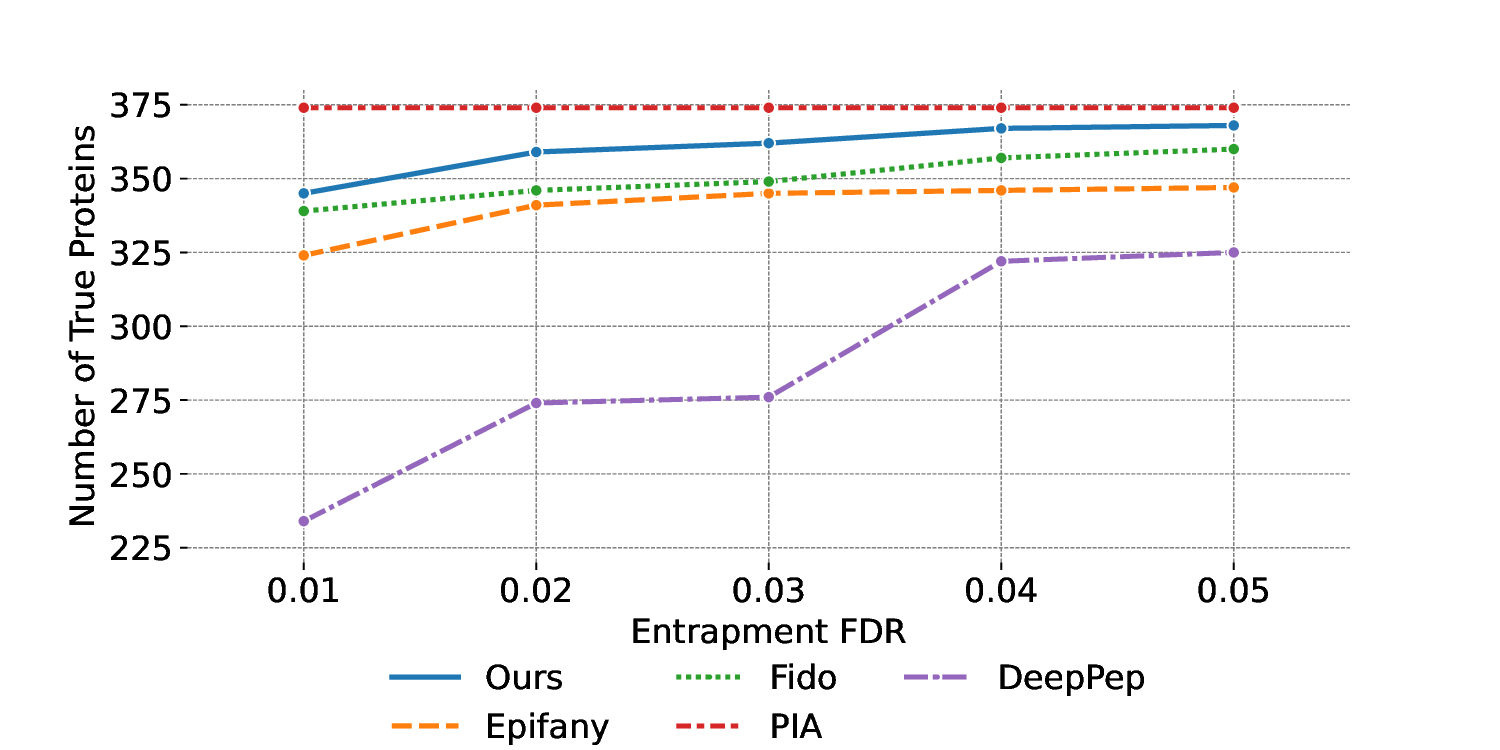}
    \caption{iPRG2016 AB}
    \label{fig: iprg_ab_roc}
\end{subfigure}\hspace{\fill} % maximize horizontal separation
\hfill
\begin{subfigure}[t]{0.49\textwidth}
\centering
    \includegraphics[width=3.3in]{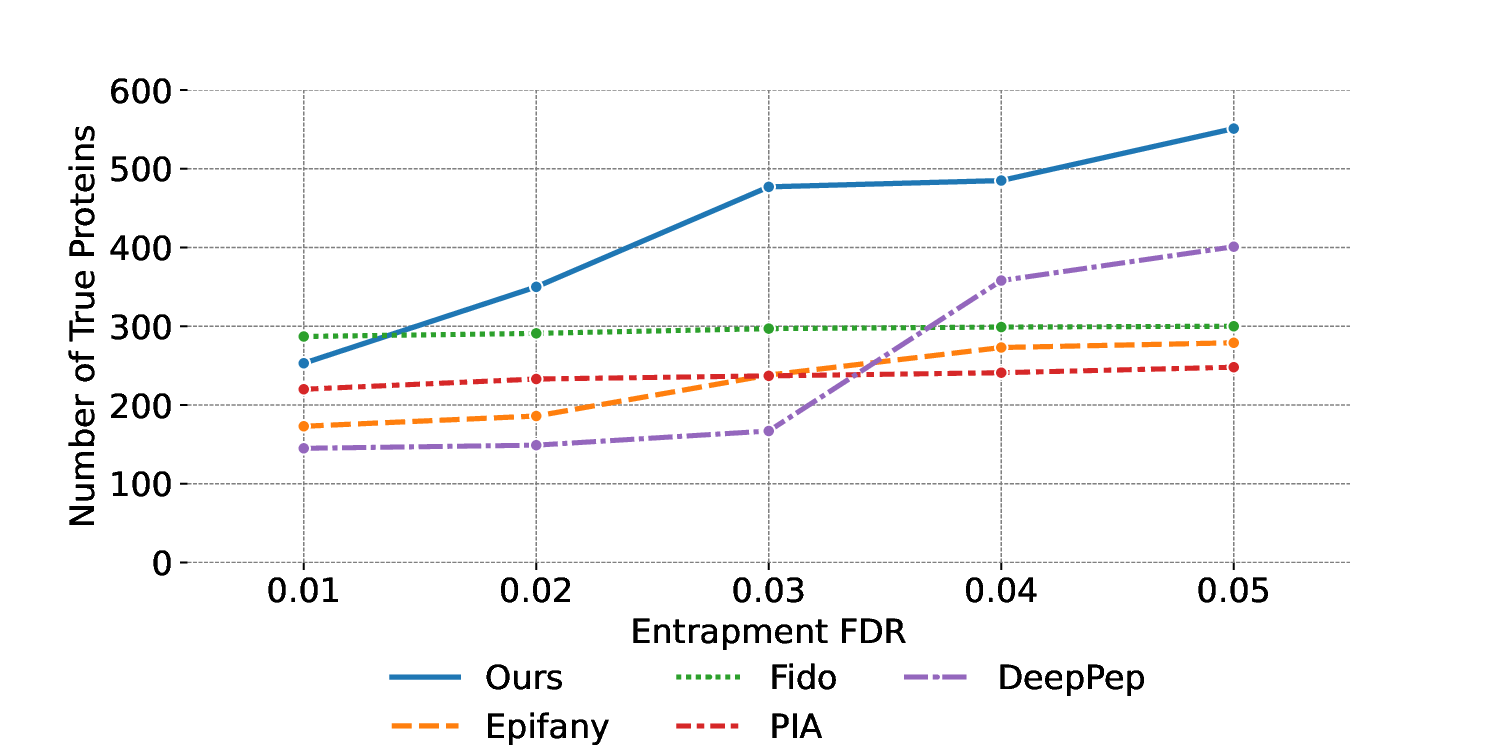}
    \caption{Yeast}
    \label{fig: yeast_roc}
\end{subfigure}

%\newline % Add a line break after the fourth image

\begin{subfigure}[t]{0.49\textwidth}
\centering
    \includegraphics[width=3.3in]{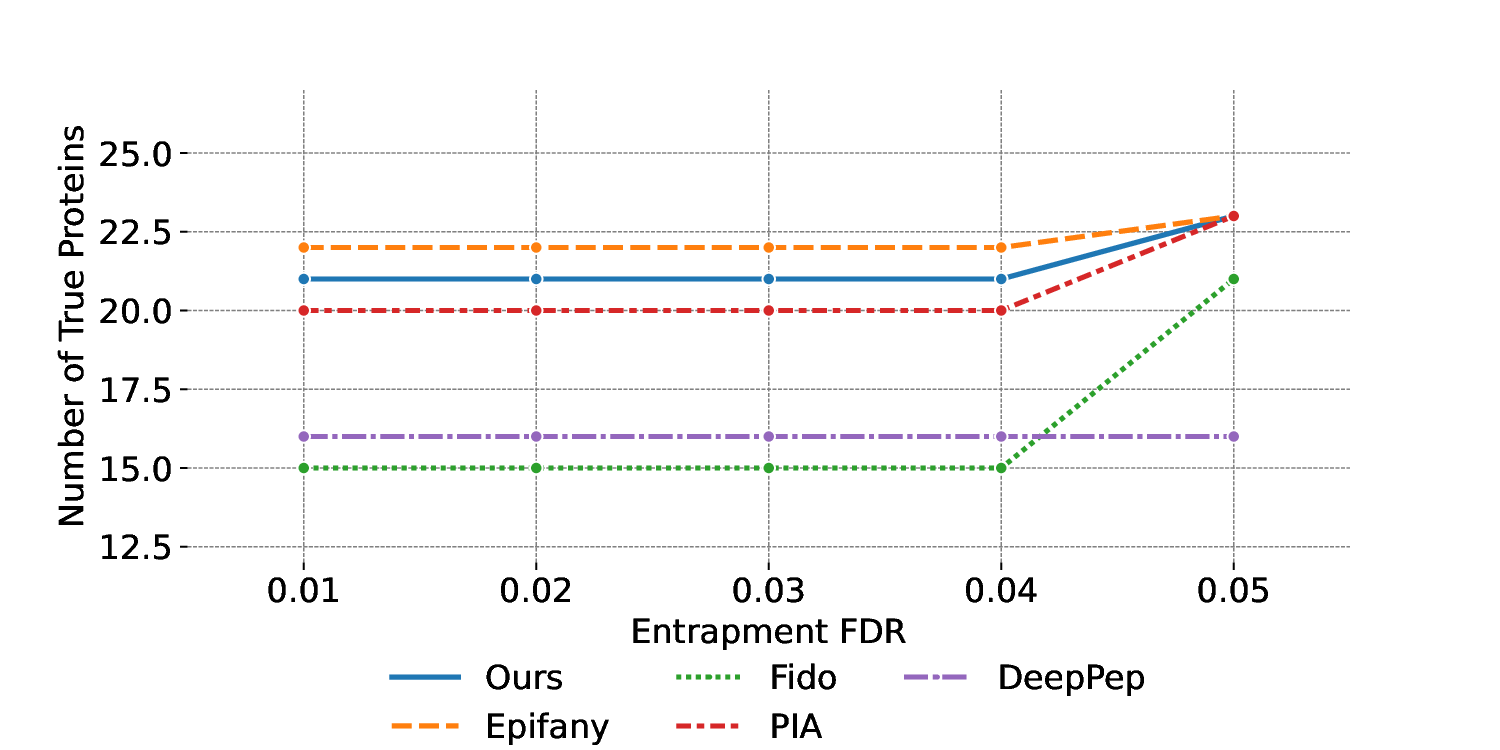}
    \caption{UPS2}
    \label{fig: ups2_roc}
\end{subfigure}\hspace{\fill} % maximize horizontal separation
\hfill
\begin{subfigure}[t]{0.49\textwidth}
\centering
    \includegraphics[width=3.3in]{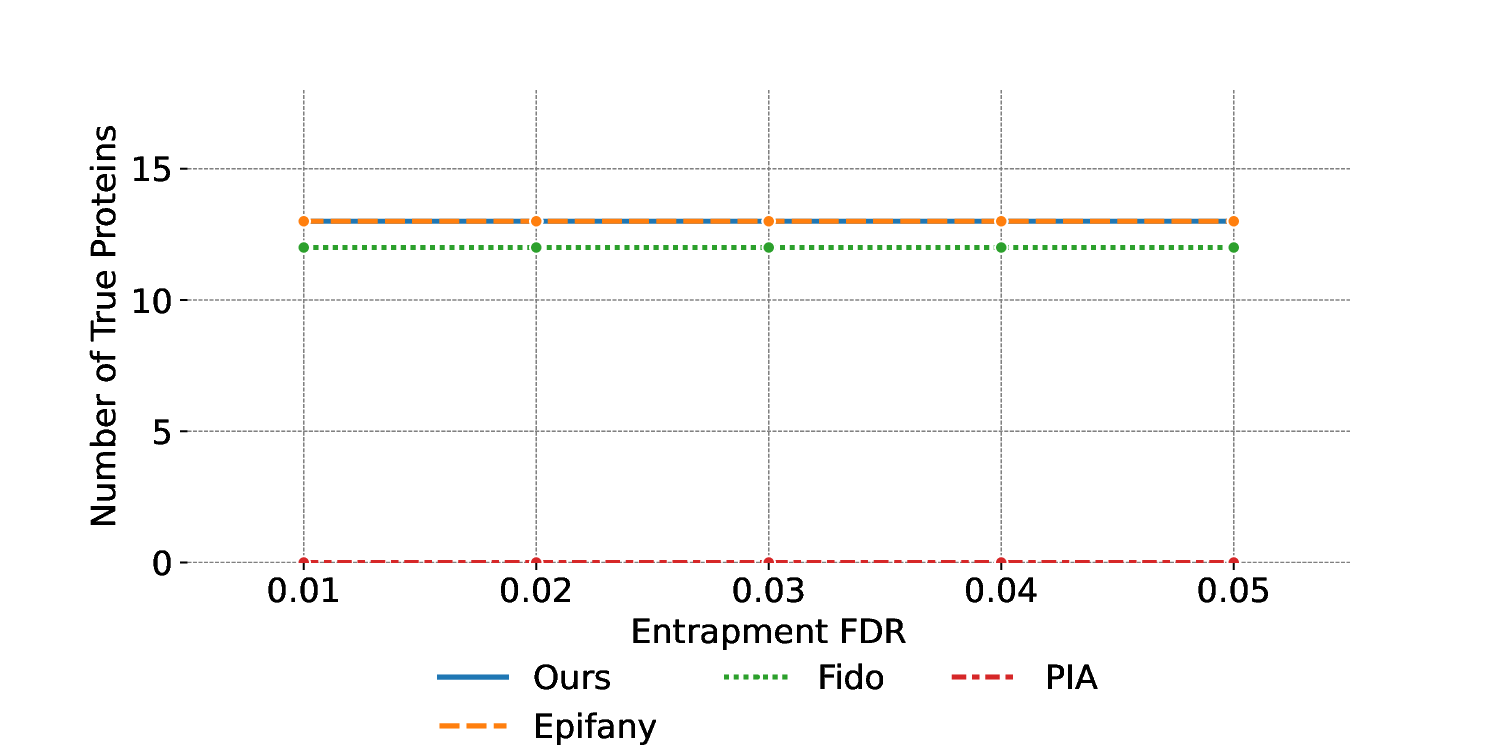}
    \caption{18Mix}
    \label{fig: 18mix_roc}
\end{subfigure}

\begin{subfigure}[t]{0.49\textwidth}
\centering
    \includegraphics[width=3.3in]{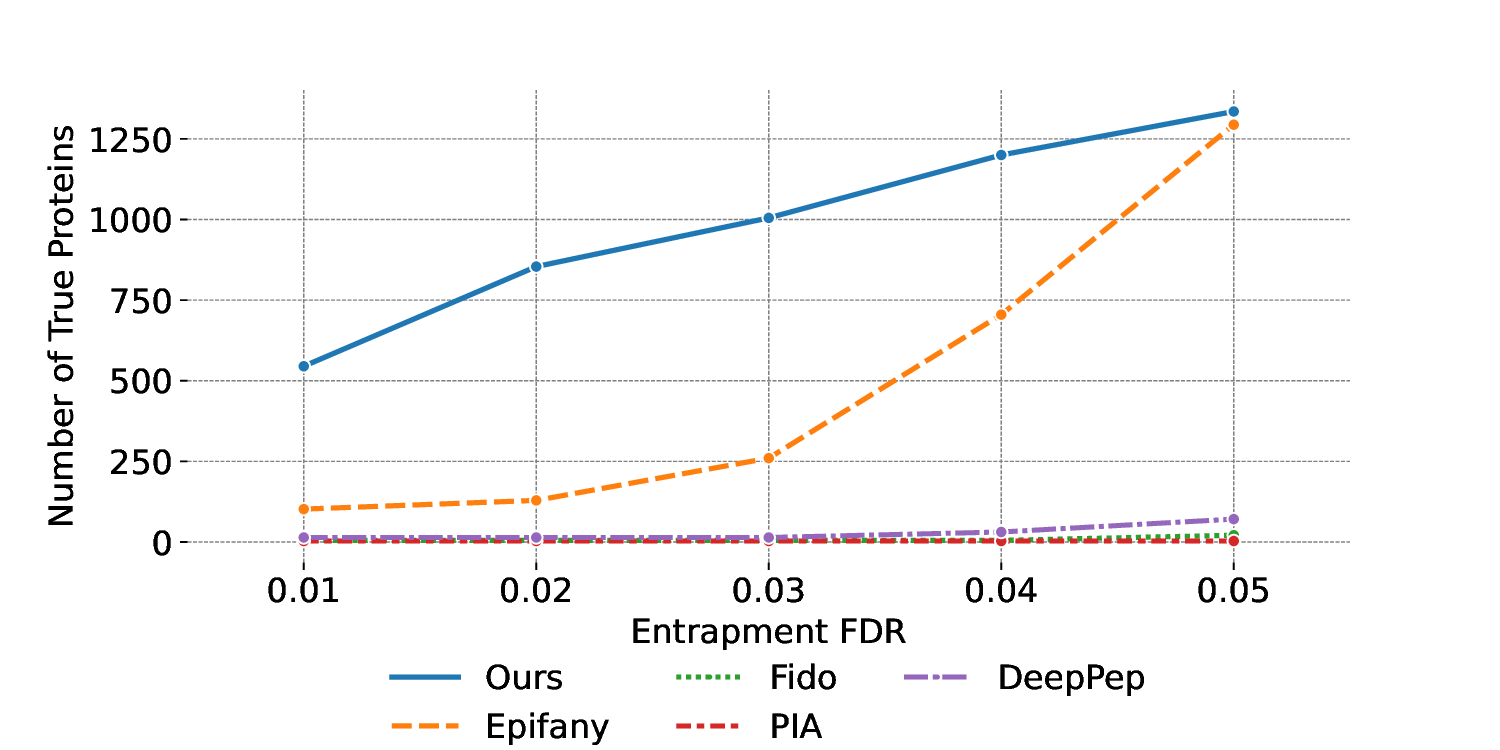}
    \caption{Hela}
    \label{fig: hela_roc}
\end{subfigure}\hspace{\fill} % maximize horizontal separation
\hfill
\begin{subfigure}[t]{0.49\textwidth}
\centering
    \includegraphics[width=3.3in]{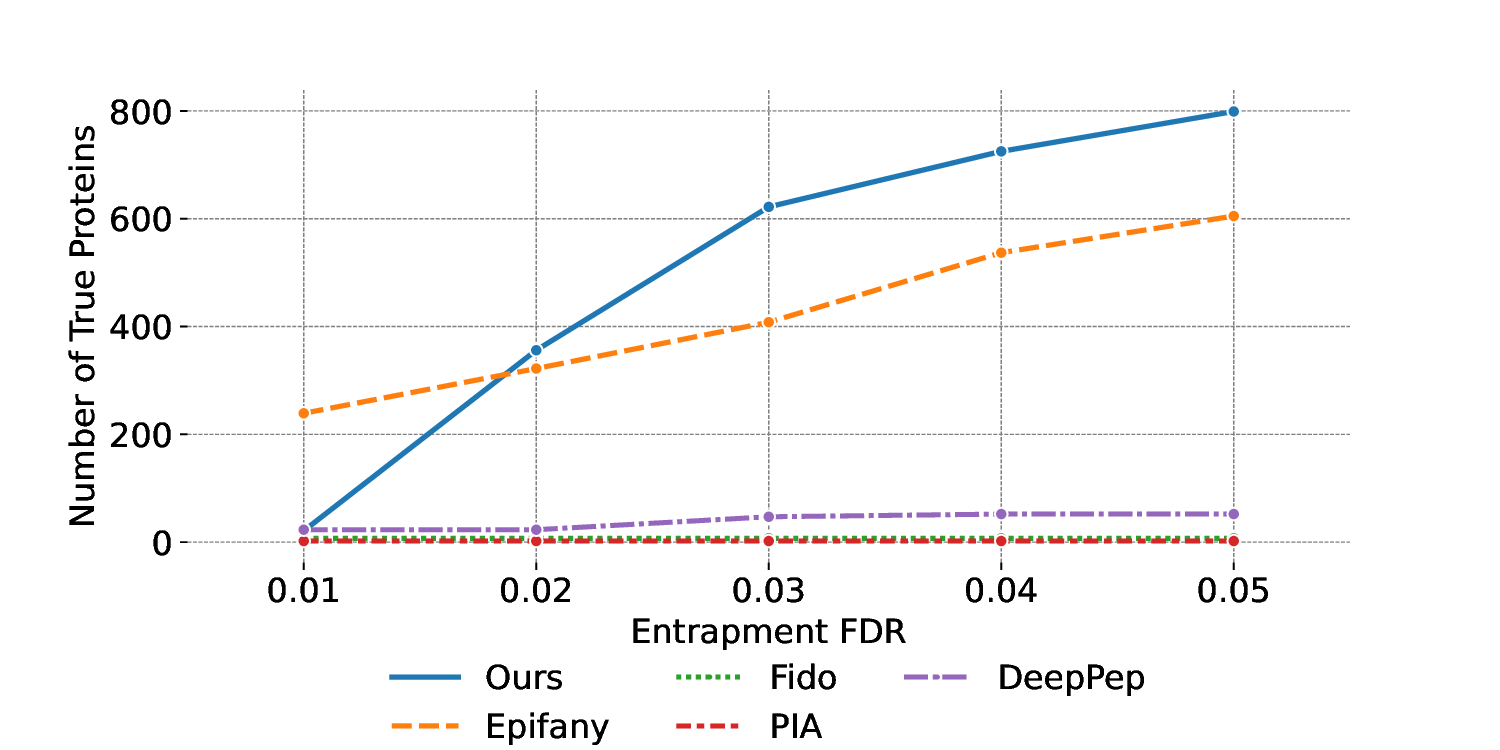}
    \caption{3T3}
    \label{fig: 3t3_roc}
\end{subfigure}

\caption{ROC curve (entrapment FDR vs. number of true proteins) of various models on the benchmark datasets: (a) iPRG2016 A, (b)
iPRG2016 B, (c) iPRG2016 AB, (d) Yeast, (e) UPS2, (f) 18Mix, (g) Hela, and (h) 3T3.}
\label{fig: comparison1}
\end{figure}

The performances of all methods are evaluated by the receiver operator characteristic (ROC) curve, which plots the number of true positive proteins (i.e. the number of ground truth proteins) as a function of entrapment FDR (the portion of contaminate proteins in the identified list, where ``contaminate'' proteins refers to real proteins in the database, but known not to exist in the biological sample). The curve is plotted by varying the FDR threshold above which a protein will appear in the identified list. Given that the test datasets all have ground truth attached (except Hela-3T3, where we implement a two-species approach), we can evaluate based on empirical FDR instead of decoy FDR. Since we are mostly interested in the performance of the methods when the FDR is small, we plot the curve within the entrapment FDR range of $[0.01, 0.05]$. The results are shown in Figure~\ref{fig: comparison1}.

\begin{figure}[t!]
%\captionsetup[subfigure]%{justification=Centering}
\centering
\begin{subfigure}[t]{0.49\textwidth}
\centering
    \includegraphics[width=2.7in]{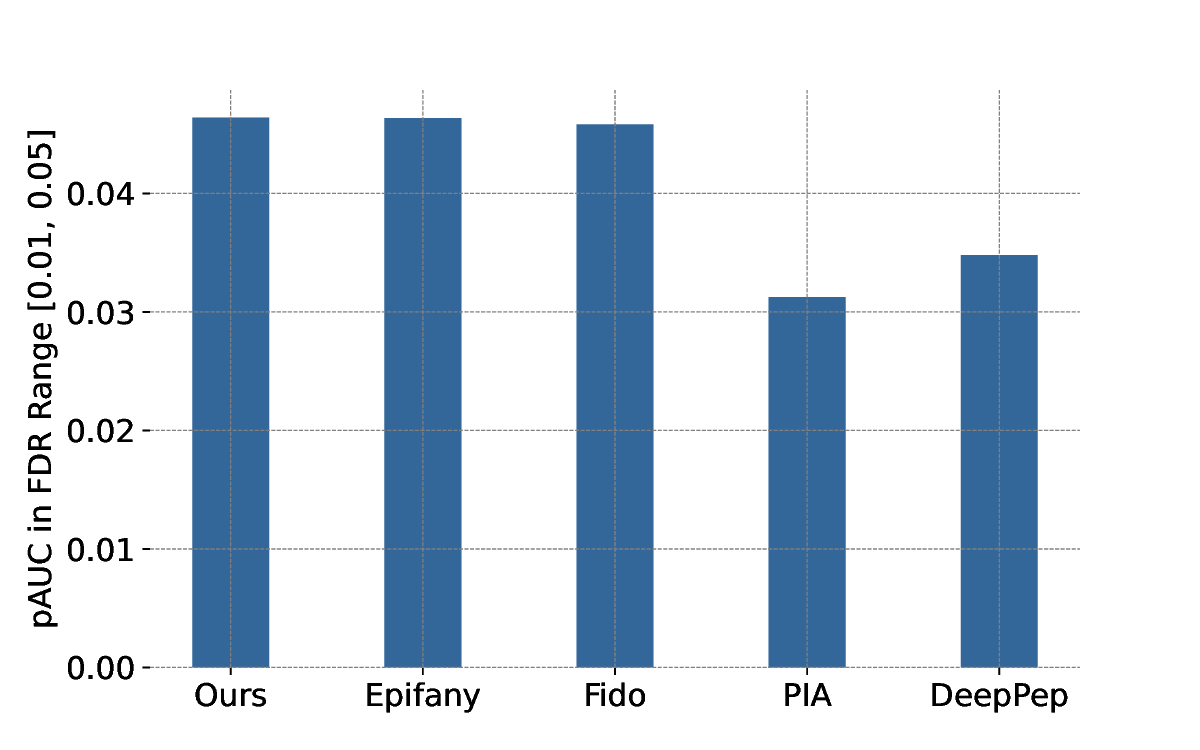}
    \caption{iPRG2016 A}
    \label{fig: iprg_a_pauc}
\end{subfigure}\hspace{\fill} % maximize horizontal separation
\hfill
\begin{subfigure}[t]{0.49\textwidth}
\centering
    \includegraphics[width=2.7in]{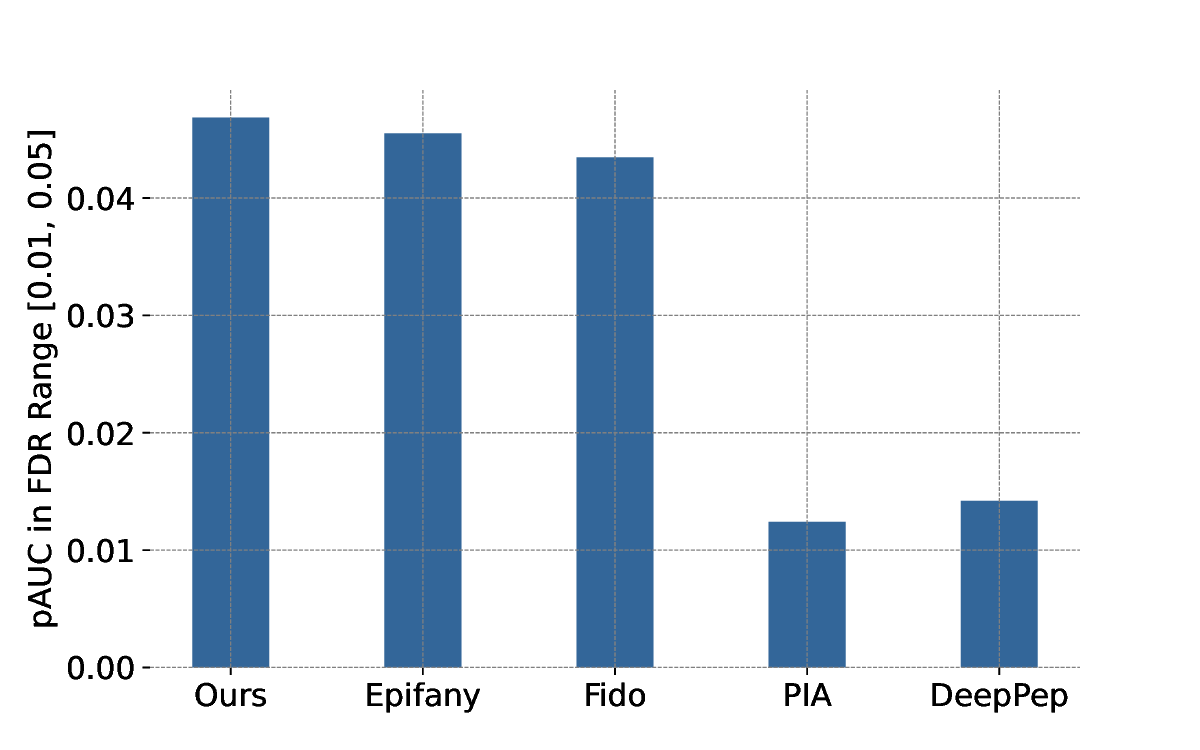}
    \caption{iPRG2016 B}
    \label{fig: iprg_b_pauc}
\end{subfigure}

 % Add a line break after the second image

\begin{subfigure}[t]{0.49\textwidth}
\centering
    \includegraphics[width=2.7in]{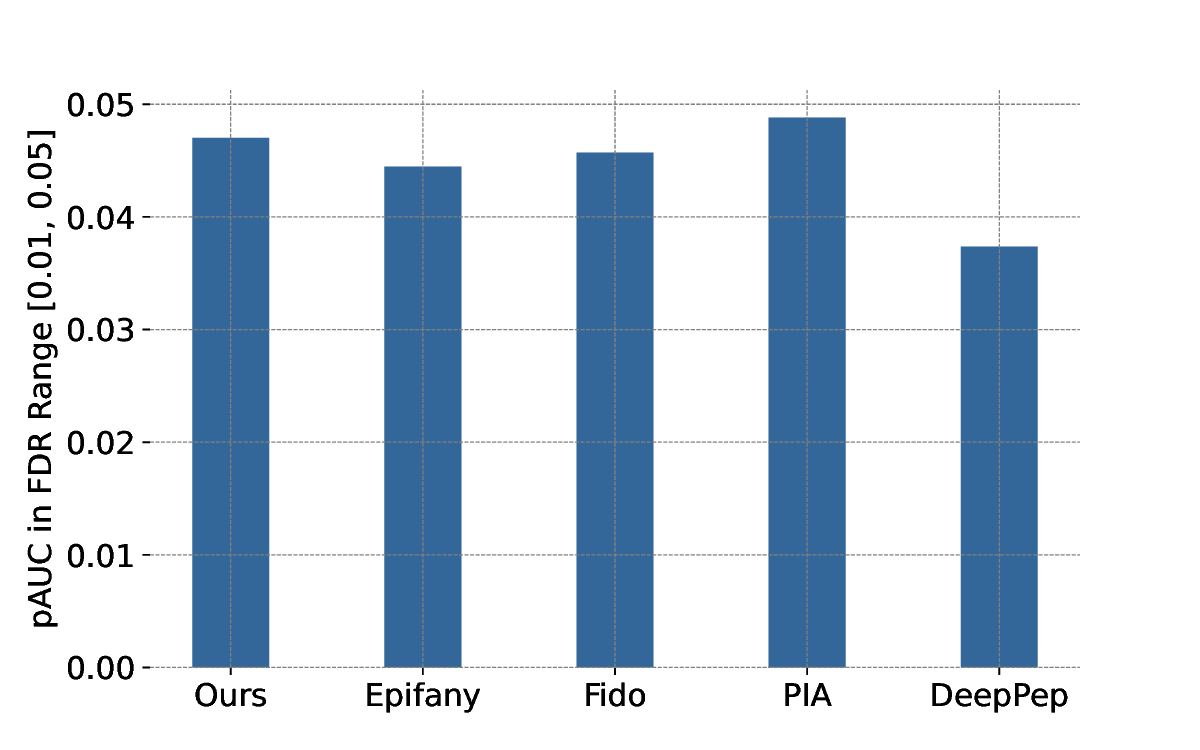}
    \caption{iPRG2016 AB}
    \label{fig: iprg_ab_pauc}
\end{subfigure}\hspace{\fill} % maximize horizontal separation
\hfill
\begin{subfigure}[t]{0.49\textwidth}
\centering
    \includegraphics[width=2.7in]{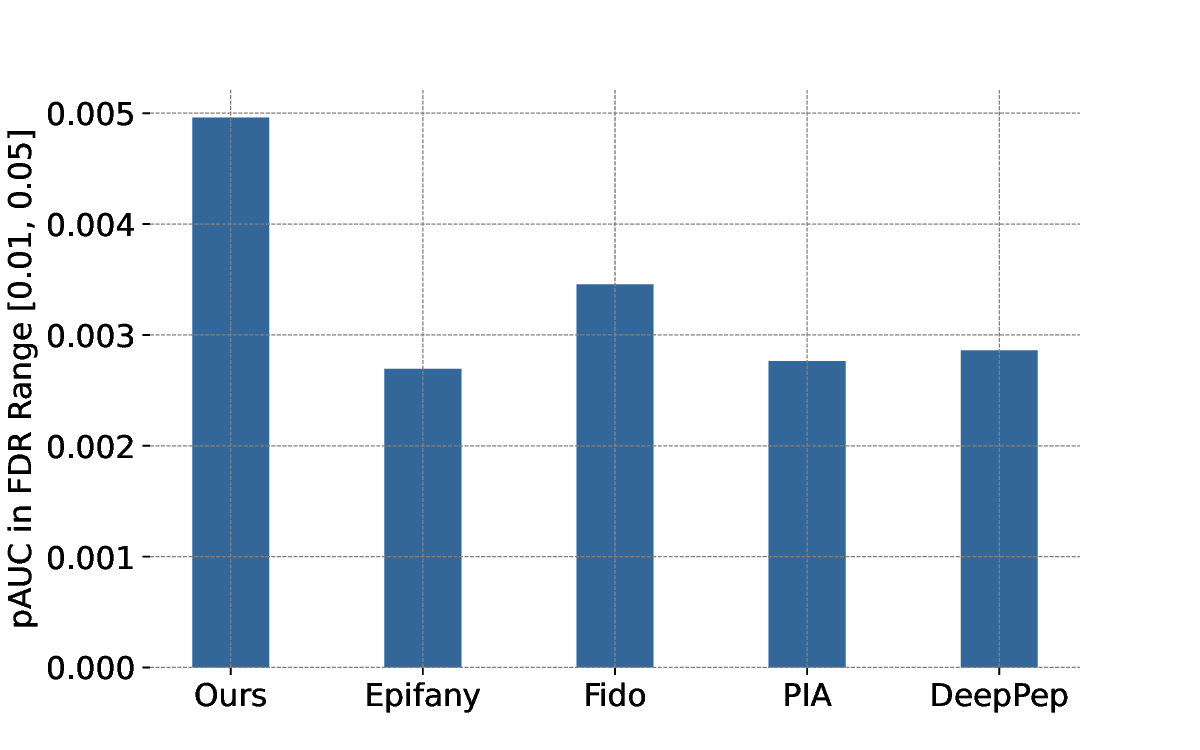}
    \caption{Yeast}
    \label{fig: yeast_pauc}
\end{subfigure}

%\newline % Add a line break after the fourth image

\begin{subfigure}[t]{0.49\textwidth}
\centering
    \includegraphics[width=2.7in]{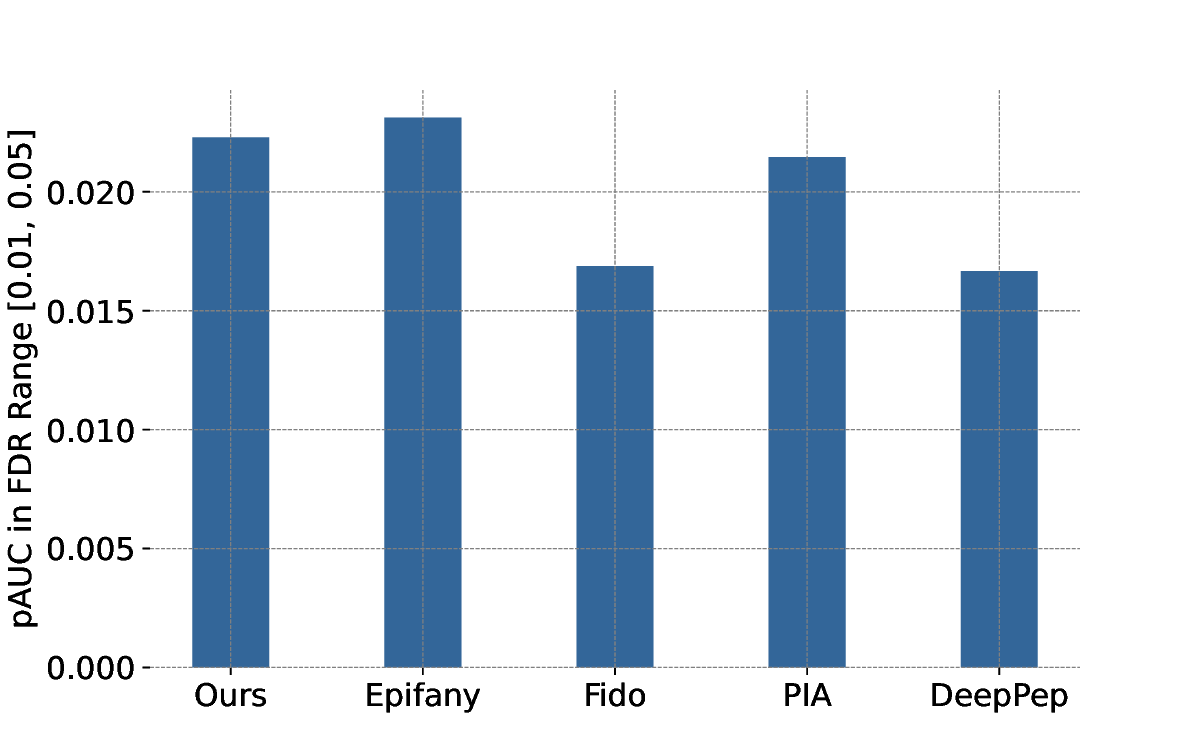}
    \caption{UPS2}
    \label{fig: ups2_pauc}
\end{subfigure}\hspace{\fill} % maximize horizontal separation
\hfill
\begin{subfigure}[t]{0.49\textwidth}
\centering
    \includegraphics[width=2.7in]{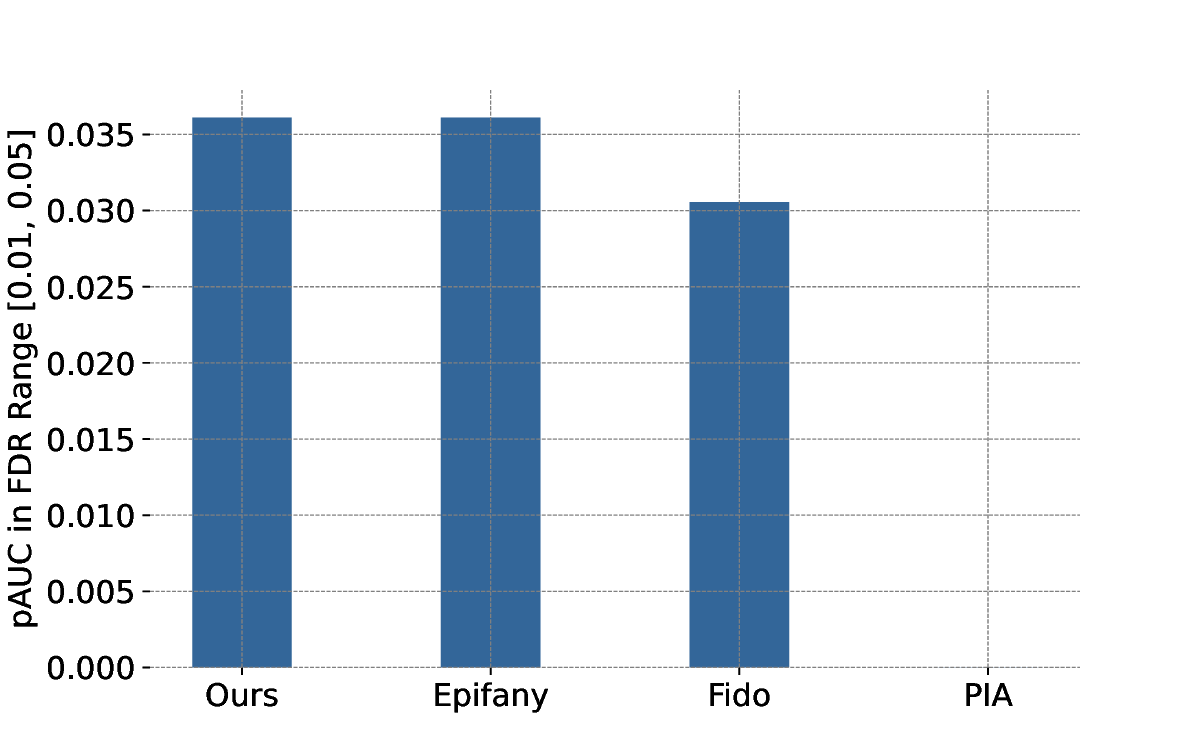}
    \caption{18Mix}
    \label{fig: 18mix_pauc}
\end{subfigure}

\begin{subfigure}[t]{0.49\textwidth}
\centering
    \includegraphics[width=2.7in]{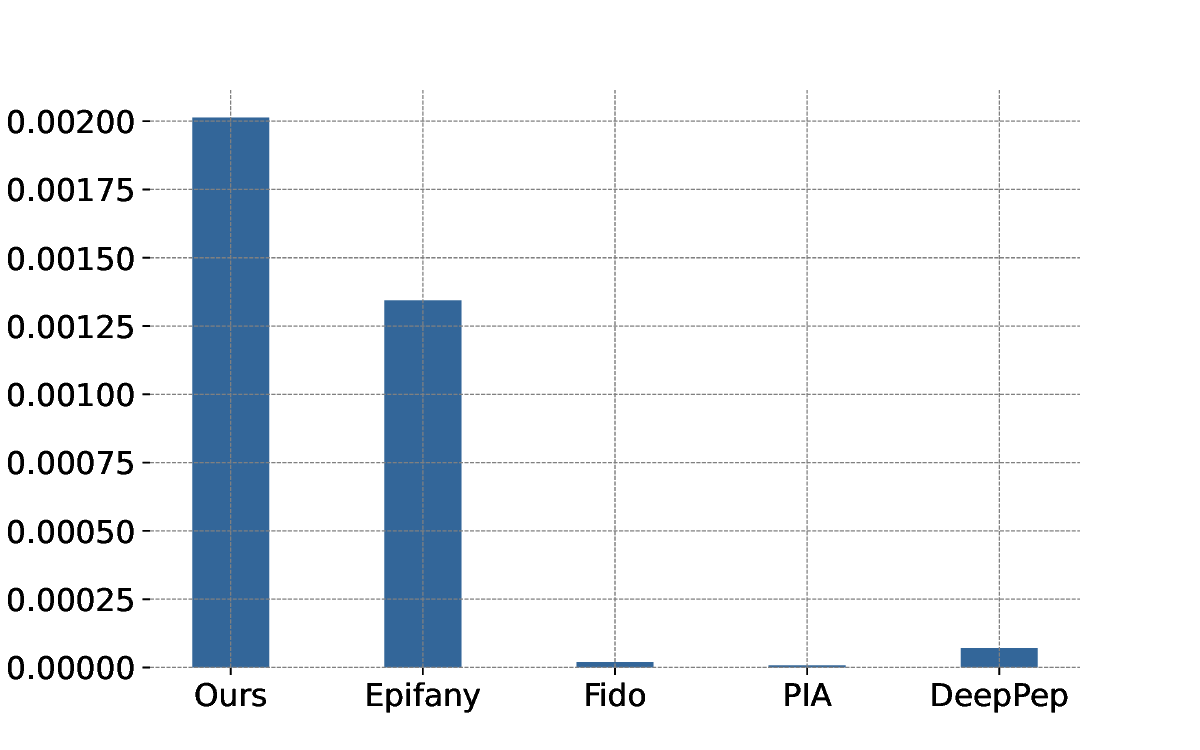}
    \caption{Hela}
    \label{fig: hela_pauc}
\end{subfigure}\hspace{\fill} % maximize horizontal separation
\hfill
\begin{subfigure}[t]{0.49\textwidth}
\centering
    \includegraphics[width=2.7in]{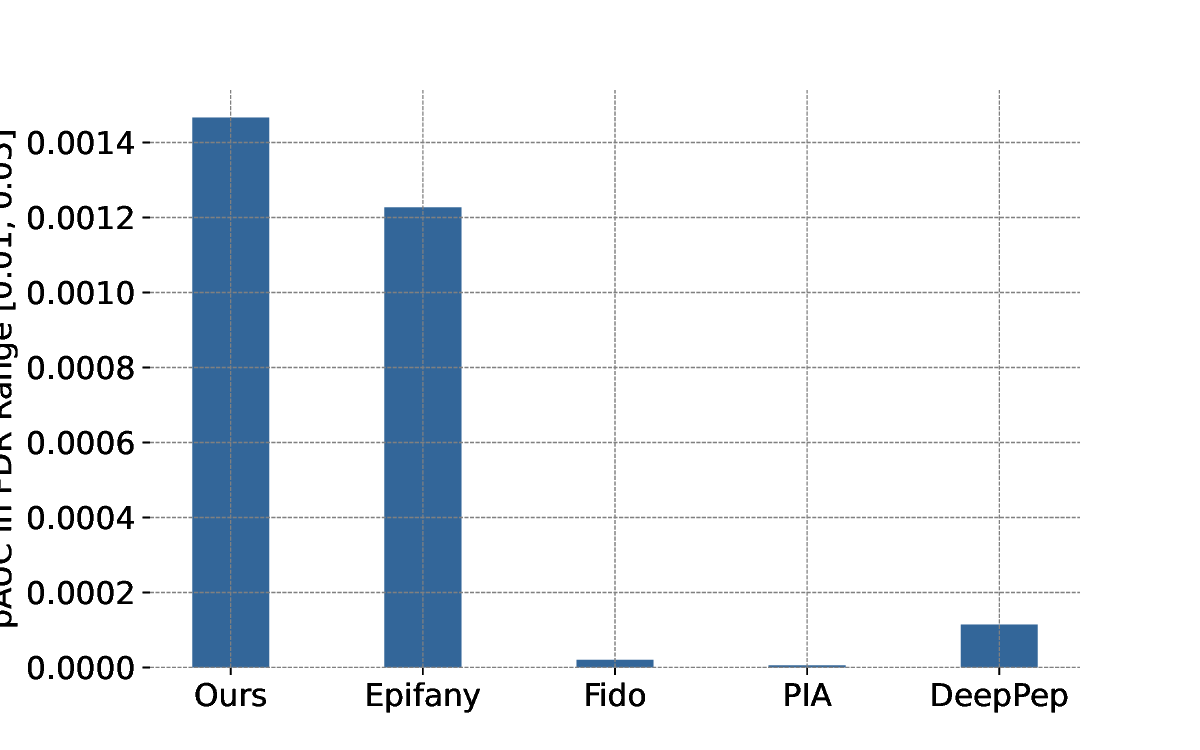}
    \caption{3T3}
    \label{fig: 3t3_pauc}
\end{subfigure}

% \caption{pAUC (partial AUC) score of various models on the benchmark datasets.}
\caption{pAUC (partial AUC) score of various models on the benchmark datasets: (a) iPRG2016 A, (b)
iPRG2016 B, (c) iPRG2016 AB, (d) Yeast, (e) UPS2, (f) 18Mix, (g) Hela, and (h) 3T3.}
\label{fig: comparison2}

\end{figure}

Overall, our method shows competing performance across all datasets. For both the iPRG2016 A and B datasets, GraphPI is the only one that can compete with Epifany in terms of identification performance. In addition, our ROC curves ascend faster than any other method in these two datasets, showing improved performance in the low FDR range. In iPRG2016 AB, GraphPI ranks the second throughout the whole FDR range, only falling behind PIA, which performs extraordinarily well in this dataset. In the Yeast dataset, our method again shows overall best performance. In the UPS2 dataset, GraphPI acquires identification performance similar to that of Epifany, surpassing Fido, PIA, and DeepPep. With the 18Mix dataset, we rank the top along with Epifany. 
Additionally, we tested our model on the Hela-3T3 \cite{saltzman2018gpgrouper} dataset, where we take the samples of 100\% Hela cells and 100\% 3T3 cells, and evaluate the performance based on a two-species library strategy. Our model achieves the top most performance in the comparison, only falling behind Epifany on the 3T3 dataset at 1\% FDR. Moreover, only GraphPI and Epifany can achieve adequate performance in these two datasets,largely due to the high prevalence of shared peptides (see Figure S1 and S2), which adds complexity to the analysis. It is worth noting that, while Epifany can produce state-of-the-art performance in iPRG2016 A and B, it falls short in other datasets like AB and Yeast, but our approach retains competitive and consistent performance across all datasets. 

We also compared the pAUC based on FDR from $1\%$ to $5\%$ of test methods, summarized in Figure~\ref{fig: comparison2}. pAUC computes the area under our ROC curve (between 1\% and 5\% FDR), relative to the perfect curve (a horizontal line with all groundtruth proteins identified at any FDR). A higher pAUC value indicates we identifies more proteins within the FDR range of interest. The results are consistent with our previous analysis, with our method achieving the top performance in iPRG2016 A, B, Yeast, 18Mix, Hela, and 3T3, being second in iPRG2016 AB and UPS2.

% \subsection{Experiments on the Hela-3T3 Dataset}
% We also test the models on the PXD008560 \cite{saltzman2018gpgrouper} dataset. This dataset provides a protein mixture of human hela cells and mouse 3t3 cells, with different ratios. We conducted our protein inference under the database of combined human and mouse proteome.

% For this dataset, we do not have a groundtruth of the hela and 3t3 cell lines, so test under the two species evaluation scheme: first, we take the data consisting of 100\% hela cells and 0\% 3t3 cells, and use the human proteins as true labels, while mouse proteins being contaminates. We conduct the same experiments on data with 0\% hela and 100\% 3t3 cells as well, with mouse proteins being true labels. 

\subsection{Computational Efficiency}
To demonstrate the computational efficiency of GraphPI, we report the runtimes in minutes on the Yeast dataset, which is the largest dataset in our test data. The runtimes are measured on an Intel Core i7 8700K processor. Every algorithm is executed with 12 threads to take advantage of parallelism. GPU acceleration is disabled during inference time to make the comparison fair.

Our method shows a significant advantage when compared to other models, especially Bayesian methods, like Fido or Epifany. On the Yeast dataset, our method takes only 88 seconds to run, while Epifany takes over 14 minutes. Partially this is due to the inherent speed advantage of neural networks over Bayesian networks. On the other hand, Epifany and Fido need to run a grid search of their parameters $\alpha$, $\beta$, and $\gamma$ for every test dataset, while our approach does not need such a procedure, further improving efficiency.

\begin{figure}[t!]
%\captionsetup[subfigure]%{justification=Centering}
\centering
\begin{subfigure}[t]{0.49\textwidth}
\centering
    \includegraphics[width=3.2in]{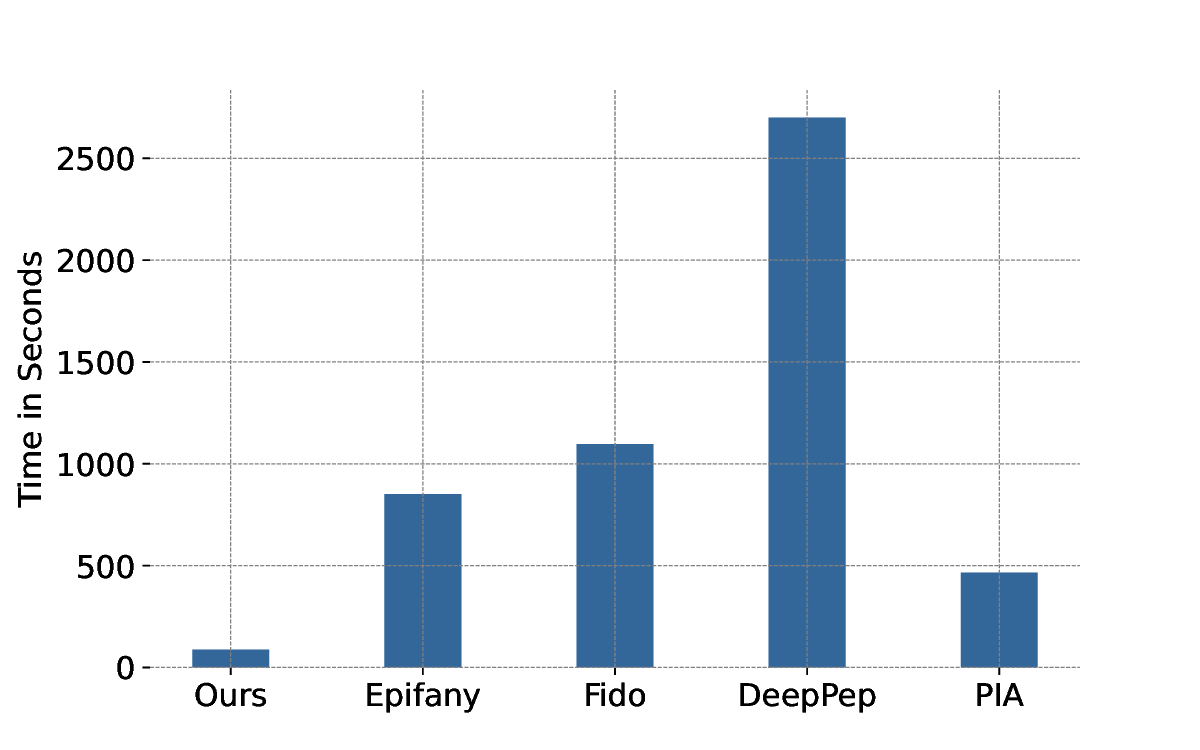}
    \caption{}
    \label{fig:yeast_time}
\end{subfigure}%\hspace{\fill} % maximize horizontal separation
\hfill
\begin{subfigure}[t]{0.49\textwidth}
\centering
    \includegraphics[width=3.2in]{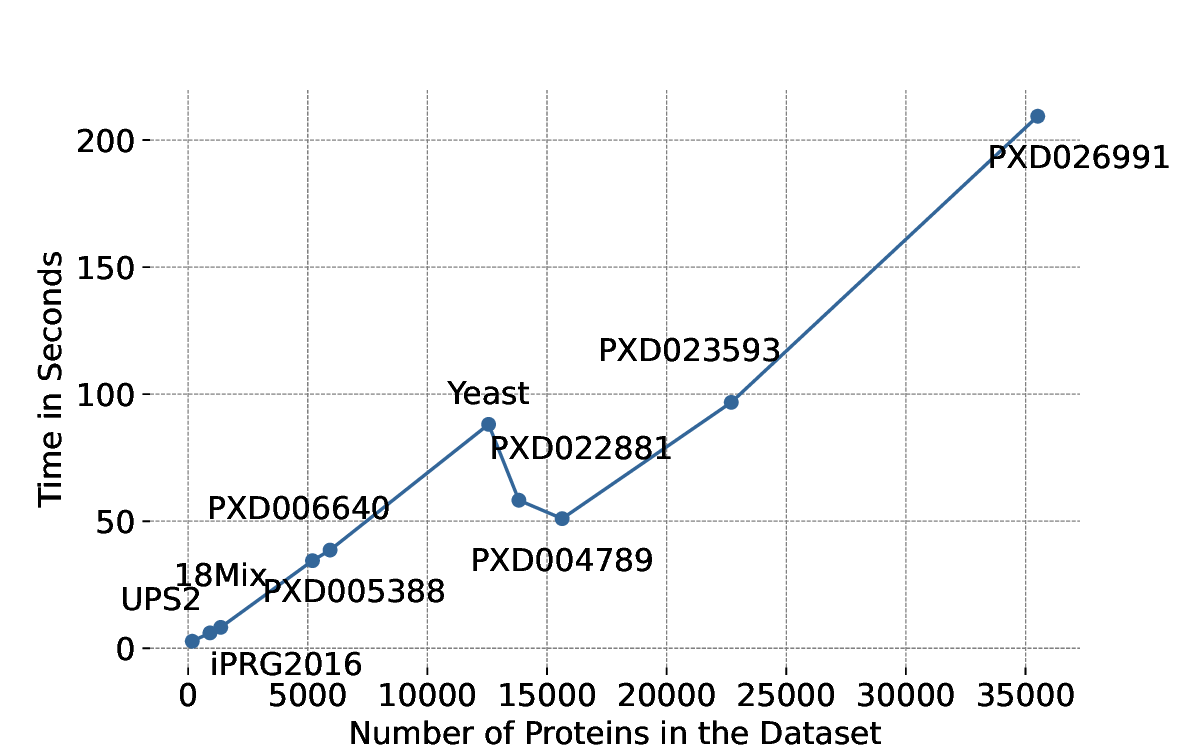}
    \caption{}
    \label{fig:linear_time}
\end{subfigure}
\caption{(a) Inference time of the benchmarked methods on the Yeast dataset. (b) Scaling of inference time for our model on datasets of different sizes.}
\end{figure}

Additionally, the design of the GNN model architecture guarantees that the runtime scales linearly with respect to the dataset size (number of proteins to be evaluated), eliminating the undesired effect of higher-order scaling. Figure~\ref{fig:linear_time} shows the runtime of our method plotted against the number of proteins in the dataset, demonstrating the linear scaling.

The efficiency of our method enables its application in large-scale datasets, which might be impossible for other methods due to computational constraints. 

\subsection{Extended Investigation}
\subsubsection{Experiments on Target-Decoy Labels}
Noticing that Barista~\citep{spivak2012direct} is removed from the latest version of Crux~\citep{mcilwain2014crux}, we create a comparable model using our GNN network as the foundational architecture. We trained this model on each test dataset, designating decoy proteins as negative samples and the remainder as positive. Adhering to Barista's documented procedures, we employed 5-fold cross-validation, averaging protein scores across the five models. We've termed this model ``Ours with Target Decoy Training''.

\begin{figure}[t!]
%\captionsetup[subfigure]%{justification=Centering}
\centering
\begin{subfigure}[t]{0.49\textwidth}
\centering
    \includegraphics[height=5.3cm]{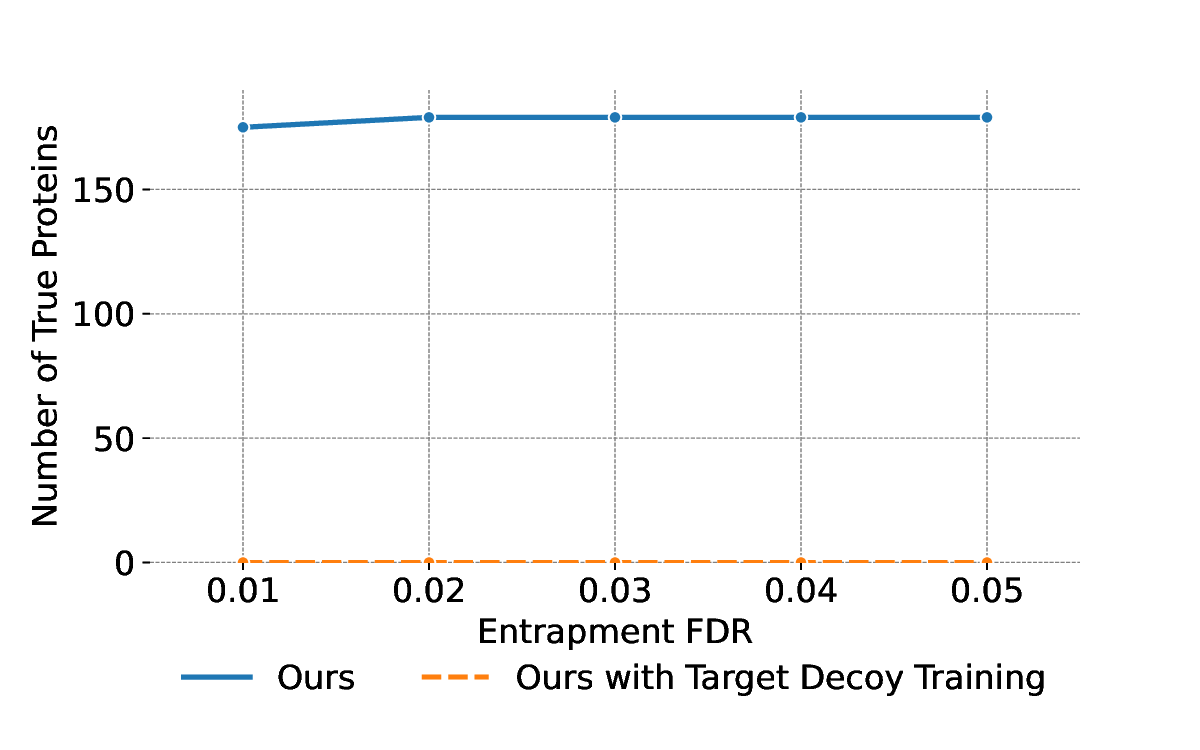}
    \caption{iPRG2016 A}
    \label{fig: iprg_a_roc_td}
\end{subfigure}\hspace{\fill} % maximize horizontal separation
\hfill
\begin{subfigure}[t]{0.49\textwidth}
\centering
    \includegraphics[width=3.2in]{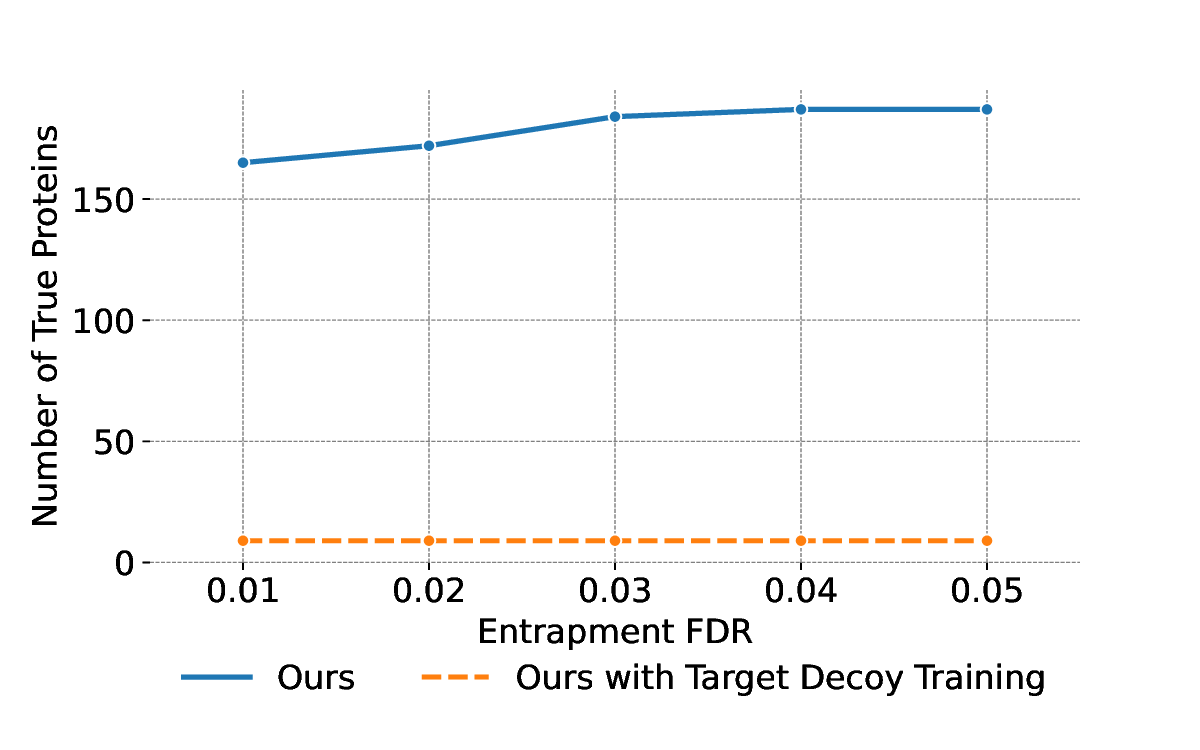}
    \caption{iPRG2016 B}
    \label{fig: iprg_b_roc_td}
\end{subfigure}

 % Add a line break after the second image

\begin{subfigure}[t]{0.49\textwidth}
\centering
    \includegraphics[width=3.2in]{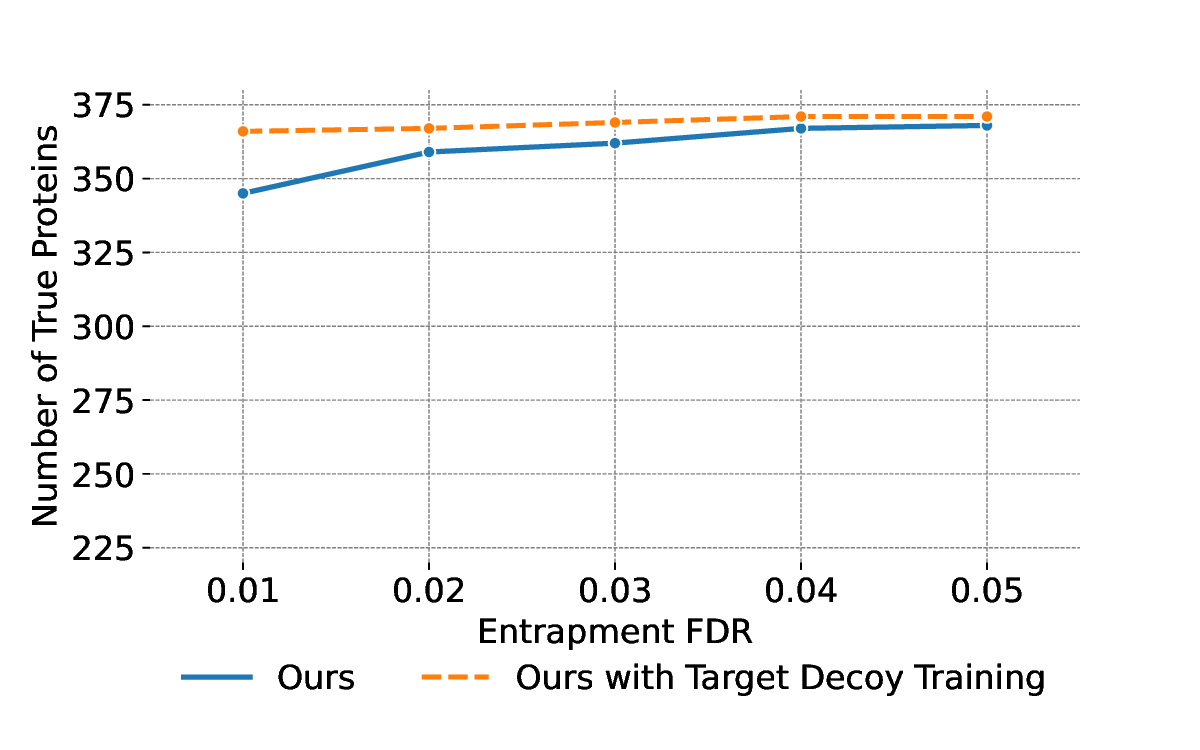}
    \caption{iPRG2016 AB}
    \label{fig: iprg_ab_roc_td}
\end{subfigure}\hspace{\fill} % maximize horizontal separation
\hfill
\begin{subfigure}[t]{0.49\textwidth}
\centering
    \includegraphics[width=3.2in]{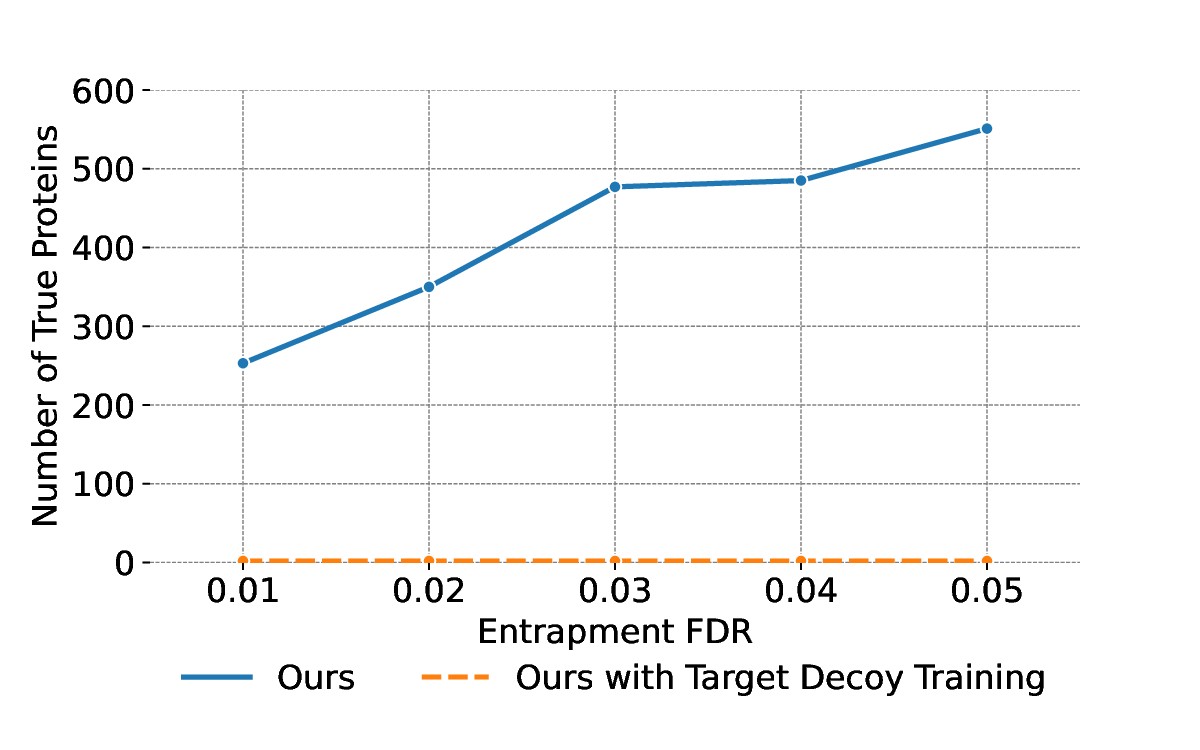}
    \caption{Yeast}
    \label{fig: yeast_roc_td}
\end{subfigure}

%\newline % Add a line break after the fourth image

\begin{subfigure}[t]{0.49\textwidth}
\centering
    \includegraphics[width=3.2in]{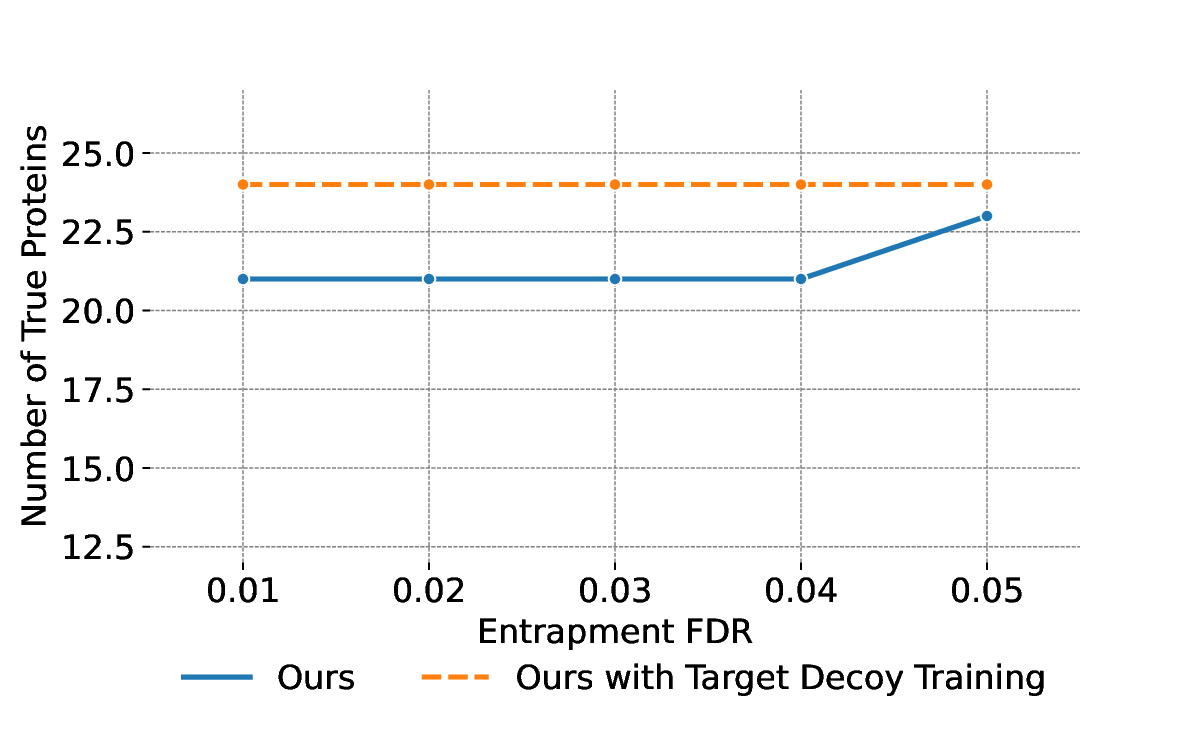}
    \caption{UPS2}
    \label{fig: ups2_roc_td}
\end{subfigure}\hspace{\fill} % maximize horizontal separation
\hfill
\begin{subfigure}[t]{0.49\textwidth}
\centering
    \includegraphics[width=3.2in]{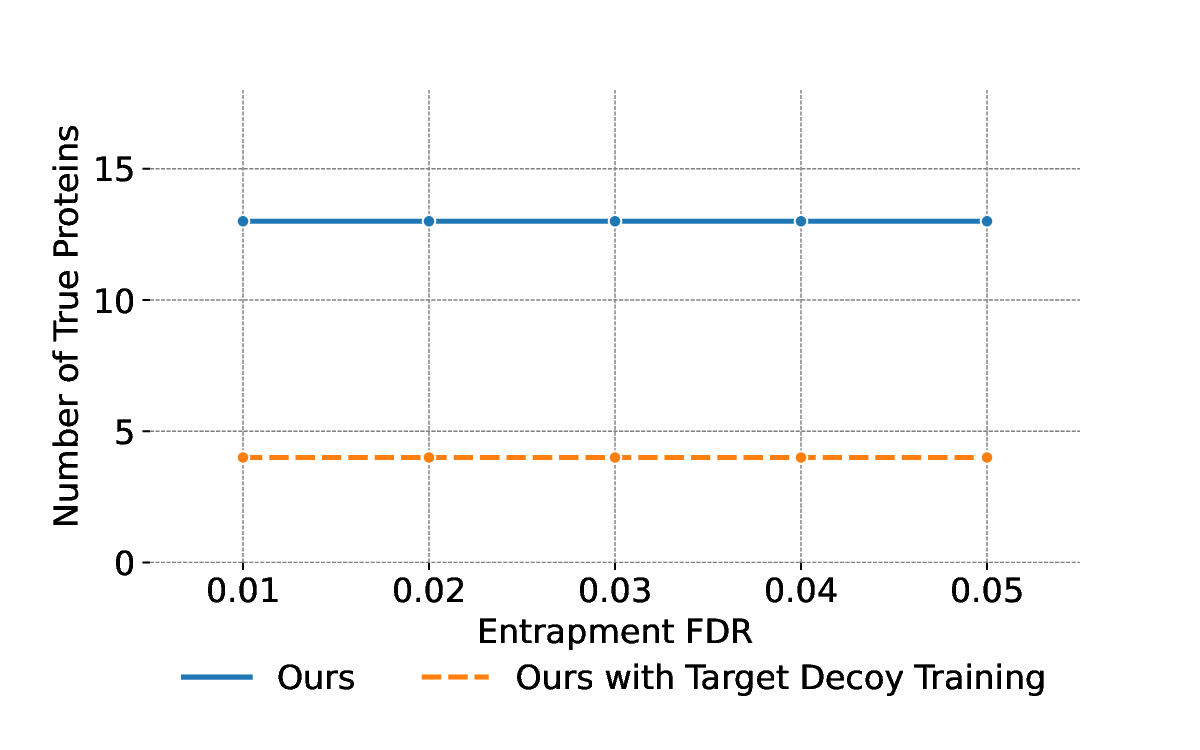}
    \caption{18Mix}
    \label{fig: 18mix_roc_td}
\end{subfigure}

% \caption{ROC curve (entrapment FDR vs. number of true proteins) of GraphPI and GraphPI with target-decoy training on the benchmark datasets.}
\caption{ROC curve (entrapment FDR vs. number of true proteins) of GraphPI and GraphPI with target-decoy training on the benchmark datasets: (a) iPRG2016 A, (b) iPRG2016 B, (c) iPRG2016 AB, (d) Yeast, (e) UPS2, and (f) 18Mix.}
\label{fig: comparison_td}
\end{figure}

The ROC curves of our GraphPI and the target-decoy setting is plotted in Figure \ref{fig: comparison_td}. It is evident that while this training setting yields favorable outcomes in the iPRG 2016 AB and UPS2 datasets, it underperforms in datasets featuring shared peptides among proteins (such as iPRG2016 A and B). This limitation stems from the fact that decoy proteins typically do not share peptides with target proteins, leading to a training gap where the model lacks exposure to examples necessitating differentiation between proteins that share peptides. Our approach effectively addresses this issue by incorporating pseudo-labels from Epifany which additionally imposes penalization on degenerate proteins.

% {\color{blue} The target decoy approach also underperforms when the dataset is larger, closer to real-world datasets, where contaminates proteins can have closer distribution to groundtruth proteins (such as Yeast and 18Mix), this caveat is also alleviated by our semi-supservised training scheme, where the model have seen real world contaminate proteins as negative labels.}

\subsubsection{Analysis on Epifany Results}
\label{sec: case study}
The benefit of GraphPI, a data-driven deep-learning-based approach, is its ability to learn the contribution of each peptide to its parental protein through the data distribution of a common set of datasets. In contrast, Epifany relies on prior probabilities and makes strong assumptions about the data distribution, which can lead to inaccurate protein scores. We provide two visual examples in the following paragraphs, supported by explanations that demonstrate the limitations of Epifany in dealing with certain types of proteins.

\begin{figure}[t!]
%\captionsetup[subfigure]%{justification=Centering}

\centering
\begin{subfigure}[t]{1\textwidth}
\centering
    \includegraphics[height=0.8cm]{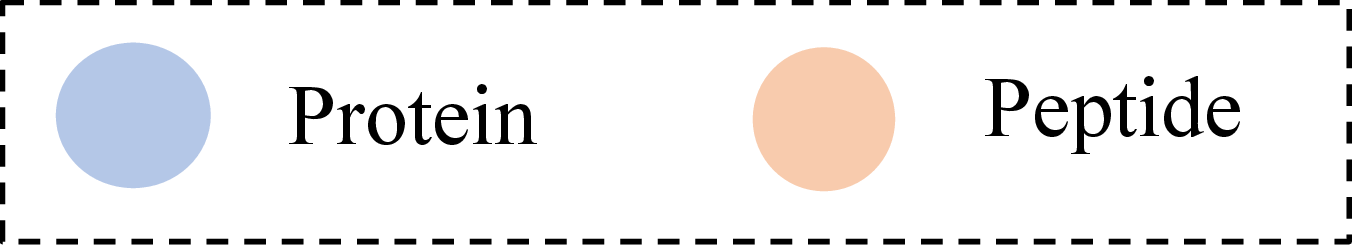}
\end{subfigure}
\newline
\centering
\begin{subfigure}[t]{0.43\textwidth}
\centering
    \includegraphics[height=5cm]{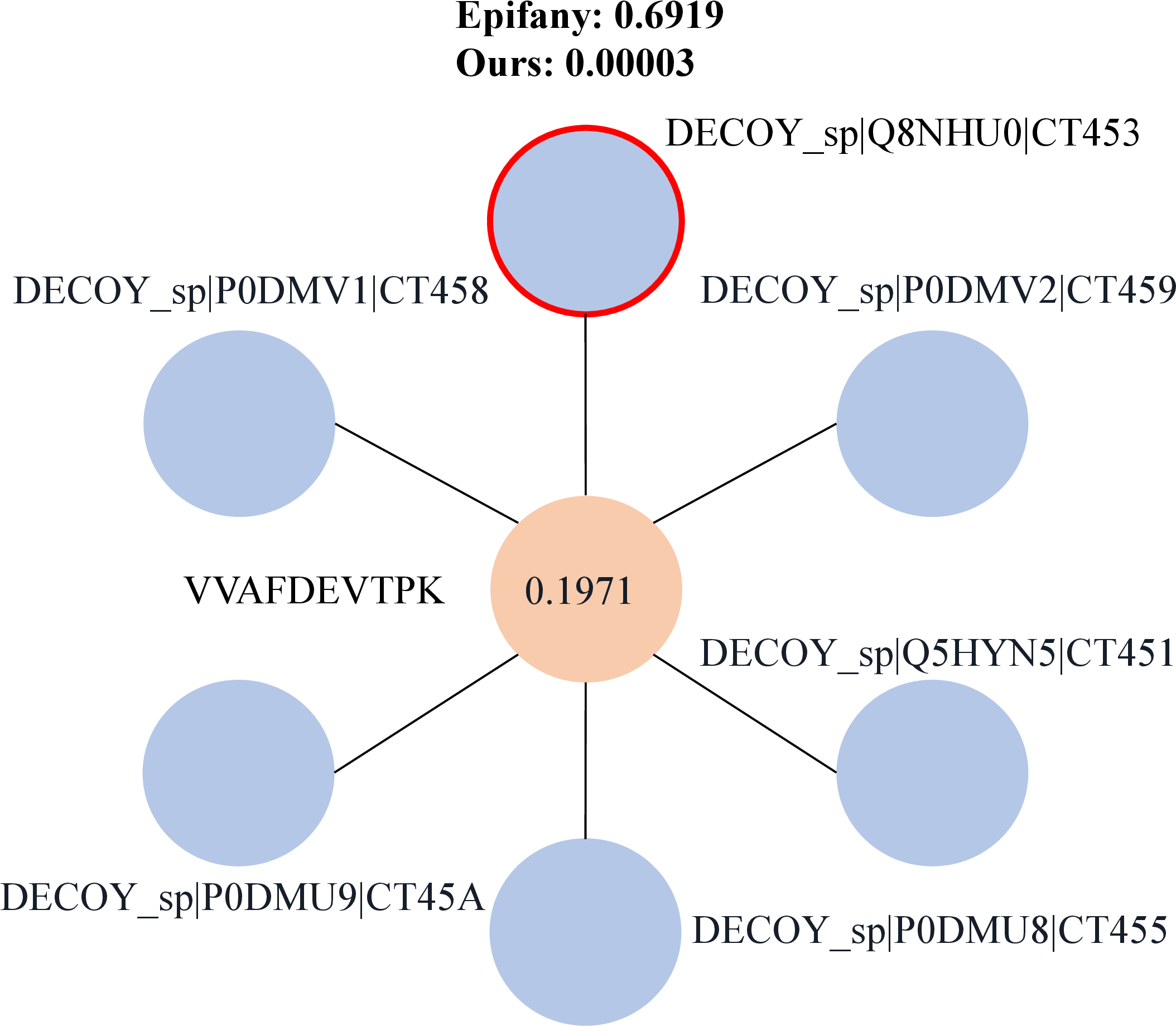}
    \caption{}
    \label{fig: case study - protein prior a}
\end{subfigure}%\hspace{\fill} % maximize horizontal separation
%\hfill
% \begin{subfigure}[t]{0.28\textwidth}
% \centering
%     \includegraphics[height=4cm]{figures/case_study_protein_prior_a.eps}
%     \caption{}
%     \label{fig: case study - protein prior b}
% \end{subfigure}
\begin{subfigure}[t]{0.54\textwidth}
\centering
    \includegraphics[height=5cm]{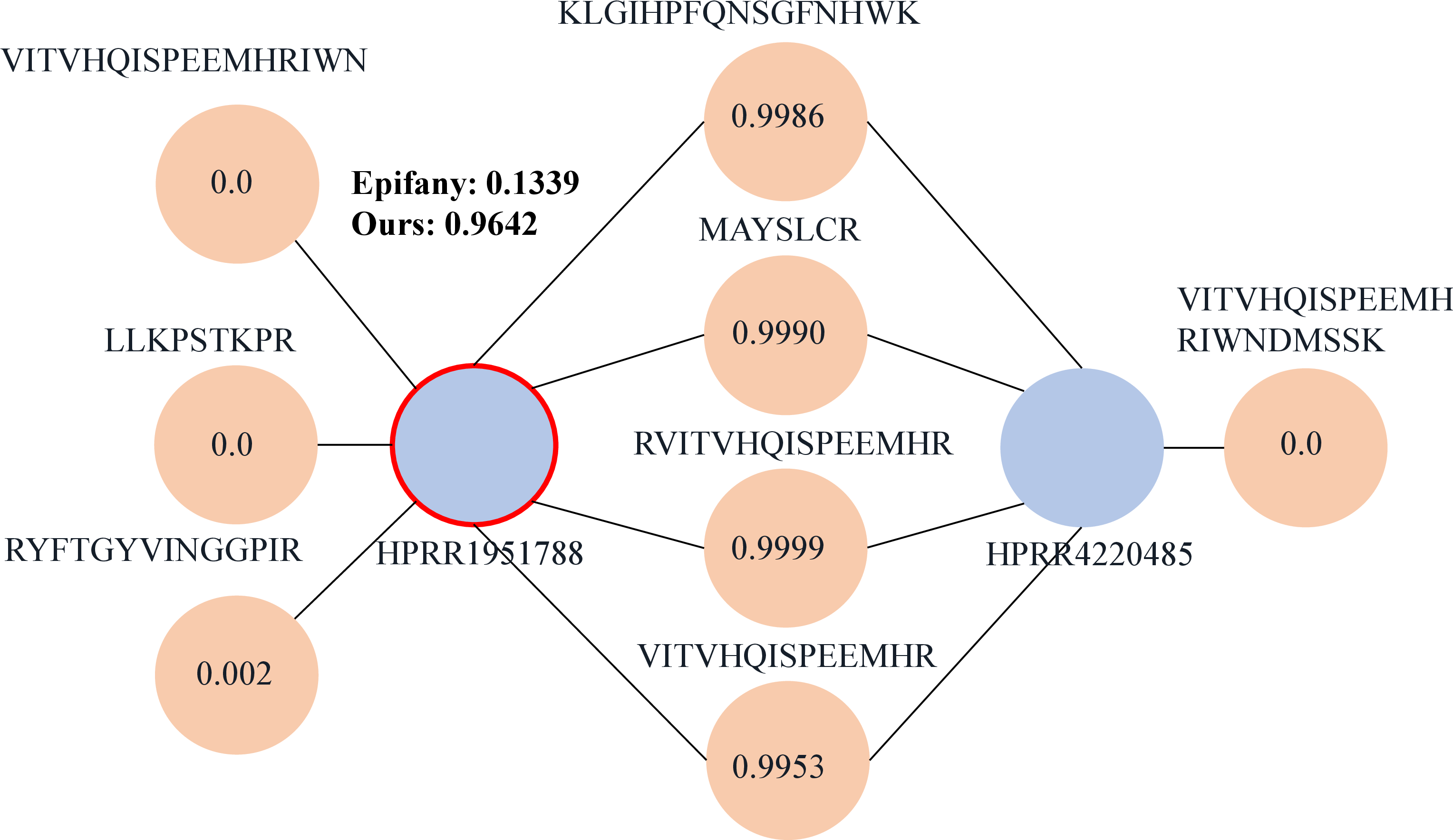}
    \caption{}
    \label{fig: case study - protein degenerate a}
\end{subfigure}
\caption{In (a)\&(b), the orange circle represents a peptide with its peptide identification score inside, while the blue circle represents a protein. The scores generated by both Epifany and our model are given for the highlighted protein in each graph, respectively. Accordingly, (a) shows the bipartite graphs for the selected proteins and their connected peptides from dataset ``PXD005388''. The protein prior is set to 0.7 based on the grid search program of Epifany. The highlighted protein is a decoy protein, which should has a lower score. (b) shows the bipartite graph for proteins selected from the dataset ``iPRG2016 AB''. The highlighted protein is a true protein, which should have a higher score.}
\end{figure}

\paragraph{Protein prior distribution presents an important role:}
In situations where a peptide has multiple siblings, Epifany tends to downweight the contribution of the peptide score and place greater reliance on the prior score of the protein, particularly in cases where there is no other supporting evidence (i.e., unique peptides connecting to the protein). This can lead to inaccuracies in protein inference, especially when the prior score of the protein is high and the dataset contains many peptides with shared proteins. We demonstrate an example of this phenomenon in Figure~\ref{fig: case study - protein prior a}, where the highlighted protein is a decoy protein that should not have received a high score, yet Epifany assigns a score that is close to its protein prior score, which is around 0.7. This is attributed to the fact that the prior score of proteins is optimized by the grid search program of Epifany based on Decoy FDR, which has the risk of overfitting the distribution of decoy rather than the actually contaminated proteins.

\paragraph{Epifany has strong penalization on degenerate proteins:} We use Figure~\ref{fig: case study - protein degenerate a} as an example for illustration: Specifically, the highlighted protein present in the figure is supposed to belong to the true protein group (having higher protein scores). However, Epifany assigns a relatively lower score to it regardless of the high-score peptides that connect to it. The reason behind this is that Epifany assigns a strong prior to peptides that have connections to multiple proteins, having the purpose of lowering their contribution to each parental protein. Specifically, Epifany sets the probability of a peptide that a certain number of proteins can generate to $1/N$, where $N$ is the number of proteins that the peptide is connected to. This prior probability affects the posterior probability score of the peptide, leading to a significant penalty in the contribution of the peptide to the calculation of each parental protein score. This approach gives Epifany superior performance in iPRG2016 A and B where most true proteins do not have shared peptides connected, but the performance on other datasets are lacking. 

\subsubsection{Learning beyond Epifany Scores}

Some may posit that GraphPI, which is trained on Epifany's pseudo-labels, might merely mimic Epifany's outputs without surpassing its performance. We argue that the divergence between GraphPI and Epifany can be distilled into three primary distinctions. 
\paragraph{Generalization over overfitting:}
As elucidated earlier in the case study, Epifany is prone to overfitting, particularly given its optimization strategy on priors anchored in Decoy FDR. In contrast, the training of GraphPI harnesses a set of public protein datasets. By building upon Epifany's results generated from diverse datasets, GraphPI inherently promotes broader generalization capabilities. As shown in both Figure~\ref{fig: case study - protein prior a} and Figure~\ref{fig: case study - protein degenerate a}, our method is able to produce a relatively lower score for the decoy protein and a higher score for the true protein, whereas the scores generated from Epifany is either inflated or deflated primarily due to its overfitting of prior probabilities on decoys of individual datasets.

\paragraph{Discerning label incorporation:}
In our label-generation mechanisms, decoy proteins, known to be artificial constructs and absent from biological samples, are consistently tagged as negative, irrespective of the scoring of Epifany. This information enables GraphPI to learn patterns surpassing those identified solely by Epifany. 
Supporting this claim, a comparison of GraphPI's performance with and without the presence of decoy proteins is shown in Table~\ref{table:decoy_ablation_1}. It is evident that the absence of decoy proteins significantly hampers our model's performance.

\begin{table}[H]
    \centering
    \begin{tabular}{| l|l| }
    \hline
    \textbf{Setting} & \textbf{True Positives} \\
    \hline
    Normal Train Data & 187 \\
    \hline
    Without Decoy & 176 \\
    \hline
    Decoys within 5\% FDR as Positive Samples & 183 \\
    \hline
    \end{tabular}
    \caption{Number of true positive (TP) proteins identified by our algorithm, with decoy proteins, without decoy proteins, or with decoys within 5\% FDR labeled as positive proteins. The numbers are acquired under 5\% entrapment FDR, from iPRG2016 B dataset, with only pseudo-labeled training (without self-training) to demonstrate the performance difference.}
    \label{table:decoy_ablation_1}
\end{table}

Nonetheless, an argument could be made that incorporating decoy proteins merely expands the training set, which in turn leads to enhanced performance. To address this concern, we conducted another experiment, where decoy proteins within a 5\% FDR threshold are labeled as positive (mimicking their incorrect identification by Epifany), and the remaining decoy proteins are labeled as negative. In this scenario, positive and negative sample labels are derived by thresholding Epifany's scores based on a 5\% FDR, resulting in labels that precisely match those obtained from Epifany. The results in Table~\ref{table:decoy_ablation_1} show that while integrating decoy data as hard negatives does enhance the model's efficacy, the presence of a few decoy proteins in the positive class can impair performance relative to our standard configuration.

\paragraph{Tailored GNN architecture, graph design, and self-training} Another aspect of GraphPI's enhanced capability is its GNN architecture, specifically tailored for a tripartite graph encompassing proteins, peptides, and PSMs.  This tailored structure adeptly captures the complex relationships within proteomic data, reducing the likelihood of overfitting to inaccuracies in Epifany's initial labels. Complementing this, GraphPI employs iterative self-training to refine these pseudo-labels from Epifany. This process allows our model to continually learn and adjust, potentially rectifying inaccuracies in the initial labels.

\subsubsection{Accuracy of FDR Estimation}

We provided an additional analysis on the relationship between entrapment FDR and decoy FDR in Figure S3. Generally speaking, our method provides less conservative FDR estimates than Epifany, while being more conservative than other methods in our comparison.

%% file: chapters/Conclusion.tex
\section{Conclusion}

In this study, we present GraphPI, a deep-learning framework designed to tackle the protein inference problem in proteomics. Our method conceptualizes proteins, peptides, and PSMs as interconnected entities within a protein-peptide-spectrum graph. By formulating the protein inference problem as a node classification task, we designed a GNN architecture inspired by GraphSAGE to handle the heterogeneous nature of the tri-partite graph. Recognizing the hurdle of limited labeled data in proteomics, We leverage large, unlabeled public protein datasets in a semi-supervised learning setting, utilizing pseudo-labels generated by existing protein inference algorithms. We further refined these labels by incorporating hard negative decoy protein information and employing self-training to iteratively improve the performance of the model. The experimental results demonstrated that our approach achieved superior performance across diverse test datasets. The use of GNNs and the semi-supervised training scheme contribute to significant improvements in protein inference accuracy. Furthermore, our method exhibited enhanced computational efficiency by leveraging the inherent parallelizability of neural networks. The universal applicability of GraphPI, trained on a common set of peptide identification data, eliminated the need for repetitive training processes on different datasets. In prospect, the realm of protein inference holds vast research potential. Future research directions could explore incorporating additional features and information, such as protein-protein interactions or post-translational modifications, to further enhance the performance of protein inference methods.

%% file: chapters/Supplementary.tex
\section{Supporting Table S1: PSM Features Utilized by Our Model}

\begin{table}
  \centering
\begin{tabularx}{1\textwidth}{| l | X| }
\hline
\textbf{Feature} & \textbf{Description} \\
    \hline
    pep & Peptide error probability from Percolator.\\
    \hline
    lnrSp & The natural logarithm of the rank of the match based on the Sp score.\\
    \hline
    deltLCn & The difference between this PSM's XCorr and the XCorr of the last-ranked PSM for this spectrum, divided by this PSM's XCorr or 1, whichever is larger.  \\
    \hline
    deltCn	& The difference between this PSM's XCorr and the XCorr of the next-ranked PSM for this spectrum, divided by this PSM's XCorr or 1, whichever is larger. \\
    \hline
    XCorr  & The SEQUEST cross-correlation score. \\
    \hline
    Sp &	The preliminary SEQUEST score. \\
    \hline
    IonFrac	& The fraction of b and y ions theoretical ions matched to the spectrum. \\
    \hline
    Mass & The observed mass $[M+H]^+$. \\
    \hline
    PepLen	& The length of the matched peptide, in residues. \\
    \hline
    Charge[n] & Is this a $1^+/2^+/3^+$ charged spectrum? \\
    \hline
    enzN & Is the peptide preceded by an enzymatic (tryptic) site? \\
    \hline
    enzC & Does the peptide have an enzymatic (tryptic) C-terminus? \\
    \hline
    enzInt & Number of missed internal enzymatic (tryptic) sites. \\
    \hline
    lnNumSP & The natural logarithm of the number of database peptides within the specified precursor range. \\
    \hline
    dM & The difference between the calculated and observed mass. \\
    \hline
    absdM & The absolute value of the difference between the calculated and observed mass. \\
    \hline
    %\captionof{PSM features we included in PSM nodes.}
    %\label{table:psm_features}
\end{tabularx}
    \caption{PSM features that we included in PSM nodes~\citep{mcilwain2014crux}.}
    \label{table:psm_features}
\end{table}

\newpage

\section{Supporting Table S2: Information about Training Data}

\begin{table}[h!]
\footnotesize
\centering
\resizebox{\linewidth}{!}{
\begin{tabular}{| l | l | l | l | l | l | l|}
\hline
\textbf{Identifier} & \textbf{\makecell{MS1\\ Tolerance}} & \textbf{\makecell{MS2 \\Tolerance}} & \textbf{Modifications} & \textbf{\makecell{Missed \\Cleavages}} & \textbf{Digestion} & \textbf{\# Replicates} \\
    \hline
    PXD004789 & 10 ppm & 0.05 Da & \makecell[l]{C(+57.02), M(+15.99), \\ N-term Acetyl} & 2 & Trypsin/Lys-C & 6 \\
    \hline
        PXD005388 & 10 ppm & 0.05 Da & \makecell[l]{C(+57.02), M(+15.99),\\N-term Acetyl} & 2 & Trypsin/Lys-C  & 4 \\
    \hline
        PXD006640 & 10 ppm & 0.02 Da & \makecell[l]{C(+57.02), M(+15.99), \\NQ(+0.98)} & 2 & Trypsin & 27\\
    \hline
        PXD010319 & 20 ppm & 0.01 Da & \makecell[l]{C(+57.02), M(+15.99),\\N-term Acetyl} & 2 & Trypsin/Lys-C  & 5 \\
    \hline
        PXD022881 & 20 ppm & 0.5 Da & \makecell[l]{C(+57.02), M(+15.99),\\N-term Acetyl} & 2 & Trypsin/P  &  11\\
    \hline

    PXD023034 & 10 ppm & 0.6 Da & \makecell[l]{C(+57.02), M(+15.99),\\STY(+79.99)} & 2 & Trypsin& 7\\
    \hline
        PXD023593 & 10 ppm & 0.01 Da & C(+57.02), M(+15.99) & 1 & Trypsin & 7 \\
    \hline
        PXD025701 & 4.5 ppm & 0.02 Da & \makecell[l]{C(+57.02), M(+15.99),\\N-term Acetyl} & 2 & Trypsin & 11\\
    \hline
        PXD026991 & 10 ppm & 0.6 Da & C(+57.02), M(+15.99) & 2 & Trypsin/Lys-C & 7 \\
    \hline
    
        PXD030330 & 10 ppm & 0.01 Da & \makecell[l]{C(+57.02), M(+15.99),\\N-term Acetyl} & 5 & Trypsin&  5\\
    \hline
        PXD030448 & 10 ppm & 0.01 Da & \makecell[l]{C(+57.02), M(+15.99),\\NQ(+0.98)} & 1 & Trypsin & 12\\
    \hline
        PXD032035 & 10 ppm & 0.02 Da & C(+57.02), M(+15.99) & 2 & Trypsin & 5\\
    \hline
        PXD032284 & 10 ppm & 0.01 Da & C(+57.02), M(+15.99) & 1 & Trypsin  & 37\\
    \hline

        PXD034012 & 10 ppm & 0.01 Da & \makecell[l]{C(+57.02), M(+15.99),\\NQ(+0.98)} & 1 & Trypsin & 9\\
    \hline
        PXD035125 & 20 ppm & 0.01 Da & \makecell[l]{C(+57.02),M(+15.99),\\STY(+79.99), \\ N-term Acetyl} & 2 & Trypsin/P & 9\\
    \hline
        PXD036171 & 10 ppm & 0.01 Da & C(+57.02), M(+15.99) & 1 & Trypsin  & 4\\
    \hline

        PXD039272 & 0.5 Da & 0.01 Da & \makecell[l]{C(+57.02), M(+15.99),\\NQ(+0.98)} & 2 & Trypsin & 13\\
    \hline
\end{tabular}}
\caption{Search parameters for training datasets.}
\label{table:training_datasets_search}
\end{table}

% Every training dataset is searched against the Uniprot (\url{https://www.uniprot.org/}) human isoform database (July 27 2022, 220301 entries), since every dataset lists only homo sapiens as species. We chose only unlabelled DDA data for our experiments for simplicity. 

Every training dataset is searched against the Uniprot human isoform database (July 27 2022, 220301 entries), since every dataset lists only homo sapiens as species. We chose only unlabelled DDA data for our experiments for simplicity. 

%\newpage

\section{Supporting Material S1: Detailed Description of Testing Data}

In this section we introduce each test dataset, and their characteristics. Also, data preprocessing protocols are included.

\paragraph{iPRG2016~\citep{iprg2016}:} The iPRG2016 dataset (with identifier PXD008425) is designed specifically for benchmarking protein inference algorithms on their ability to process proteins sharing peptides. 
To create the dataset, two different samples of defined protein content were created from pairs of proteins that share peptides. Recombinant protein fragments from the Human Proteome Atlas project are selected in silico, resulting in 191 pairs of sequences such that for each pair, the two proteins share at least one identical fully trptic peptide. Selected recombinant protein fragments were then cloned into an expression vector and thereafter transformed into an E.coli strain for production. Each protein sequence has a similar length between 50 and 150 acids.
    
Each recombinant protein in a pair is ramdonly assigned into one of the mixtures (A or B) that are created, while a third mixture is created by mixing mixtures A and B, resulting in the mixture AB.

An amount of 1.8 pmol of each recombinant protein fragment is added to either mixture A or B, and an amount of 0.9 pmol of all fragments is added to mixture AB. Each pool is mixed into a background of a tryptic digest of 100 ng E. coli.
Finally, each mixture is analyzed by LC-MS/MS with a Q-Exactive HF mass spectrometer in DDA mode. In addition, we are provided with fasta files containing the amino acid sequences of the 191 chosen pairs of protein fragments, along with a set of entrapment sequences of 1,000 recombinant protein fragments that are not selected for the mixtures, hence are absent in the sample. These entrapment sequences have a similar length distribution compared to the selected sequences. The combined search database, consisting of proteins from mixture A, B, as well as common contaminates and decoys, has in total 2,766 proteins. 3 replicates are analyzed together for each of the 3 mixtures, making a total of 9 raw files.

Shown in Figure \ref{fig: proteinshare} and Figure \ref{fig: pepshare}, this dataset contains a large amount of shared peptides. For iPRG2016 B, 48.69\% proteins in the groundtruth share peptide with the contaminate proteins.
    
\paragraph{UPS2~\citep{ups2}:} UPS2\footnote{http://www.marcottelab.org/MSdata/Data\_13/} is a mixture of 48 individual human proteins, which presents a dynamic range of 5 orders of magnitude, ranging from 50 pmols to 500 amols. However, unlike iPRG2016, the proteins in this dataset does not share peptides between each other. This dataset serves as a nice addition to iPRG2016, allowing us to test the performance of protein inference algorithms on regular proteins. Proteins are digested with trypsin, then analyzed by a LTQ-Orbitrap mass spectrometer.

We choose the raw file corresponding to the highest concentrated sample (30 $\mu L$) to produce the most discoverable proteins. The provided fasta file included the 48 known protein sequences, plus 6 different shuffled decoy sequences for each protein, making a total of 350 proteins. One set of the six shuffled decoys is treated as entrapment proteins, being masked so that Comet search will not be provided the full ground-truth information.

This is considered a ``simpler'' dataset, since few peptides are shared between proteins.

\paragraph{18Mix~\citep{18mix}:} In this dataset{\footnote{http://regis-web.systemsbiology.net/PublicDatasets/} 18 proteins from different organisms are chosen to produce the sample, with 1 nmol of each protein being utilized. Unlike UPS2, the ground-truth proteins and entrapment proteins can share peptides with each other. The same sample is analyzed by a variety of mass spectrometers to evaluate their performance.

In our experiment, we choose the data (11 replicates) from LTQ to evaluate the test algorithms. Based on the original fasta file containing the 18 chosen sequences, as well as 1,802 entrapment protein sequences, we generated a decoy database with shuffling. 

 Note that this is a dataset generated from a low-resolution mass spectrometry experiment, showing that our model can be directly adopted to data with various precision.

\paragraph{Yeast~\citep{ramakrishnan2009gold}:}Unlike the above three datasets, where proteins are artificially selected or generated to create a mixture, the yeast dataset consists of the whole proteome of wild-type yeast grown in rich medium to log-phase. Cell lysate was processed and analyzed with LC-MS/MS with an Thermo LTQ Orbitrap, using a total of 8 injections with varying conditions for optimization, resulting in a total of 32 raw files. The link to the dataset is included in the citation.

Since we do not have a mixture of defined protein content, it is difficult to obtain a ground truth for this specific dataset. To counter this, the author assembled other datasets generated by the same biological sample (wild-type yeast in log-phase), and documented their identified proteins. The ground truth is generated by the union of all identified proteins, which consist of 4,265 proteins, since the original biological sample is the entire protein expression of yeast. The total number of proteins in the yeast genome is 6,330, and the remaining ones are likely to be absent in this stage of growth. These absent proteins are treated as entrapment sequences when computing false discovery rate. The database of yeast proteome and decoys has size 13,428. Moreover, it is considered as a ``simpler'' dataset given that the majority of proteins in the dataset do not share peptides, as can be seen from Figure \ref{fig: yeast_proteinshare} and \ref{fig: yeast_pepshare}.

\paragraph{Hela-3T3~\citep{saltzman2018gpgrouper}:} 
We also evaluated the models on the PXD008560 \cite{saltzman2018gpgrouper} dataset. This dataset provides a protein mixture of human hela cells and mouse 3t3 cells, with different ratios. We conducted our protein inference under the database of combined human and mouse proteome (with 20,419 human proteins, and 17,202 mouse proteins).
For this dataset, we do not have a groundtruth of the hela and 3t3 cell lines, so we test under the two-species scheme: first, we take the data consisting of 100\% hela cells and 0\% 3t3 cells (5 replicates), and adopt the human proteins as true labels, while mouse proteins being contaminates. We conduct the same experiments on data with 0\% hela and 100\% 3t3 cells as well (5 replicates), with mouse proteins being true labels. 

We observe a large level of peptide sharing in the data, similar to the level of iPRG2016 A and B. This allows us to test our algorithm on real-world complex dataset.

\section{Supporting Figures S1-S2: Peptide Sharing Statistics Among Test Datasets}

We include two pie charts for each dataset: the first pie chart represents the portion of groundtruth proteins sharing peptides with contaminates proteins. Since the objective is to differentiate groundtruth proteins from contaminates, peptide sharing between them makes the task more difficult, and the model will need strategies such as the ``explain away effect'' to adjust the scores of the contaminates.

The second pie chart describes the percentages of peptides that belongs to only 1 (no sharing), 2, or 3+ proteins, it also represents the level of peptide sharing in a dataset, but we do not need groundtruth proteins to compute this statistics. 

\begin{figure}[t!]
%\captionsetup[subfigure]%{justification=Centering}
% \centering
% \begin{subfigure}[t]{1\linewidth}
% \centering
%     \includegraphics[height=2cm]{figures/proteinsharing_legend_iPRG2016_A.eps}
%     %\caption{iPRG2016 A}
%     %\label{fig: iprg_a_proteinshare}
% \end{subfigure}
% \newline
\centering
\begin{subfigure}[t]{0.49\textwidth}
\centering
    \includegraphics[width=3.2in]{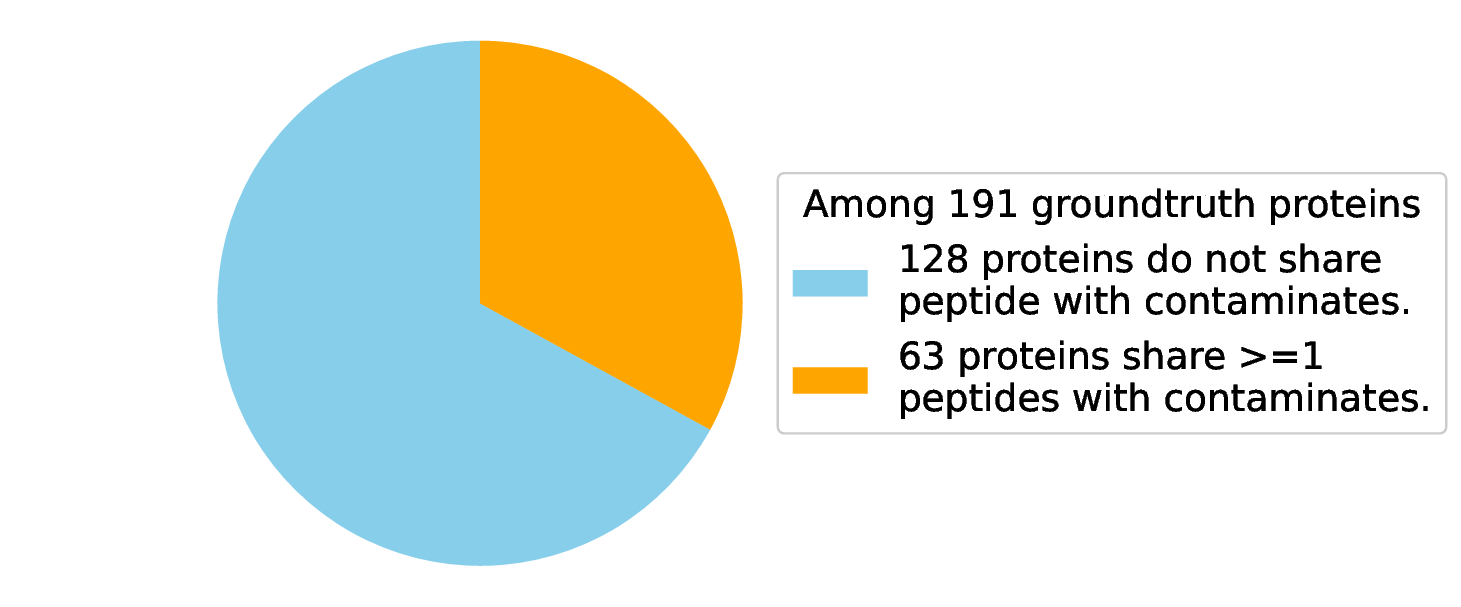}
    %\vspace{-1cm}
    \caption{iPRG2016 A}
    \label{fig: iprg_a_proteinshare}
\end{subfigure}
\centering
\begin{subfigure}[t]{0.49\textwidth}
\centering
    \includegraphics[width=3.2in]{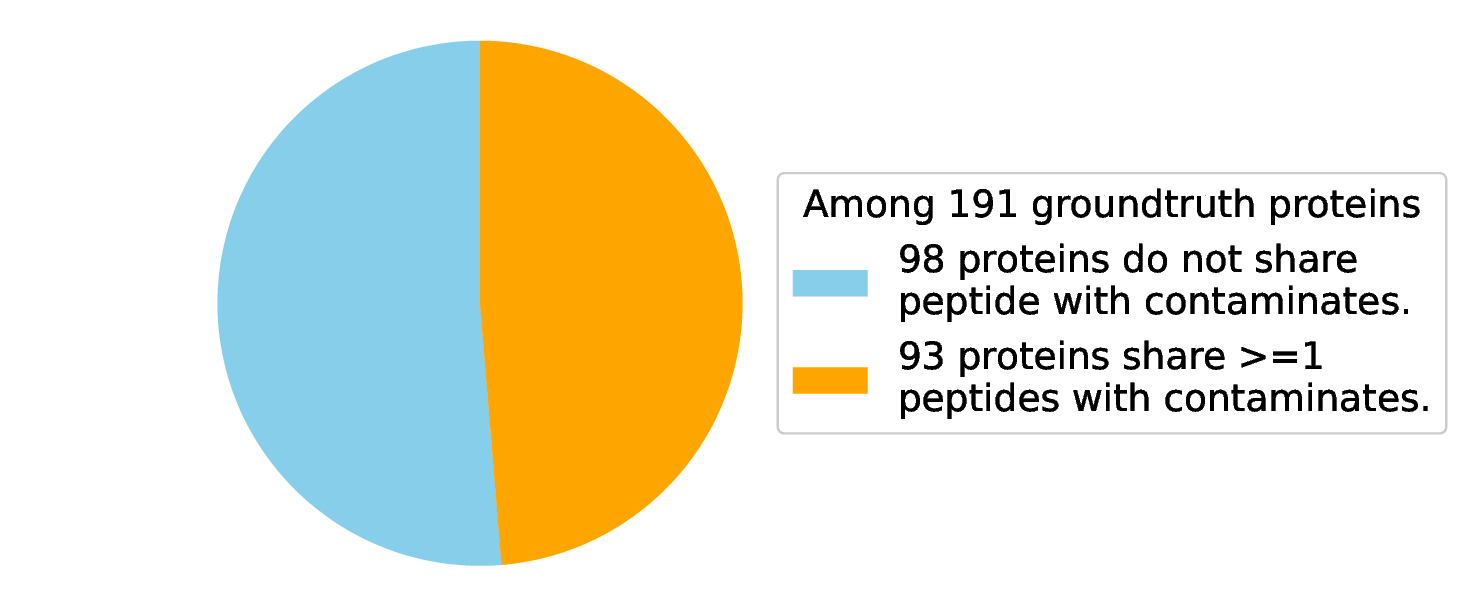}
    %    \vspace{-1cm}
    \caption{iPRG2016 B}
    \label{fig: iprg_b_proteinshare}
\end{subfigure}
 % Add a line break after the second image
\centering
\begin{subfigure}[t]{0.49\textwidth}
\centering
    \includegraphics[width=3.2in]{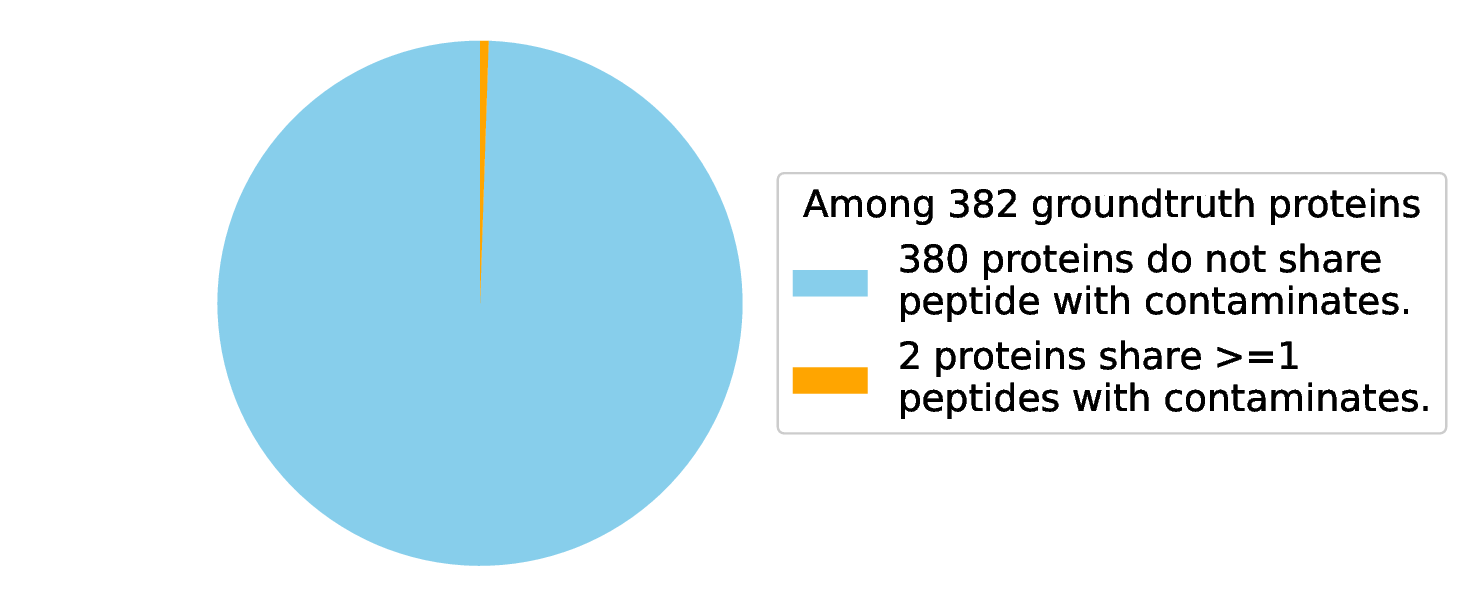}
     %   \vspace{-1cm}
    \caption{iPRG2016 AB}
    \label{fig: iprg_ab_proteinshare}
\end{subfigure}\hspace{\fill} % maximize horizontal separation
\hfill
\centering
\begin{subfigure}[t]{0.49\textwidth}
\centering
    \includegraphics[width=3.2in]{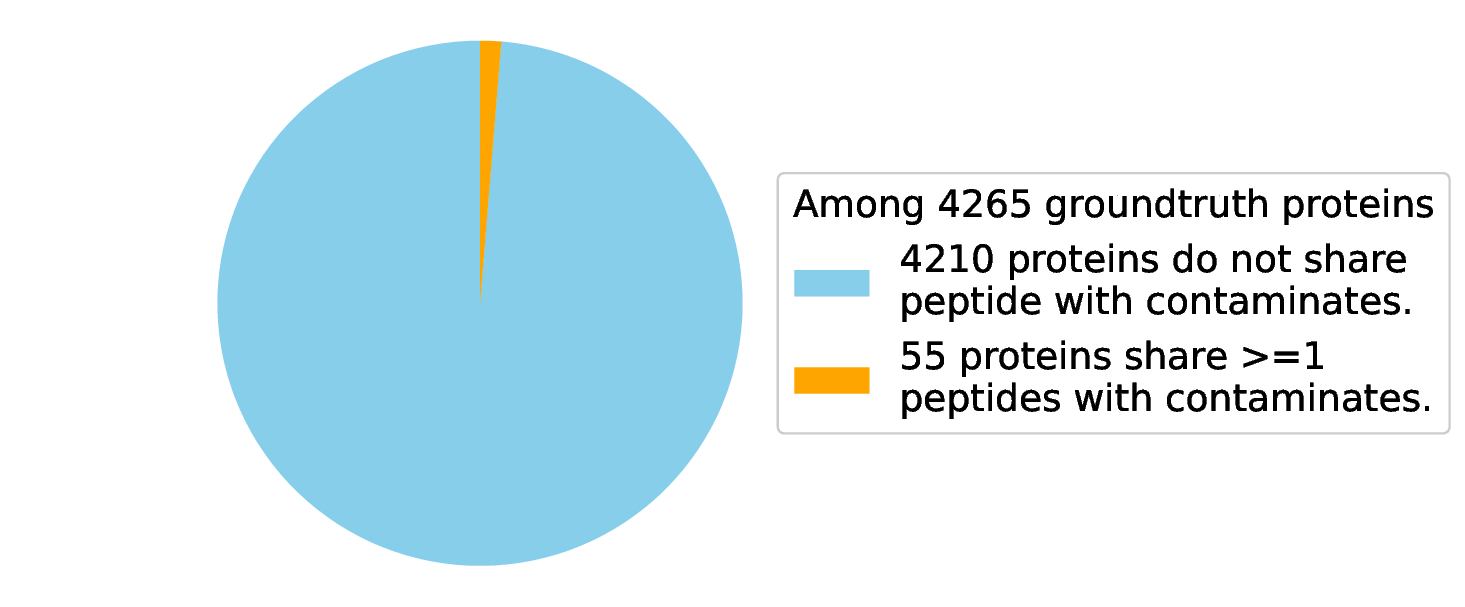}
     %   \vspace{-1.5cm}
    \caption{Yeast}
    \label{fig: yeast_proteinshare}
\end{subfigure}

%\newline % Add a line break after the fourth image
\centering
\begin{subfigure}[t]{0.49\textwidth}
\centering
    \includegraphics[width=3.2in]{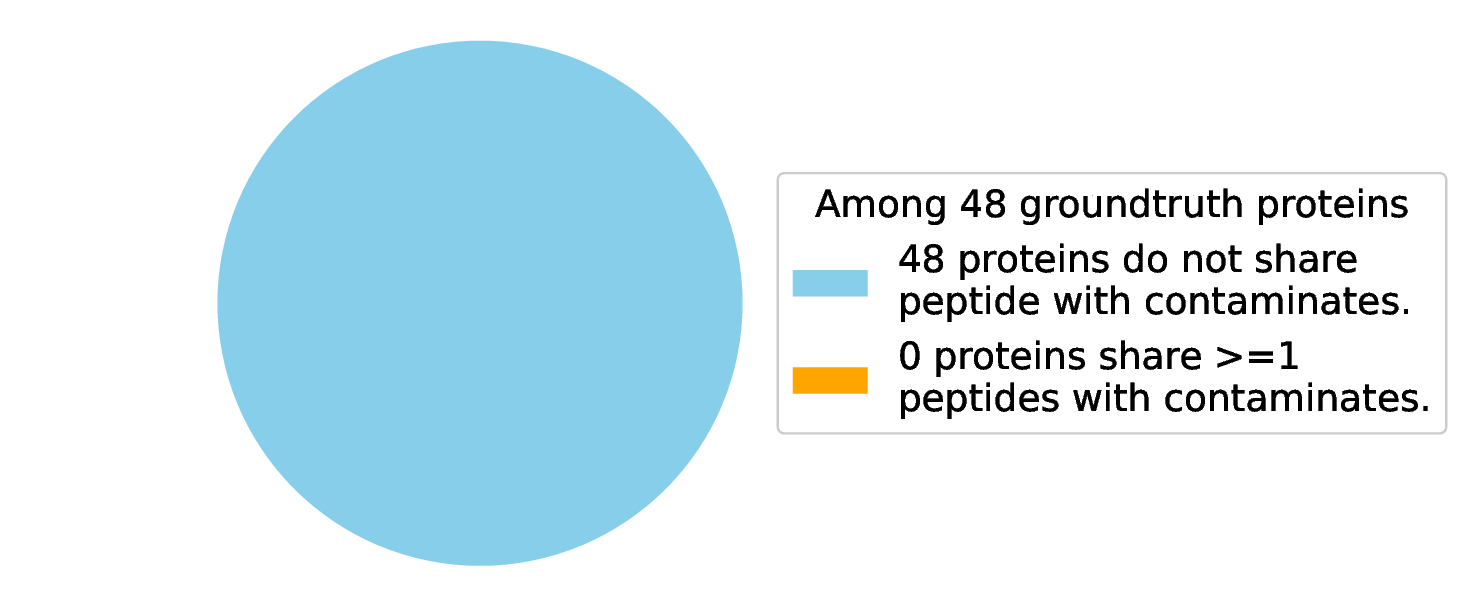}
     %   \vspace{-1.5cm}
    \caption{UPS2}
    \label{fig: ups2_proteinshare}
\end{subfigure}\hspace{\fill} % maximize horizontal separation
\hfill
\centering
\begin{subfigure}[t]{0.49\textwidth}
\centering
    \includegraphics[width=3.2in]{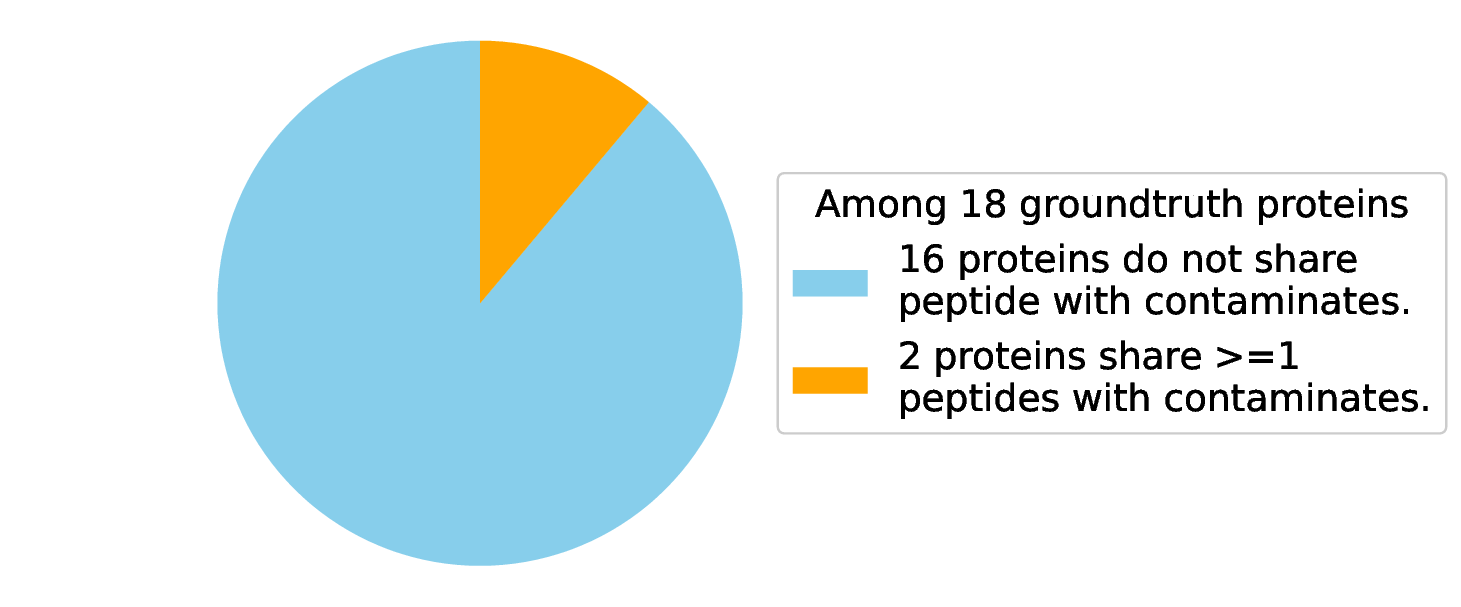}
    %    \vspace{-1.5cm}
    \caption{18Mix}
    \label{fig: 18mix_proteinshare}
\end{subfigure}

\centering
\begin{subfigure}[t]{0.49\textwidth}
\centering
    \includegraphics[width=3.2in]{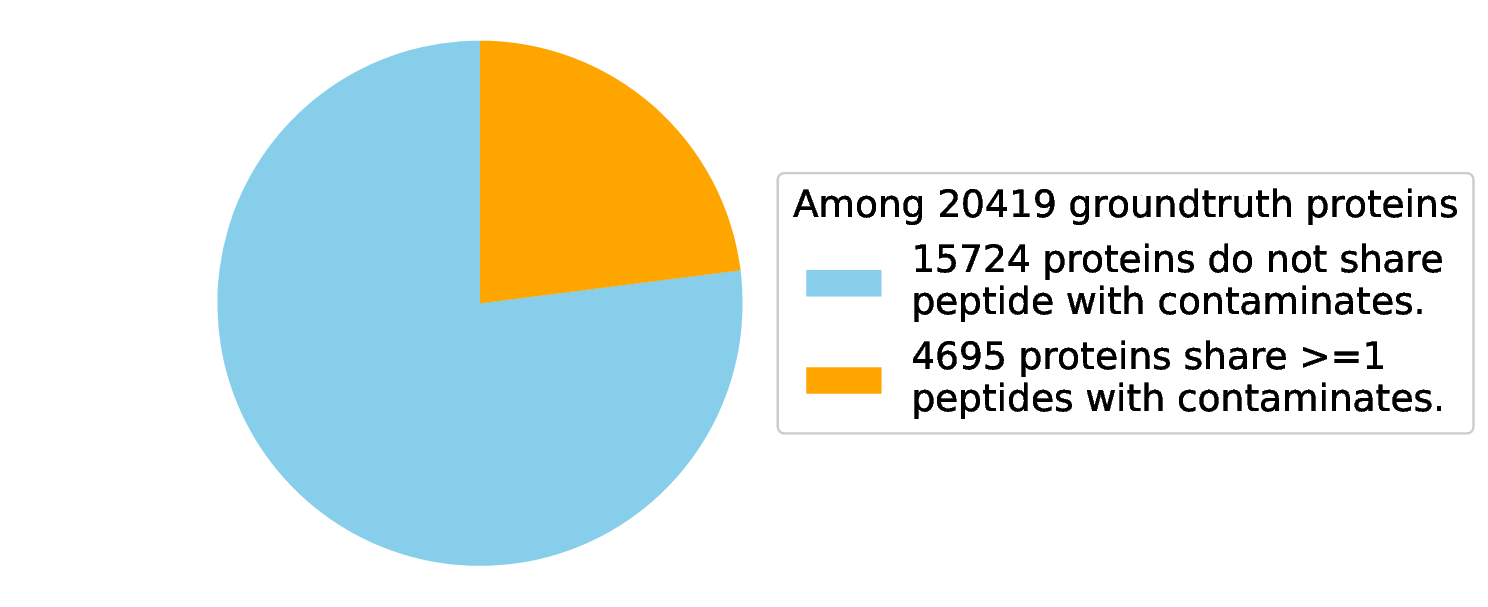}
    %    \vspace{-1.5cm}
    \caption{Hela}
    \label{fig: hela_proteinshare}
\end{subfigure}\hspace{\fill} % maximize horizontal separation
\hfill
\centering
\begin{subfigure}[t]{0.49\textwidth}
\centering
    \includegraphics[width=3.2in]{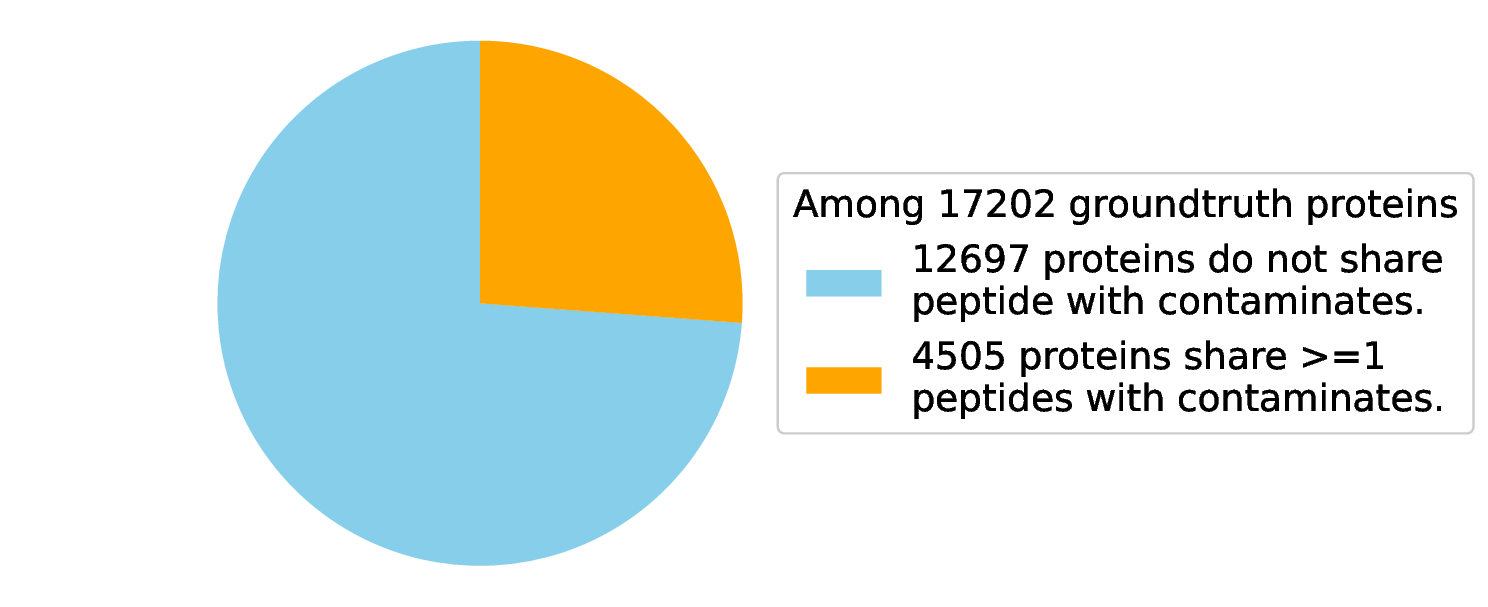}
     %   \vspace{-1.5cm}
    \caption{3T3}
    \label{fig: 3t3_proteinshare}
\end{subfigure}

\caption{Portion of groundtruth proteins sharing peptides with contaminate proteins.}
\label{fig: proteinshare}
\end{figure}

\begin{figure}[t!]
%\captionsetup[subfigure]%{justification=Centering}
\centering
\begin{subfigure}[t]{0.49\textwidth}
\centering
    \includegraphics[width=3.2in]{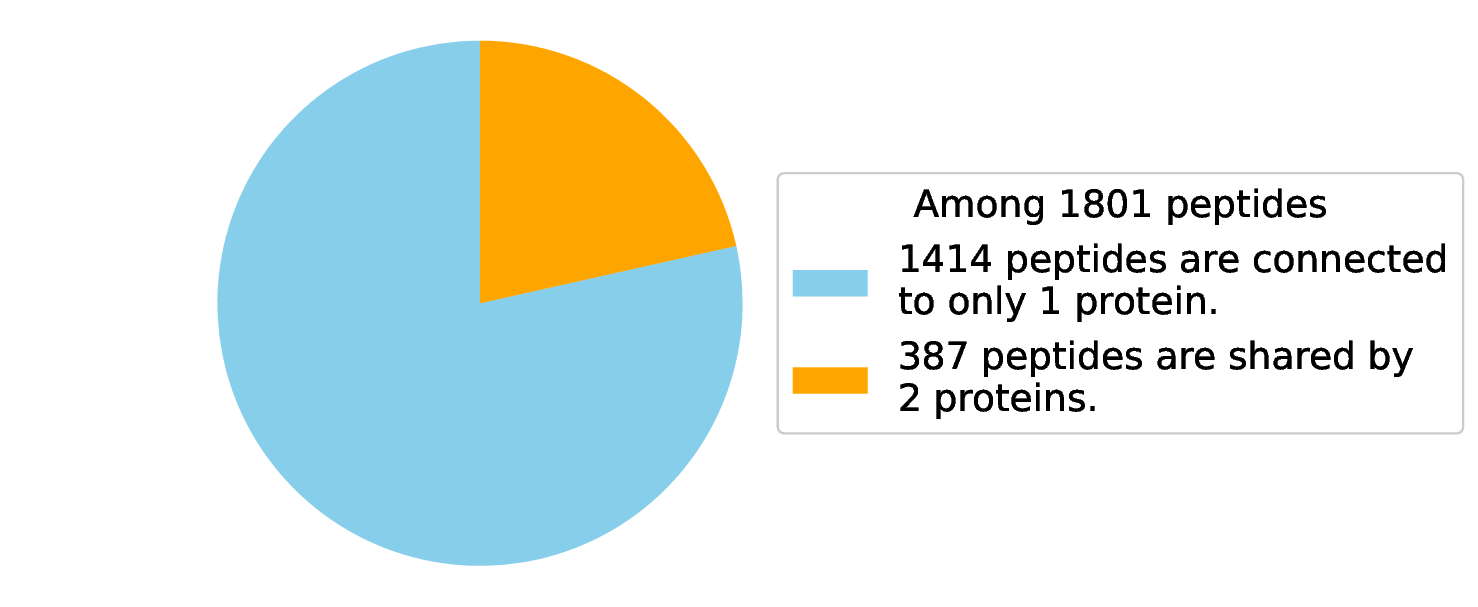}
    \caption{iPRG2016 A}
    \label{fig: iprg_a_pepshare}
\end{subfigure}\hspace{\fill} % maximize horizontal separation
\hfill
\begin{subfigure}[t]{0.49\textwidth}
\centering
    \includegraphics[width=3.2in]{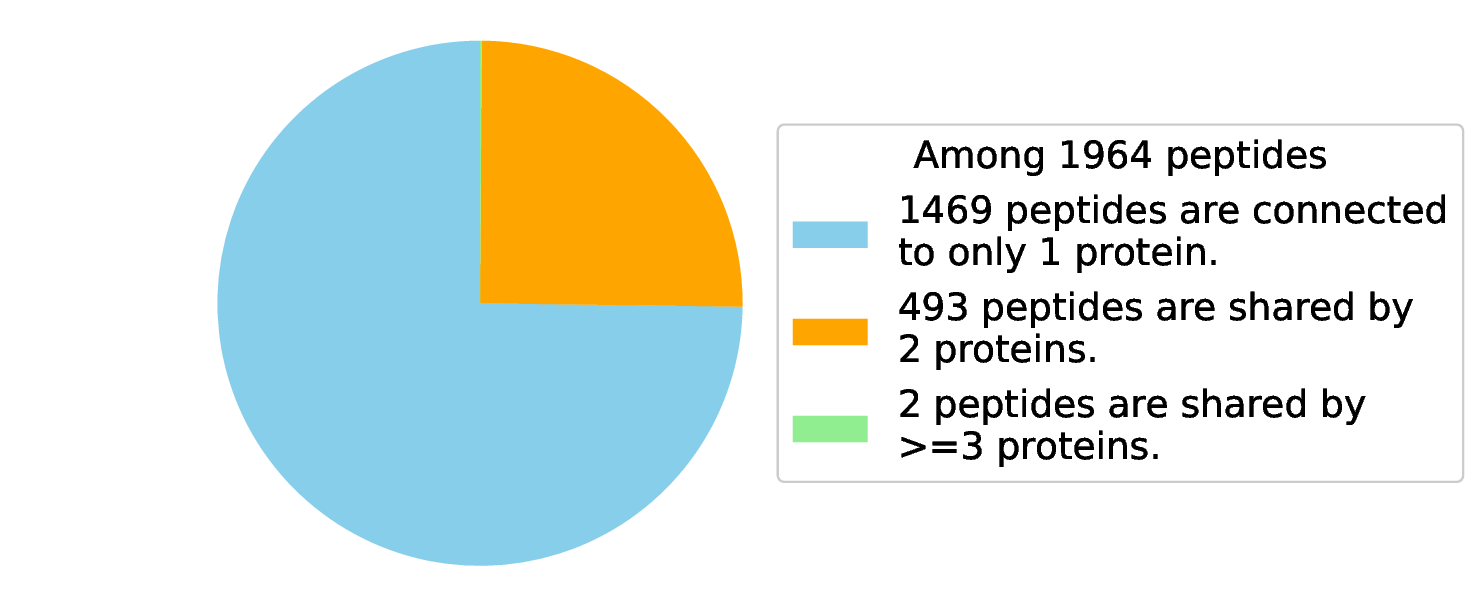}
    \caption{iPRG2016 B}
    \label{fig: iprg_b_pepshare}
\end{subfigure}

 % Add a line break after the second image

\begin{subfigure}[t]{0.49\textwidth}
\centering
    \includegraphics[width=3.2in]{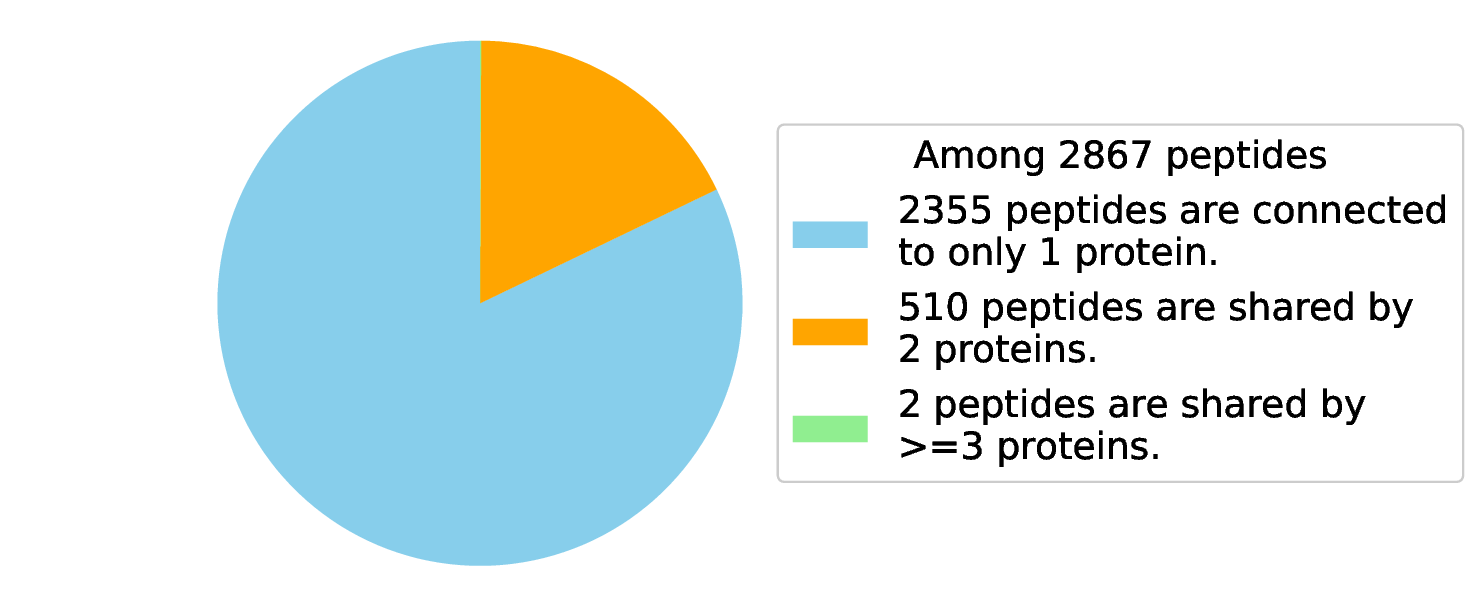}
    \caption{iPRG2016 AB}
    \label{fig: iprg_ab_pepshare}
\end{subfigure}\hspace{\fill} % maximize horizontal separation
\hfill
\begin{subfigure}[t]{0.49\textwidth}
\centering
    \includegraphics[width=3.2in]{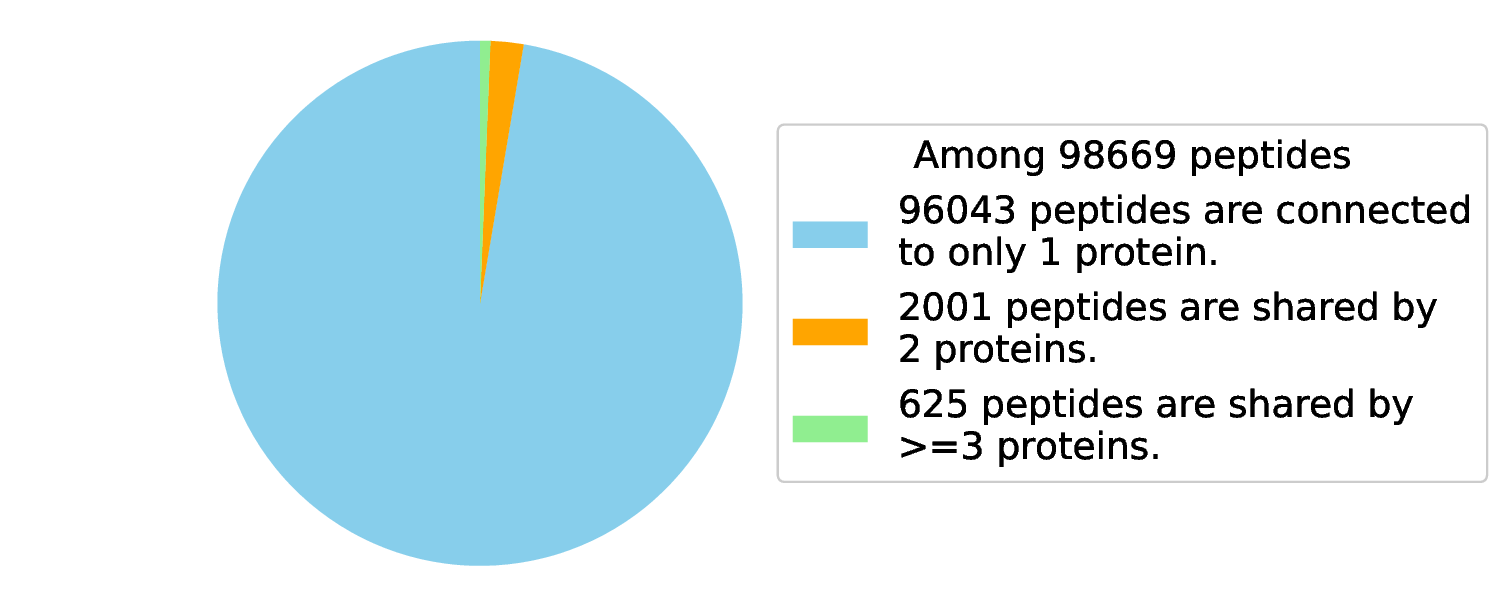}
    \caption{Yeast}
    \label{fig: yeast_pepshare}
\end{subfigure}

%\newline % Add a line break after the fourth image

\begin{subfigure}[t]{0.49\textwidth}
\centering
    \includegraphics[width=3.2in]{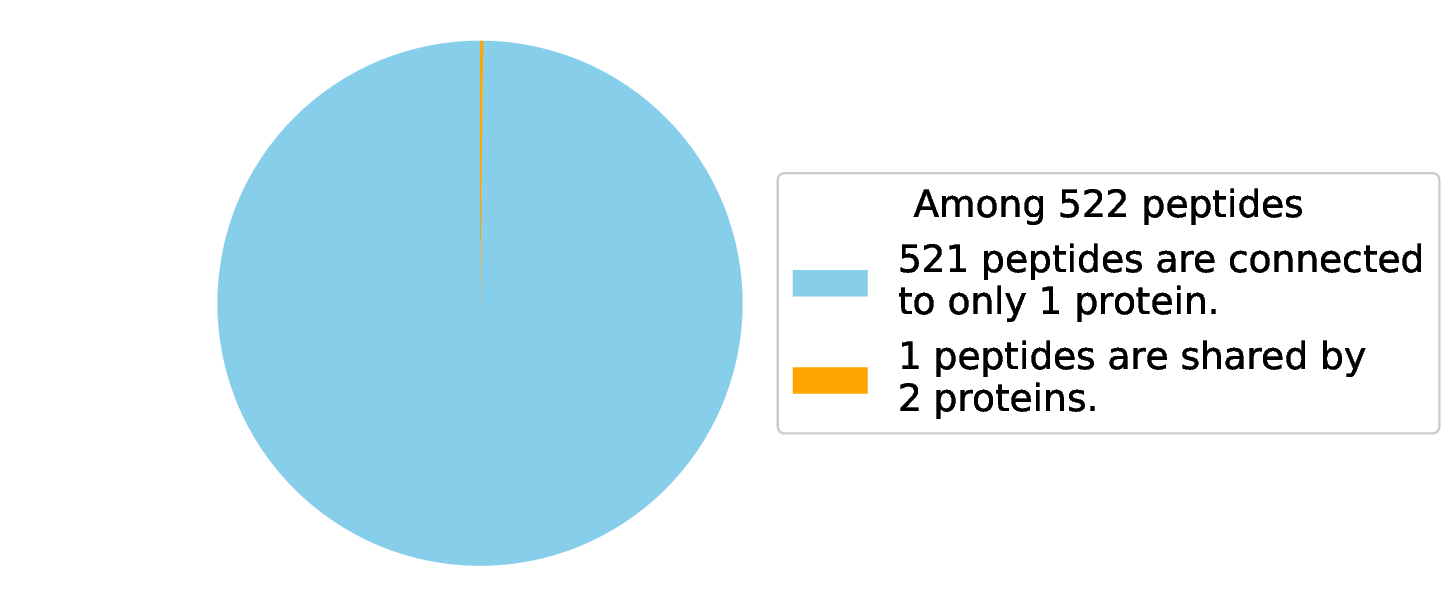}
    \caption{UPS2}
    \label{fig: ups2_pepshare}
\end{subfigure}\hspace{\fill} % maximize horizontal separation
\hfill
\begin{subfigure}[t]{0.49\textwidth}
\centering
    \includegraphics[width=3.2in]{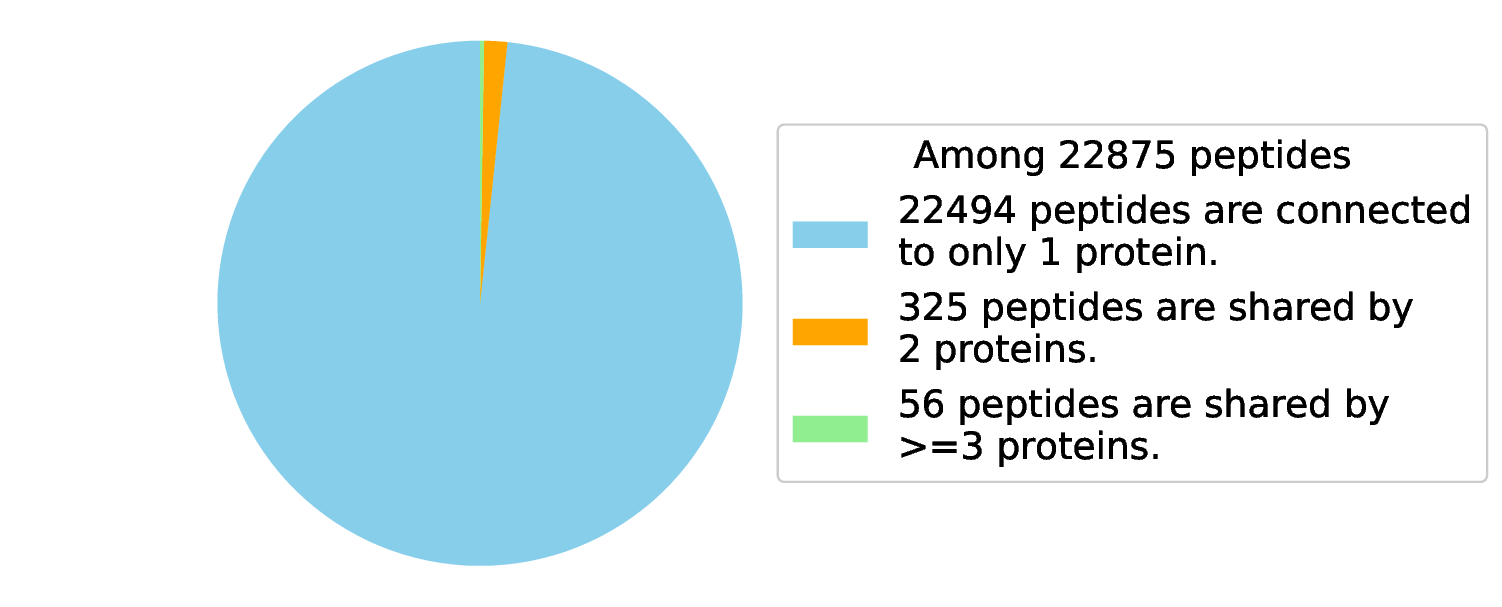}
    \caption{18Mix}
    \label{fig: 18mix_pepshare}
\end{subfigure}

\begin{subfigure}[t]{0.49\textwidth}
\centering
    \includegraphics[width=3.2in]{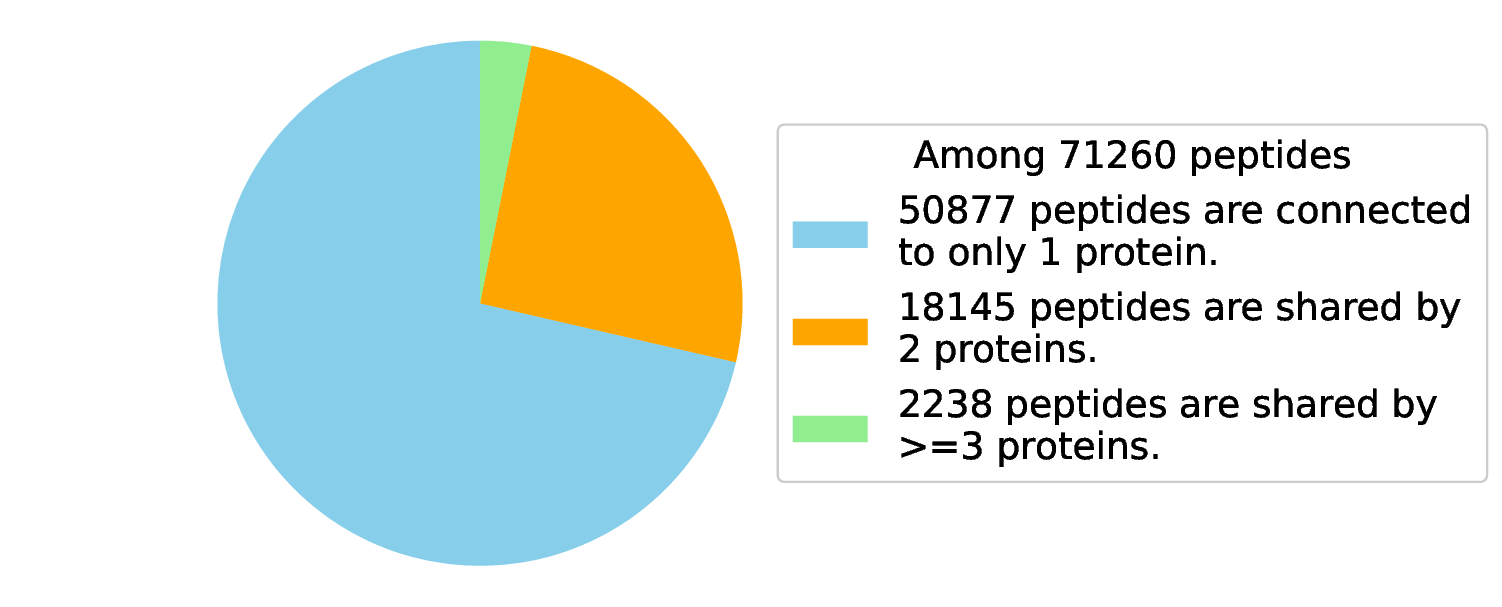}
    \caption{Hela}
    \label{fig: hela_pepshare}
\end{subfigure}\hspace{\fill} % maximize horizontal separation
\hfill
\begin{subfigure}[t]{0.49\textwidth}
\centering
    \includegraphics[width=3.2in]{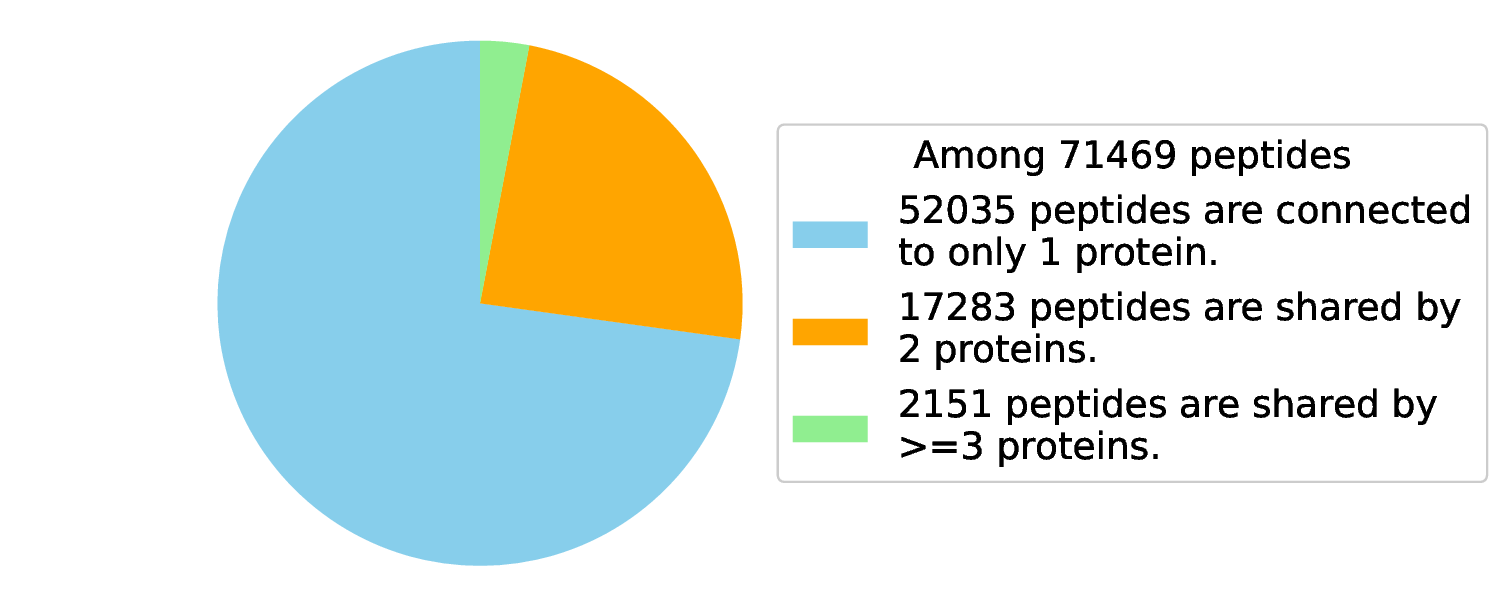}
    \caption{3T3}
    \label{fig: 3t3_pepshare}
\end{subfigure}

\caption{Portion of peptides connected to 1, 2, or $\geq3$ proteins.}
\label{fig: pepshare}
\end{figure}

% \newpage
\clearpage
\section{Supporting Table S3: Software Versions and Configurations in Our Experiments}

\begin{table}[h!]
\footnotesize
\centering
\begin{tabular}{| l | l | l | }
\hline
\textbf{Software} & \textbf{Version} & \textbf{Remark}\\
    \hline 
    msConvert~\citep{chambers2012msconvert} & 3.0.22208-6839020 & Centroid with vendor peaks picking \\
    \hline
    KNIME \cite{berthold2009knime} & 4.7.7 & \\
    \hline
    OpenMS~\citep{rost2016openms} & 2.7.0.202109131426 & \\
    \hline
    Percolator~\cite{kall2007semi} & within OpenMS & \\
    \hline
    Comet~\citep{eng2015deeper} & within OpenMS & \\
    \hline
    Epifany~\citep{pfeuffer2020epifany} & within OpenMS & greedy\_group\_resolution set to remove\_proteins\_wo\_evidence \\
    \hline
    Fido~\citep{serang2010efficient} & within OpenMS & greedy\_group\_resolution set to True \\
    \hline
    PIA~\citep{uszkoreit2015pia} & 1.4.6.v202206101448 & ``Report all'' mode \\
    \hline
    DeepPep~\cite{kim2017deeppep} & v1 & https://github.com/IBPA/DeepPep \\
    \hline
    PyTorch & 1.13.1 & \\
    \hline
    CUDA & 11.7 & \\
    \hline
    CuDNN & 8.5.0 & \\
    \hline
    Torch-Geometric & 2.2.0 & \\
    \hline
\end{tabular}
\caption{Software versions.}
\label{table:software_versions}
\end{table}

\section{Supporting Figure S3: Accuracy of FDR Estimation}

\begin{figure}[t!]
%\captionsetup[subfigure]%{justification=Centering}
\centering
\begin{subfigure}[t]{0.49\textwidth}
\centering
    \includegraphics[width=2.7in]{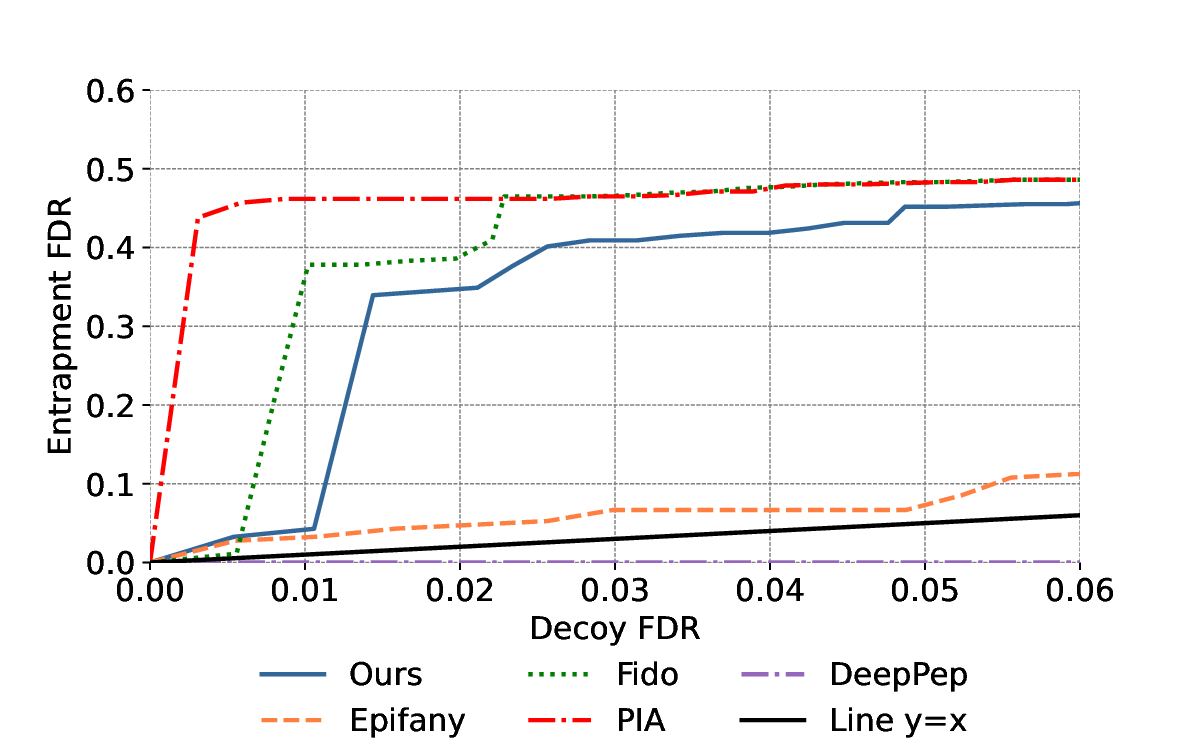}
    \caption{iPRG2016 A}
    \label{fig: iprg_a_fdr}
\end{subfigure}\hspace{\fill} % maximize horizontal separation
\hfill
\begin{subfigure}[t]{0.49\textwidth}
\centering
    \includegraphics[width=2.7in]{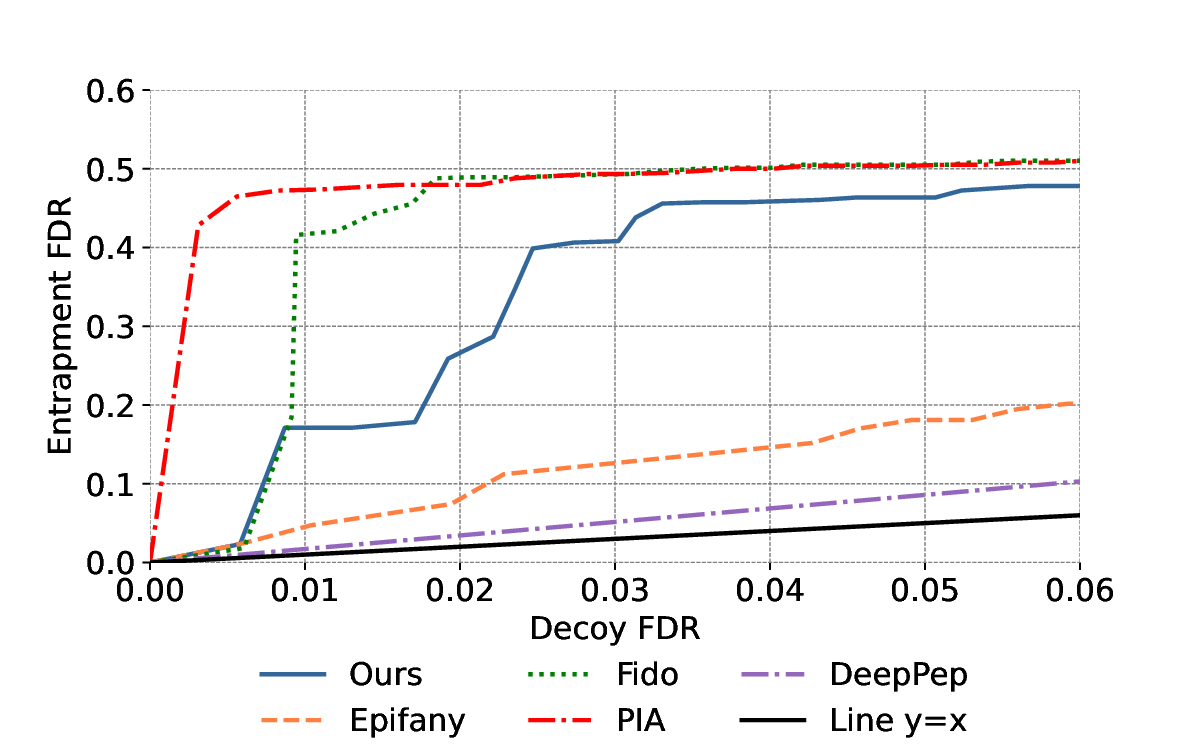}
    \caption{iPRG2016 B}
    \label{fig: iprg_b_fdr}
\end{subfigure}

 % Add a line break after the second image

\begin{subfigure}[t]{0.49\textwidth}
\centering
    \includegraphics[width=2.7in]{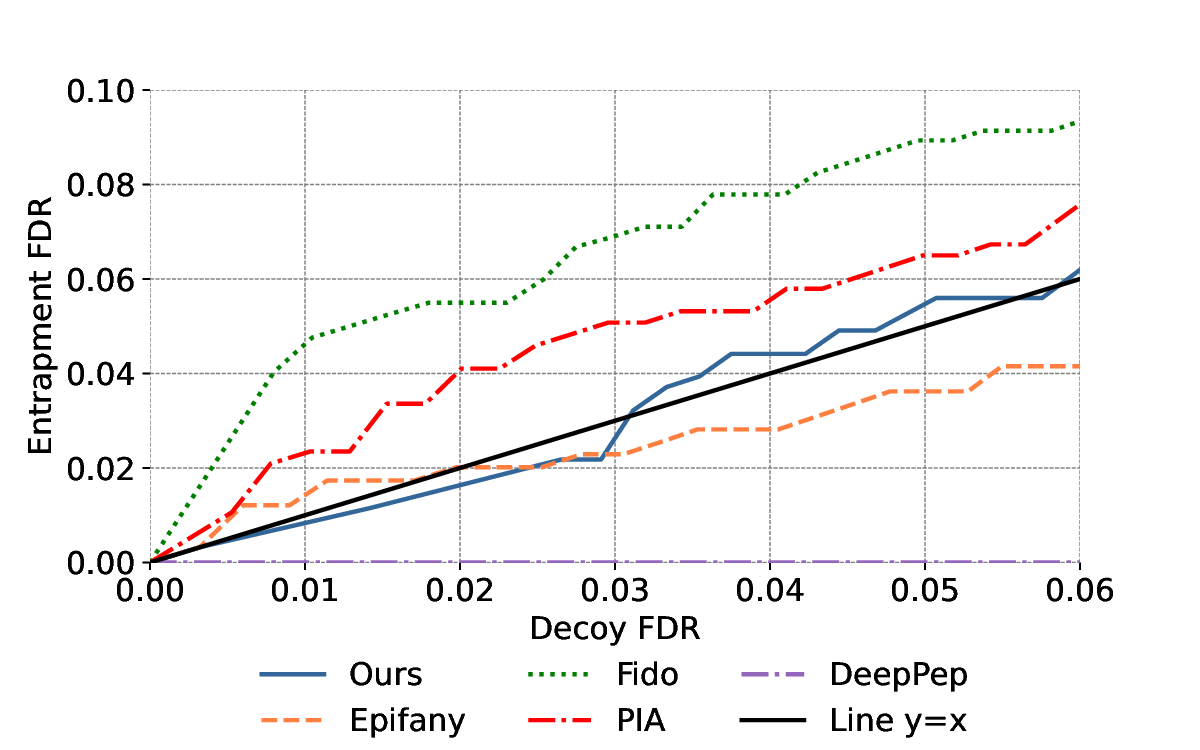}
    \caption{iPRG2016 AB}
    \label{fig: iprg_ab_fdr}
\end{subfigure}\hspace{\fill} % maximize horizontal separation
\hfill
\begin{subfigure}[t]{0.49\textwidth}
\centering
    \includegraphics[width=2.7in]{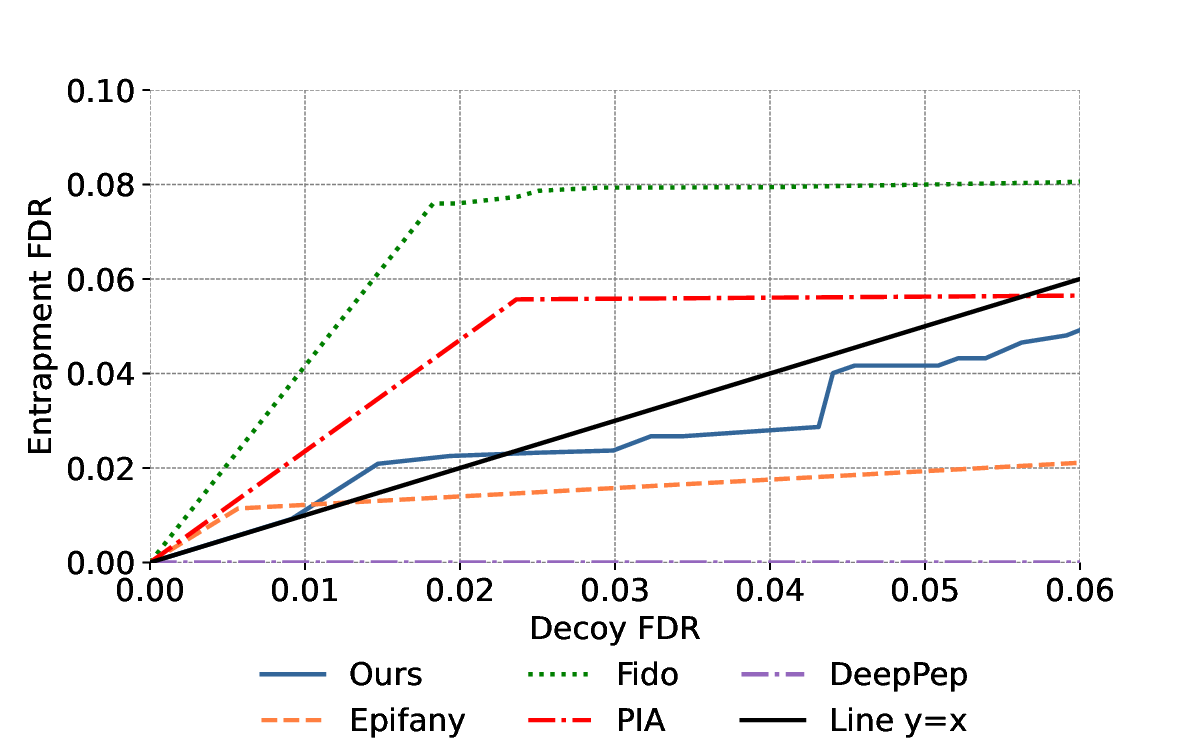}
    \caption{Yeast}
    \label{fig: yeast_fdr}
\end{subfigure}

%\newline % Add a line break after the fourth image

\begin{subfigure}[t]{0.49\textwidth}
\centering
    \includegraphics[width=2.7in]{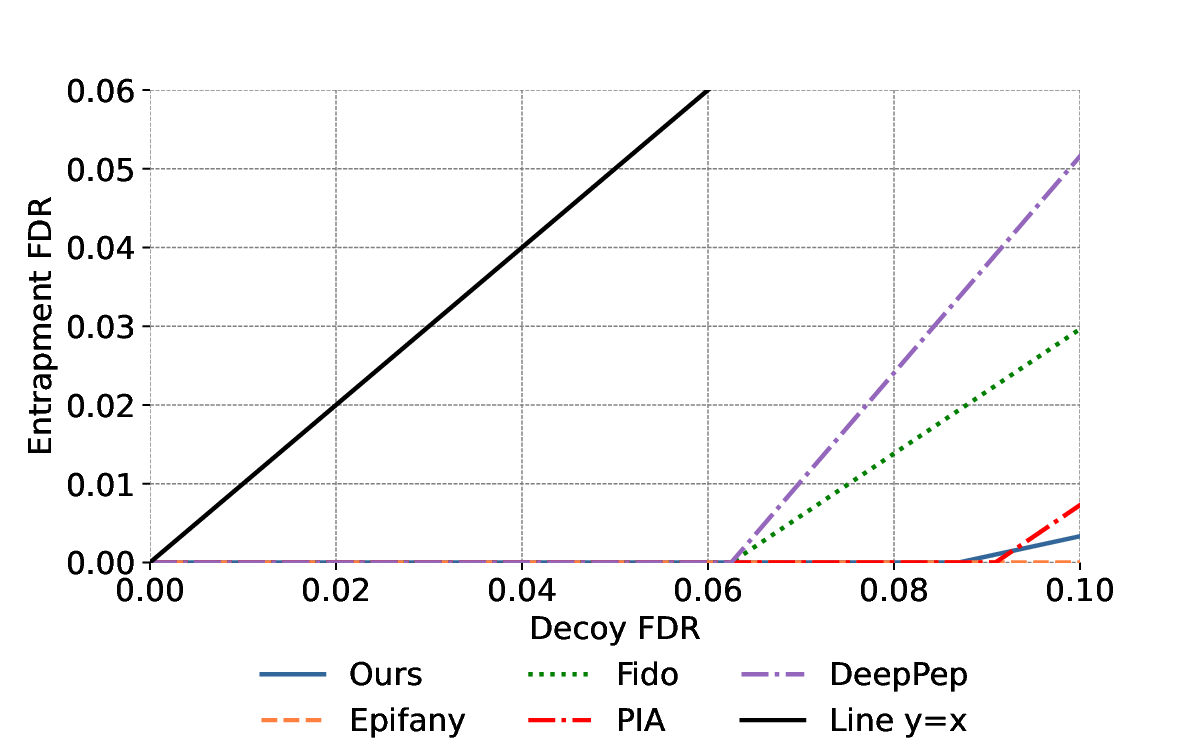}
    \caption{UPS2}
    \label{fig: ups2_fdr}
\end{subfigure}\hspace{\fill} % maximize horizontal separation
\hfill
\begin{subfigure}[t]{0.49\textwidth}
\centering
    \includegraphics[width=2.7in]{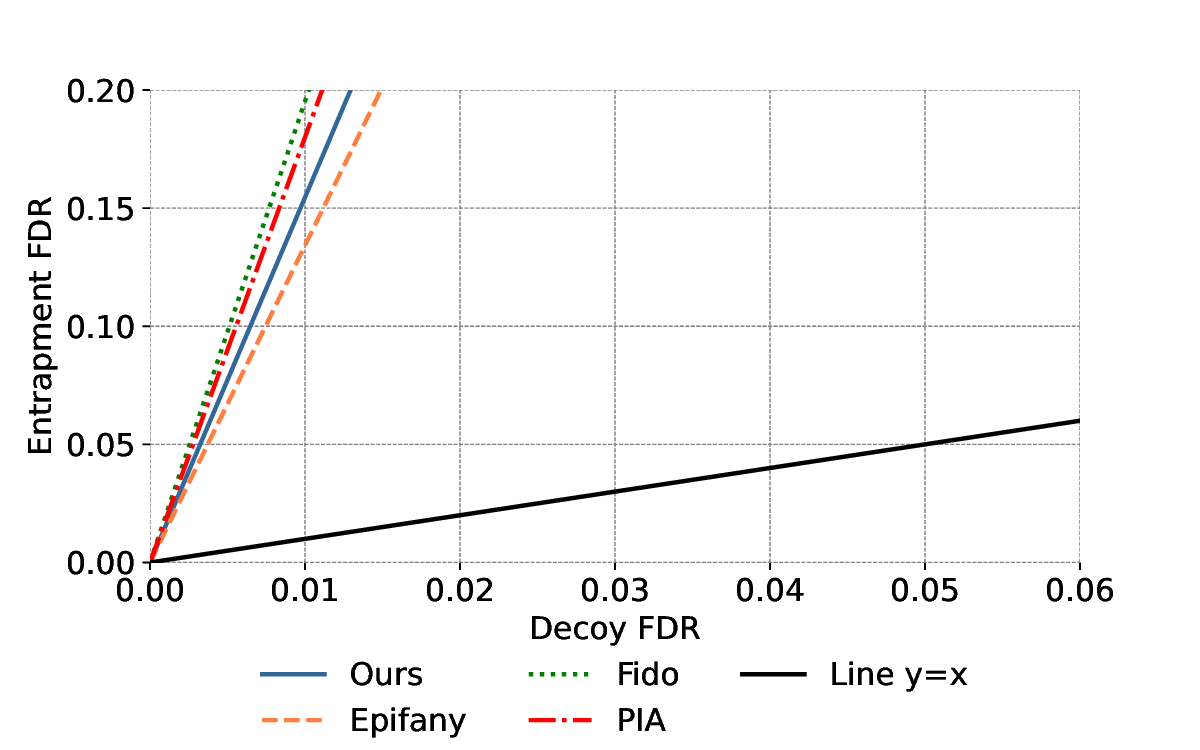}
    \caption{18Mix}
    \label{fig: 18mix_fdr}
\end{subfigure}

\begin{subfigure}[t]{0.49\textwidth}
\centering
    \includegraphics[width=2.7in]{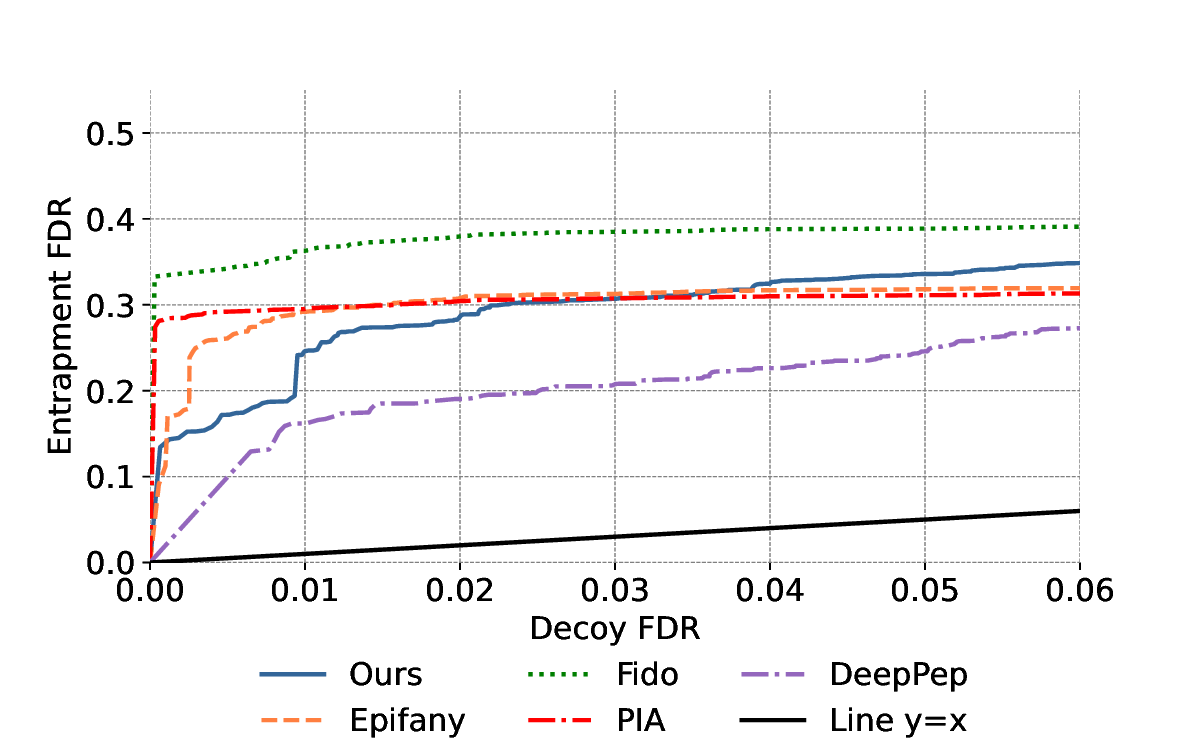}
    \caption{Hela}
    \label{fig: hela_fdr}
\end{subfigure}\hspace{\fill} % maximize horizontal separation
\hfill
\begin{subfigure}[t]{0.49\textwidth}
\centering
    \includegraphics[width=2.7in]{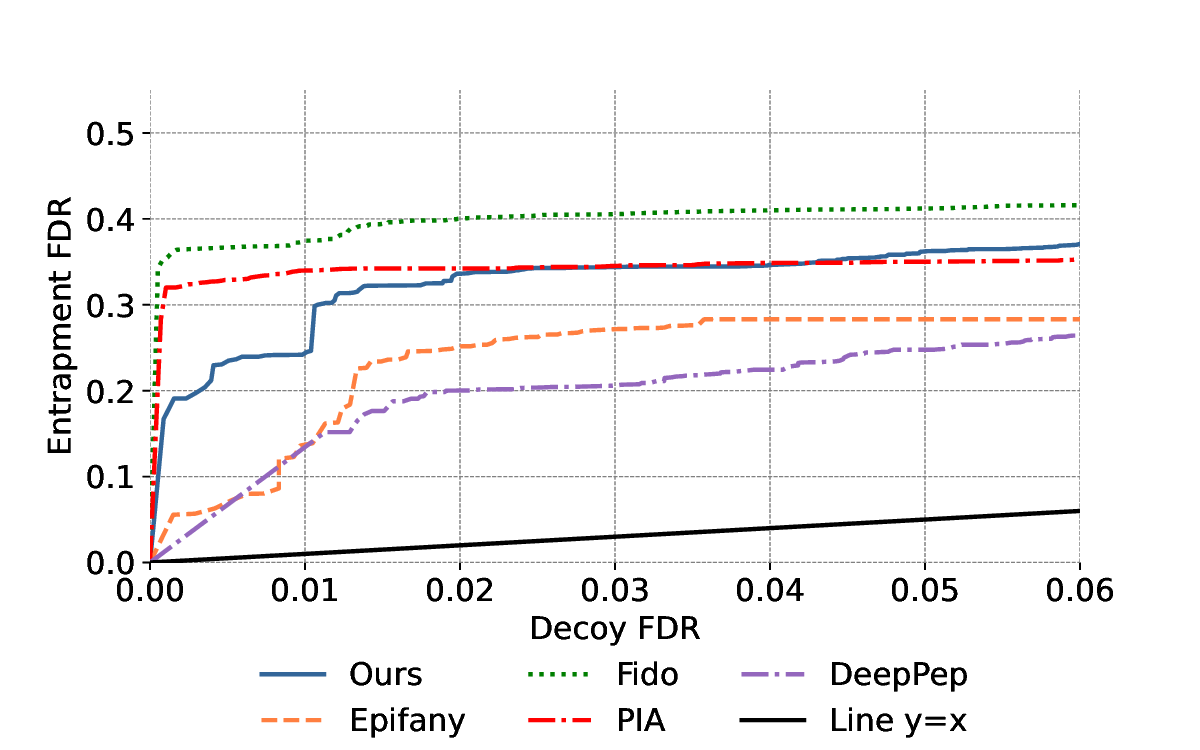}
    \caption{3T3}
    \label{fig: 3t3_fdr}
\end{subfigure}

\caption{Relationship between decoy FDR and entrapment FDR for different models, demonstrating the accuracy of FDR estimate. The dashed line is a straight line from (0, 0) to (1, 1), demonstrating the perfect FDR estimate.}
\label{fig: fdr}
\end{figure}

Figure~\ref{fig: fdr} illustrates the relationship between decoy FDR (our estimated FDR) and entrapment FDR. Generally, Epifany provides the most conservative FDR estimate in iPRG2016 A and B, while our model lies between Epifany and Fido. In iPRG2016 AB, Yeast and 18Mix dataset, our model yields the most accurate FDR estimate. {In hela, GraphPI offered the second best estimate, after DeepPep, while being after DeepPep and Epifany in 3t3. \color{blue}} Every model gets irregular FDR estimate in UPS2, making this dataset an outlier.

% We need to note that observing higher entrapment FDR at the same level of decoy FDR does not mean an inferior identification performance. It just suggests that decoy proteins are ranked behind positive and entrapment ones. Figure ~\ref{fig:id} illustrates that, under specific decoy FDR, our model identifies more proteins than Epifany in the iPRG2016 B dataset, hence higher entrapment FDR. Such differences can be easily calibrated.

Observing that a higher entrapment FDR at a consistent decoy FDR level doesn't necessarily imply subpar identification performance. Instead, it indicates that decoy proteins are ranked after both positive and entrapment ones. As illustrated in Figure~\ref{fig:id}, at a specific decoy FDR, our model identifies a greater number of proteins than Epifany in the iPRG2016 B dataset, leading to a higher entrapment FDR. Such disparities can be readily calibrated.

% \begin{figure}[t!]
% \centering
% \includesvg[height=7cm]{figures/fdr_estimate.svg}  
% \caption{\kcomment{make the fonts bigger and don't forget to change the name!}}
% \label{fig: fdr}
% \end{figure}

\begin{figure}[t!]
\centering
\begin{subfigure}[t]{0.49\textwidth}
\centering
    \includegraphics[width=2.7in]{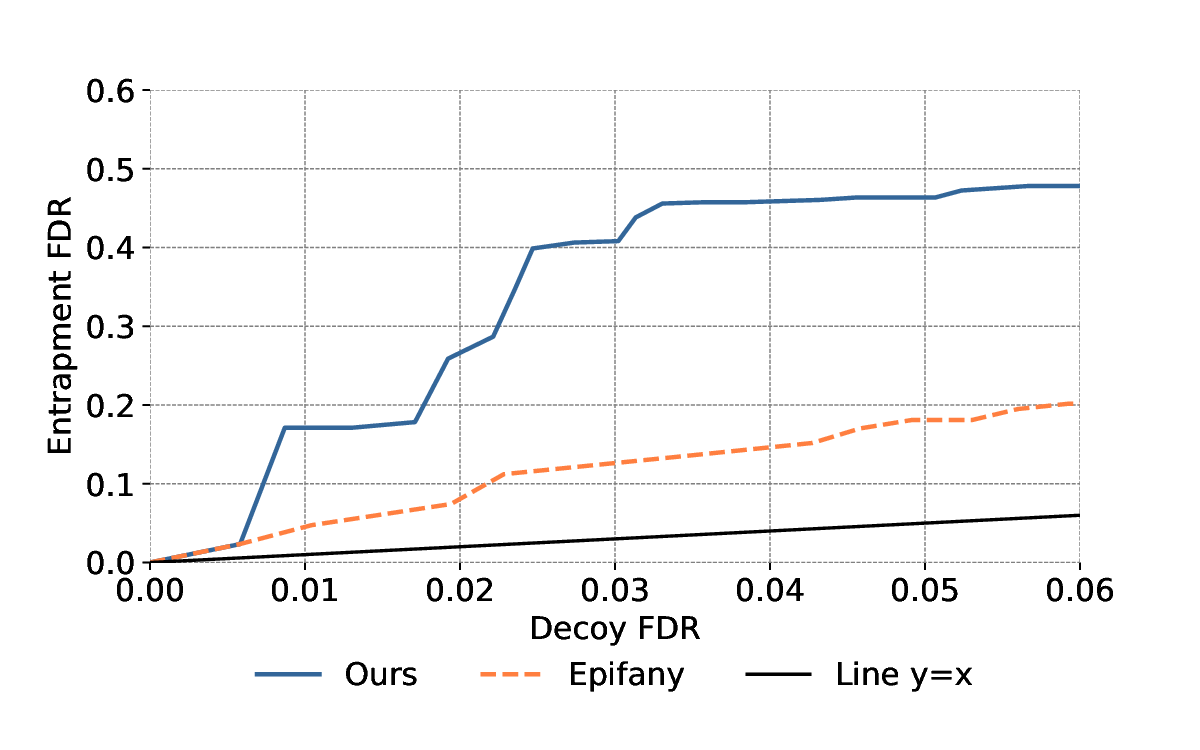}
    \caption{}
    \label{fig: iprga_oursvepifany_fdr}
\end{subfigure}\hspace{\fill} % maximize horizontal separation
\hfill
\begin{subfigure}[t]{0.49\textwidth}
\centering
    \includegraphics[width=2.7in]{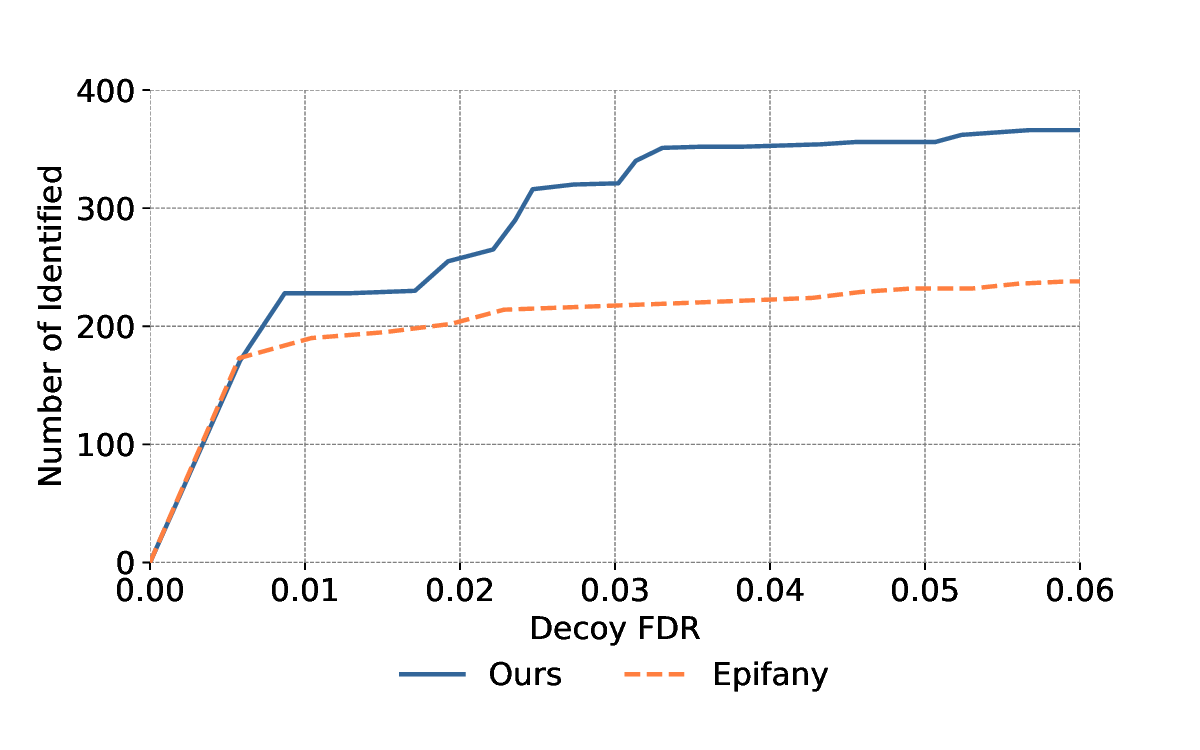}
    \caption{}
    \label{fig: iprga_oursvepifany_id}
\end{subfigure}

\caption{(a) shows the relationship between decoy FDR and entrapment FDR for iPRG2016 B, (b) shows the number of identified proteins under different decoy FDR values.}
\label{fig:id}
\end{figure}

%% file: chapters/Supplementary_nonhuman_comparison.tex
\section{Supporting Material S2: Alternative Settings for Pretraining}

% We conducted a comparative analysis to evaluate the impact of using isoform versus non-isoform databases on peptide searches. Specifically, we utilized the UniProt non-isoform database to search against the datasets listed in Section~\ref{table
% }, and the obtained PSM features were adopted to pretrain GraphPI. As shown in Figure~\ref{fig: comparison_pauc_otherdatabase}, the model trained on the isoform database slightly outperformed the model trained on the non-isoform database. This improved performance is likely due to the increased number of degenerate proteins in the isoform database, which facilitates better learning of the intricate relationships between proteins and peptides. Nonetheless, the differences were not substantial, suggesting that our model is relatively robust to the choice between isoform and non-isoform databases.

\begin{figure}[h!]
\centering
\includegraphics[width=1\linewidth]{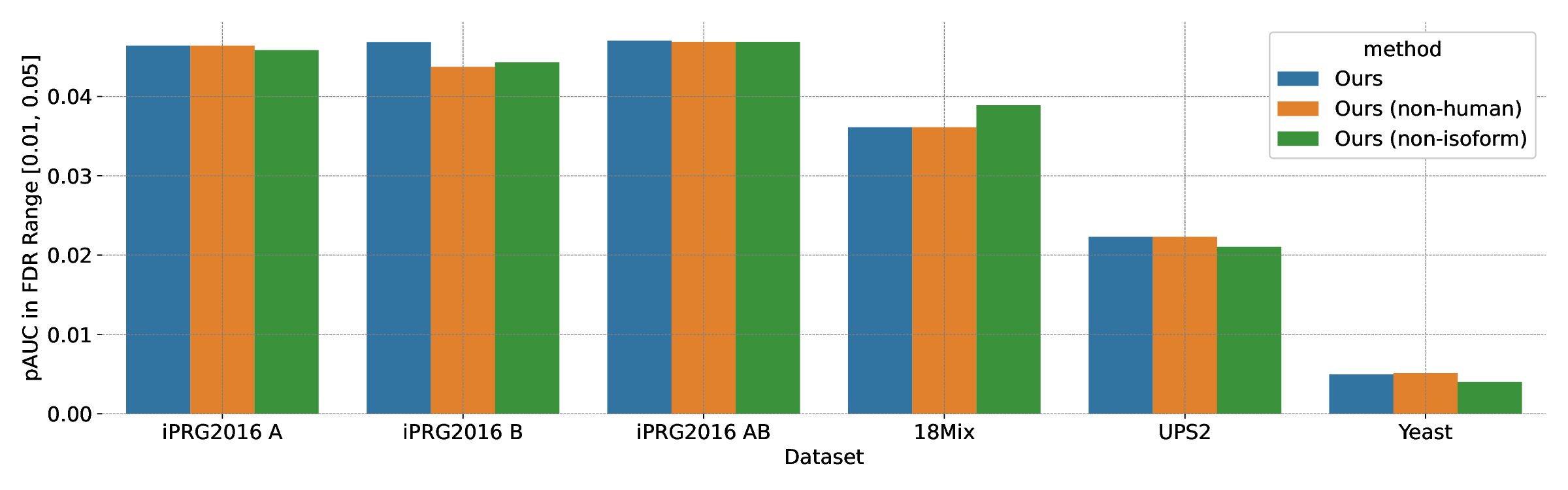}
\caption{pAUC (partial AUC) score of GraphPI trained on human isoform database (Ours), non-human isoform database (Ours non-human), and human non-isoform database (Ours (non-isoform)).}
\label{fig: comparison_pauc_otherdatabase}
\end{figure}

We conduct a comparative analysis to evaluate the impact of using isoform versus non-isoform databases on peptide searches, utilizing both human and non-human protein datasets. Specifically, we implemented two additional settings: 1) Using the UniProt non-isoform database to search against the datasets listed in Table~\ref{table:training_datasets_search}. 2) Searching against a list of non-human protein datasets using their respective isoform databases. Details of these datasets are provided in Table~\ref{table:training_datasets_search_nonhuman}.
The performance evaluated on the test sets iPRG 2016, 18Mix, UPS2, and Yeast in terms of pAUC are presented in Figure~\ref{fig: comparison_pauc_otherdatabase}. 

The results for both human isoform and non-human isoform datasets were similar, indicating that our method can effectively generalize protein-peptide structures across different species, This generalization is attributed to the inductive properties of Graph Neural Networks (GNNs), which leverage specific node features and edge weights to learn context-based graph structures. Such an approach enables the model trained on one dataset to extend its learning effectively to other datasets. Additionally, our model trained on the isoform database slightly outperformed the model trained on the non-isoform database. Nonetheless, the differences were not substantial, suggesting that our model is relatively robust to the choice between isoform and non-isoform databases.

% This improved performance is likely due to the increased number of degenerate proteins in the isoform database, which facilitates better learning of the intricate relationships between proteins and peptides. 

\section{Supporting Material S3: Comparison of Datasets from Other Species}

We further evaluated GraphPI on various public protein datasets from species other than humans. The parameter settings for the Comet search of these datasets are listed in Table~\ref{table:training_datasets_search_nonhuman}. 
% {\color{yellow}The species of these datasets are as follows: Escherichia coli (PXD040963 and PXD041209), Drosophila melanogaster (Fruit fly) (PXD044924), Limosilactobacillus fermentum (PXD050857), Mus musculus (Mouse) (PXD014146 and PXD051778), and Rattus norvegicus (Rat) (PXD047159 and PXD048641).}

\begin{figure}[t!]
    \centering
    \includegraphics[width=1\linewidth]{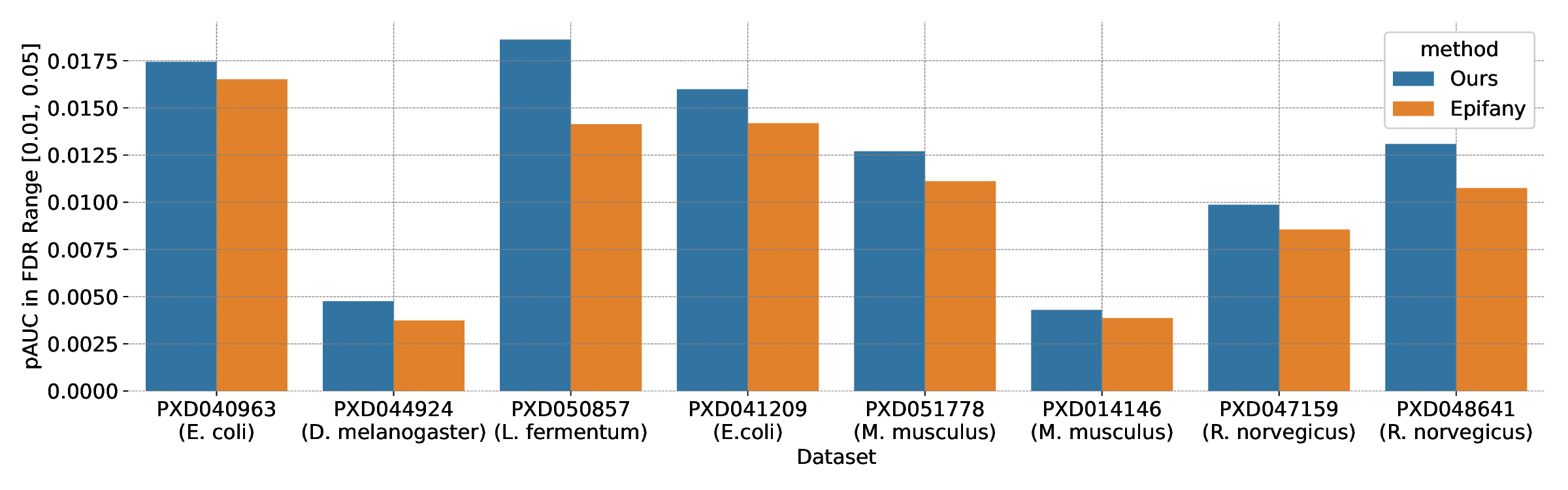}
    \caption{pAUC (partial AUC) score of GraphPI and Epifany on non-human PXD datasets with decoy proteins as contaminate proteins.}
    \label{fig: comparison nonhuman database}
\end{figure}

Given that these datasets do not have ground truth labels for evaluating entrapment FDR, we employed the target-decoy approach, with pAUC evaluated based on the Decoy FDR. Figure~\ref{fig: comparison nonhuman database} presents the pAUC performance of GraphPI compared to the best-performing baseline, Epifany. The results demonstrate the consistently superior performance of GraphPI across these non-human datasets.

We attribute this consistent perfromance to the inherent capabilities of GNNs, which facilitate the learning of patterns and structures from one set of data and apply this knowledge to new, unseen data. Meanwhile, by training GraphPI on a diverse range of datasets with varying structures, we enable it to learn a flexible representation of the graph that generalize well across different datasets.

\begin{table}[t!]
\footnotesize
\centering
\resizebox{\linewidth}{!}{
\begin{tabular}{| l |l | l | l | l | l | l | l | l |}
\hline
\textbf{Identifier}  & \textbf{\makecell{MS1\\ tolerance}} & \textbf{\makecell{MS2 \\tolerance}} & \textbf{Modifications} & \textbf{\makecell{Missed \\Cleavages}} & \textbf{Digestion} & \textbf{\# Replicates}  & \textbf{Species} \\
    \hline
    PXD040963 & 10 ppm & 0.01 Da & \makecell[l]{C(+57.02), M(+15.99)} & 2 & Trypsin &11 & E. coli \\
    \hline
    PXD044924 &   20 ppm & 0.02 Da & \makecell[l]{C(+57.02), M(+15.99) \\ Protein N-term Acetyl} & 2 & Trypsin & 7 & D. melanogaster \\
    \hline
    PXD050857 & 20 ppm & 0.01 Da & \makecell[l]{C(+57.02), M(+15.99)} & 2 & Trypsin & 4 &  L. fermentum \\
    \hline
    PXD041209 & 20 ppm & 0.02 Da & \makecell[l]{C(+57.02), M(+15.99)\\ NQ(+0.98), Acetyl(K) \\N-term Acetyl} & 3 & Trypsin & 10 & E. coli \\
    \hline
    PXD051778 &  10 ppm & 0.01 Da & \makecell[l]{C(+57.02), M(+15.99)\\ Protein N-term Acetyl\\ NQ(+0.98)} & 2 & Trypsin & 4 & M. musculus \\
    \hline
    PXD014146 &  20 ppm & 0.5 Da & \makecell[l]{C(+57.02), M(+15.99) \\ Protein N-term Acetyl \\ STY(+79.99)} & 2 & Trypsin & 5 & M. musculus \\
    \hline
    PXD047159 & 	 10 ppm & 0.01 Da & \makecell[l]{C(+57.02), M(+15.99)\\ Protein N-term Acetyl} & 2 & Trypsin & 8 & R. norvegicus \\
    \hline
    PXD048641 &  10 ppm & 0.01 Da & \makecell[l]{C(+57.02), M(+15.99) \\Protein N-term Acetyl} & 2 & Trypsin & 9 & R. norvegicus \\
    \hline
    
    PXD024746 & 10 ppm & 0.01 Da & \makecell[l]{C(+57.02), MP(+15.99) \\Protein N-term Acetyl} & 2 & Trypsin & 24 & M. musculus \\
    \hline
    PXD048393 & 20 ppm & 0.01 Da & \makecell[l]{C(+57.02),\\N-term Acetyl} & 2 & Trypsin & 7 & M. musculus \\
    \hline
    PXD045340 & 10 ppm & 0.6 Da & \makecell[l]{C(+57.02), M(+15.99)\\Protein N-term Acetyl} & 2 & Trypsin & 7 & A. thaliana \\
    \hline
    PXD040196 & 10 ppm & 0.01 Da & \makecell[l]{C(+57.02), M(+15.99) \\ K(+42.02)} & 2 & Trypsin & 4 & M. musculus \\
    \hline
    PXD038241 & 10 ppm & 0.01 Da & \makecell[l]{C(+57.02), M(+15.99) \\N-term Acetyl, NQ(+0.98)\\ STY(+79.99)} & 3 & Trypsin/P & 10 & M. musculus \\
    \hline
    PXD039581 & 10 ppm & 0.02 Da & \makecell[l]{C(+57.02), M(+15.99) \\ NQ(+0.98)} & 2 & Trypsin & 9 & E. coli \\
    \hline
    PXD049971 & 7 ppm & 0.5 Da & \makecell[l]{C(+57.02), M(+15.99) \\ Protein N-term Acetyl\\
    R(+14.02), R(+28.03)
    } & 3 & Trypsin  & 7 & A. thaliana \\
    \hline
\end{tabular}}
\caption{Search parameters for non-human datasets.}
\label{table:training_datasets_search_nonhuman}
\end{table}